%% file: main.tex
\PassOptionsToPackage{table,x11names}{xcolor}
\documentclass[12pt,letterpaper]{article}
\usepackage[T1]{fontenc}
\usepackage[utf8]{inputenc}
\usepackage[a4paper,total={7in,10in}]{geometry}

\usepackage{helvet}
\usepackage{authblk}
\usepackage{orcidlink}
\usepackage{graphicx}
\usepackage{xcolor}
\usepackage{xurl}
\usepackage{amsmath,amssymb}
\usepackage{booktabs,multirow,longtable,tabularx,array}
\usepackage{subcaption,float,makecell,bm,setspace,pifont}
\usepackage{soul,calc,etoolbox}
\usepackage{chngcntr}
\usepackage{appendix}
\usepackage{numcompress}
\usepackage[super,comma,sort&compress]{natbib}
\usepackage{hyperref}
\usepackage[capitalize,noabbrev]{cleveref}
\newcommand{\Checkmark}{\ding{51}}
\newcommand{\XSolid}{\ding{55}}

\makeatletter
\renewcommand{\maketitle}{\bgroup\setlength{\parindent}{0pt}
\begin{flushleft}
  {\LARGE\bfseries\@title\par}
  \vspace{1em}
  \@author
\end{flushleft}\egroup}
\makeatother

\title{Ensemble learning of pathology foundation models for precision oncology}
\date{}
\author[1,7]{Xiangde Luo\orcidlink{0000-0001-8574-0005}}
\author[1,7]{Xiyue Wang\orcidlink{0000-0002-3597-9090}}
\author[1]{Feyisope Eweje}
\author[2]{Xiaoming Zhang}
\author[3]{Juan Luis Gomez Marti}
\author[3]{Sarah Cascarino}
\author[1]{Sen Yang}
\author[1]{Yuchen Li}
\author[4]{Ryan Quinton}
\author[1]{Jinxi Xiang}
\author[1]{Yuanfeng Ji}
\author[1]{Zhe Li}
\author[1]{Yijiang Chen}
\author[4]{Colin Bergstrom}
\author[1]{Ted Kim}
\author[4]{Francesca Maria Olguin}
\author[2]{Kelley Yuan}
\author[5]{Matthew Abikenari}
\author[1]{Andrew Heider}
\author[4]{Sierra Willens}
\author[4]{Sanjeeth Rajaram}
\author[2]{Robert West}
\author[4]{Joel Neal}
\author[3]{Adam Schoenfeld}
\author[1]{Maximilian Diehn}
\author[3]{Chad Vanderbilt}
\author[1,6,*]{Ruijiang Li\orcidlink{0000-0002-0232-5998}}
\affil[1]{Department of Radiation Oncology, Stanford University School of Medicine, Stanford, CA, USA}
\affil[2]{Department of Pathology, Stanford University School of Medicine, Stanford, CA, USA}
\affil[3]{Department of Pathology and Laboratory Medicine, Memorial Sloan Kettering Cancer Center, New York, NY, USA}
\affil[4]{Department of Medicine (Oncology), Stanford University School of Medicine, Stanford, CA, USA}
\affil[5]{Department of Neurosurgery, Stanford University School of Medicine, Stanford, CA, USA}
\affil[6]{Stanford Institute for Human-Centered Artificial Intelligence, Stanford, CA, USA}
\affil[7]{These authors contributed equally}
\affil[*]{Correspondence and Lead Contact: Ruijiang Li, \href{mailto:rli2@stanford.edu}{rli2@stanford.edu}}

\begin{document}
\maketitle

\section{SUMMARY}\label{summary-1}

Histopathology is essential for cancer diagnosis and treatment selection, and pathology foundation models learn visual representations from whole-slide images (WSIs). However, existing foundation models are trained on disparate datasets using varying strategies, leading to inconsistent performance and limited generalizability. Here, we introduce ELF (Ensemble Learning of Foundation models), which integrates five pretrained pathology foundation models into unified slide-level representations. Trained on 53,699 WSIs spanning 20 anatomical sites, ELF leverages ensemble learning to capture complementary information across models. ELF\textquotesingle s slide-level architecture is designed for data-efficient downstream evaluation, including settings with limited data such as therapeutic response prediction. We evaluate ELF for disease classification, biomarker detection, as well as anticancer and immunotherapy response prediction across multiple cancer types. ELF achieves higher performance than the evaluated constituent and slide-level foundation models across the tested tasks, supporting further evaluation of ensemble learning for pathology applications in oncology.

\section{KEYWORDS}\label{keywords}

Pathology foundation model; ensemble learning; precision oncology; biomarker detection; treatment response prediction

\clearpage
\section{INTRODUCTION}\label{introduction-1}

Histopathology is a cornerstone of clinical practice, offering critical insights into cellular morphology and tissue architecture. As the gold standard for disease diagnosis, it also plays a vital role in guiding prognosis and informing treatment decisions. With the growing availability of digitized whole-slide histopathology images and advances in computational power, artificial intelligence (AI) has emerged as a powerful tool for histopathology image analysis, which has significant implications for precision oncology\cite{bera2019artificial,yates2025new,bhinder2021artificial,van2021deep}. Among the most promising developments in this field are pathology foundation models, large-scale models pretrained using self-supervised learning on vast collections of histopathology slides\cite{xu2024whole,wang2024pathology,xiang2025vision,hoptimus0,chen2024towards,lu2024visual,vorontsov2024foundation}. These models can be fine-tuned for a wide range of downstream clinical tasks, with the potential to enhance diagnostic accuracy, identify novel prognostic and predictive biomarkers, and support personalized treatment strategies\cite{wang2025foundation,campanella2025real,kondepudi2025foundation}.

There has been a rapid proliferation of pathology foundation models in the last few years, with at least 20 developed and counting\cite{lipkova2024age}. These foundation models are typically trained on proprietary datasets with different training strategies\cite{chen2021empirical,he2022masked}, leading to model-specific biases and divergent performance across applications\cite{neidlinger2024benchmarking,ma2024towards}. Indeed, recent benchmark studies\cite{neidlinger2024benchmarking,ma2025pathbench,campanella2025clinical} based on extensive evaluation have shown that the relative performance of pathology foundation models is highly variable across different tasks and even across datasets for the same task. In other words, no single foundation model outperforms others across all tasks. This creates a conundrum in the community about which foundation model one should use for a particular application.

There is an emerging interest in combining the relative strengths of different foundation models. One recent study attempts to resolve the differences among several existing models using knowledge distillation\cite{ma2024towards}. As a result, the model effectively learns an `average' feature representation, which may paradoxically reduce feature diversity and representation capability of the resulting model. This approach does not outperform existing foundation models as shown in benchmark studies\cite{neidlinger2024benchmarking,ma2025pathbench,bareja2025evaluating}. The optimal strategy that truly leverages the complementary value of different foundation models remains to be explored.

Here, we present ELF (Ensemble Learning of Foundation models), an ensemble framework for integrating multiple foundation models. ELF incorporates five pretrained pathology foundation models and generates a unified representation for the whole-slide image (WSI) through additional pretraining on 53,699 WSIs across 20 anatomical sites. As a pretrained slide-level foundation model, ELF is designed to improve data efficiency relative to traditional tile-level approaches and may therefore be useful when large datasets are not available for model fine-tuning. This is quite common in therapeutic response prediction which is focused on a particular drug regimen in a given indication and clinical setting.

We evaluated the performance of ELF on a wide range of clinical applications, including challenging disease diagnosis, molecular biomarker detection, and treatment response prediction. In particular, we tested ELF for predicting response across all major anticancer therapies including cytotoxic chemotherapy, molecularly targeted therapy, and immunotherapy in various cancer types. ELF achieved higher performance across the evaluated tasks than each of the five individual foundation models and the evaluated slide-level foundation models\cite{xu2024whole,wang2024pathology,ding2024multimodal}. These findings support further evaluation of ensemble learning for integrating pathology foundation models in precision medicine.

\begin{figure*}[p]
\centering
\includegraphics[width=\textwidth,height=1.0\textheight,keepaspectratio]{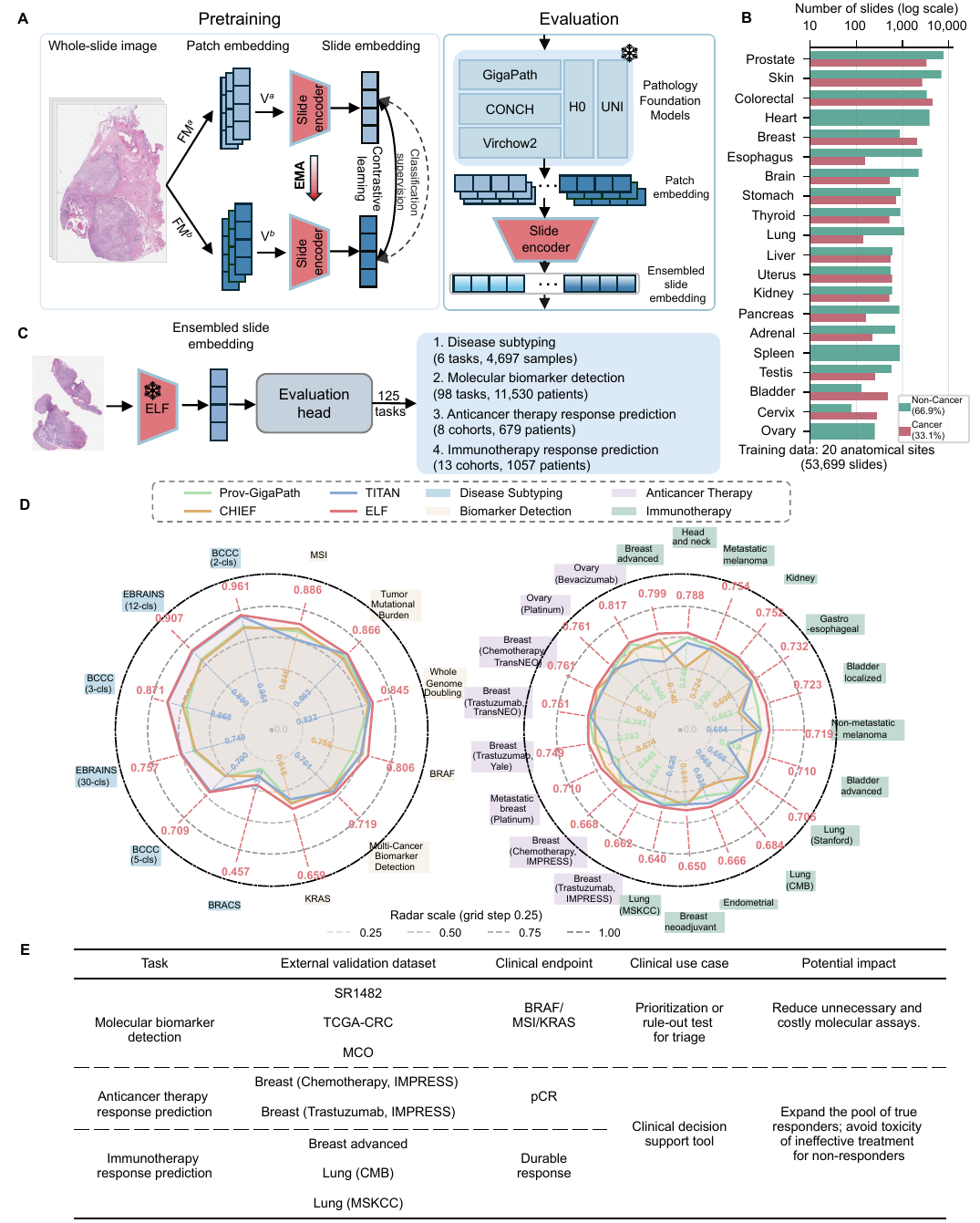}
\end{figure*}
\begin{figure*}[t]
\caption{\textbf{ELF integrates multiple pathology foundation models into a unified slide-level representation.}
\textbf{A.} Model pretraining. ELF is an ensemble framework that integrates multiple foundation models through a unified slide-level pathology foundation model. Here, five independently trained tile-level foundation models were employed to construct this ensemble (GigaPath\cite{xu2024whole}, CONCHV1.5\cite{lu2024visual,ding2024multimodal}, Virchow2\cite{zimmermann2024virchow2}, H-Optimus-0\cite{hoptimus0}, UNI\cite{chen2024towards}). During pretraining, ELF mainly employs contrastive learning to align slide-level embeddings derived from different foundation models (\(FM^{a}\) and \(FM^{b}\)) or distinct spatial views (\(V^{a}\), \(V^{b}\)) of the same whole-slide image (WSI). At the same time, two auxiliary supervision tasks (cancer detection and organ classification) were used to further enhance pan-cancer representation learning. These patch embeddings are encoded into slide representations via a shared slide encoder updated through EMA, enabling ELF to learn model-invariant and view-consistent representations. EMA: Exponential Moving Average.
\textbf{B,} Data distribution for ELF pretraining, which consists of 53,699 slides with clinical information across 20 anatomical sites from 11 public datasets.
\textbf{C,} ELF is evaluated across diverse clinical applications, including disease subtyping, biomarker detection, and response prediction for anticancer therapy and immunotherapy.
\textbf{D,} Across these clinical tasks, ELF consistently outperforms three leading slide-level pathology foundation models (Prov-GigaPath\cite{xu2024whole}, CHIEF\cite{wang2024pathology}, and TITAN\cite{ding2024multimodal}, radar plots scaled from 0 to 1). Performance is reported using balanced accuracy (BA) for disease subtyping, and area under the ROC curve (AUC) for biomarker detection and treatment response prediction.
\textbf{E,} Summary of external validation datasets, clinical endpoints, and potential impact across three predictive tasks: molecular biomarker detection, anticancer therapy response prediction, and immunotherapy response prediction.}
\label{fig:figure1}
\end{figure*}

\section{RESULTS}\label{results}

\subsection{Design and evaluation of ensemble learning of foundation models}\label{design-and-evaluation-of-ensemble-learning-of-foundation-models}

We proposed an ensemble learning approach (ELF) to integrate multiple independently trained foundation models into a single model (Figure~1). ELF generates a unified slide-level representation that combines the complementary strengths of five tile-level pathology foundation models: GigaPath\cite{xu2024whole}, CONCH\cite{lu2024visual}, Virchow2\cite{zimmermann2024virchow2}, H-Optimus-0\cite{hoptimus0}, and UNI\cite{chen2024towards}. These five models were chosen because of the large number of slides used for pretraining (\(>\)100,000) and their generally top-ranking performance (though not always) across benchmark studies\cite{neidlinger2024benchmarking,ma2025pathbench,campanella2025clinical}.

ELF was pretrained on 53,699 whole-slide images (WSIs) using unsupervised contrastive learning for feature alignment and weakly supervised learning for cancer detection and organ classification (Figures~1\textbf{A and} 1\textbf{B}). The architecture of ELF comprises two main components: (1) five pretrained tile-level feature extractors and (2) an ensemble module that synthesizes a unified slide representation. For inference, ELF processes a given slide through each base model to obtain per-model embeddings, which are then fused into a unified slide-level feature representation (Figure~1\textbf{C}). ELF was evaluated on a wide range of clinical applications, including classification and subtyping of 4,697 tumor samples, 98 combinations of biomarker-indication assessment in 11,530 patients, and therapeutic response prediction in 21 independent cohorts of 1,736 patients across 9 cancer types (Figures~1\textbf{C and} 1\textbf{D}). A summary of the external validation datasets, clinical endpoints, and potential applications across these predictive tasks is provided in Figure 1\textbf{E}. Detailed model architectures, training regimes, and evaluation protocols are provided in STAR Methods.

\subsection{Disease classification and subtyping}\label{disease-classification-and-subtyping}

We evaluated ELF on several diagnostic problems for disease classification and subtyping. These included skin cancer subtyping (2 cls, 3 cls, and 5 cls, cls means classes) using the Basal Cell Carcinoma Classification (BCCC) dataset\cite{yacob2023weakly} (1,831 samples after quality control), breast cancer subtyping (7 cls) using the BReAst Carcinoma Subtyping (BRACS) dataset (547 samples)\cite{brancati2022bracs}, and brain tumour subtyping (12 cls and 30 cls) using the EBRAINS dataset (2,319 samples)\cite{roetzer2022digital}. We compared ELF with three state-of-the-art slide-level foundation models (Prov-GigaPath\cite{xu2024whole}, CHIEF\cite{wang2024pathology}, and TITAN\cite{ding2024multimodal}), using balanced accuracy (BA) as the evaluation metric.

ELF consistently achieved the highest performance across all tasks, surpassing the second-best model, TITAN, by an average of 2.2\% across all datasets and up to 16.3\% on BRACS (Figure~2\textbf{A}). Specifically, on the BCCC dataset, ELF achieved BA of 0.961 (95\% confidence interval (CI): 0.941-0.979), 0.871 (95\% CI: 0.839-0.903), and 0.709 (95\% CI: 0.663-0.752) on the 2-, 3-, and 5 cls subtyping tasks, respectively. These results exceeded all baseline models, including TITAN (BA: 0.951, 0.868, and 0.700; \(P <\) 0.001 for all). Similarly, for brain tumour subtyping on EBRAINS, ELF achieved 0.907 (95\% CI: 0.869-0.944) BA (12 cls) and 0.757 (95\% CI: 0.716-0.795) (30 cls), outperforming TITAN by around 1\% (\(P <\) 0.001). The improvements in classification performance were more notable for challenging tasks. On the BRACS dataset (7 cls breast subtyping), ELF achieved a BA of 0.457 (95\% CI: 0.359-0.566), representing a 16.3\% relative improvement over TITAN (0.393), and substantially outperforming other models such as CHIEF (0.384) and Prov-GigaPath (0.323).

To assess the quality of slide-level representations, we performed t-SNE visualization of slide embeddings generated by ELF and Prov-GigaPath, which are both pure vision encoders. Using the EBRAINS dataset (12 cls) as an example, we evaluated their ability to separate histological subtypes in the embedding space (Figure~2\textbf{B}). ELF produced more distinct and coherent clusters corresponding to known brain tumor subtypes, including ependymal, lymphoid, and sellar tumors. These patterns were consistently observed in both development (the combined training and validation sets) and test sets. Additional details are presented in Figures S1A--S1C. Such well-separated histological representations contribute to improved classification accuracy of known disease subtypes and might facilitate unsupervised discovery of new disease categories.

To better understand ELF's performance, we conducted a comprehensive comparison to evaluate the effectiveness of its two major design components: integration of diverse foundation models and slide-level aggregation (Table S1 and Figure~2\textbf{C}). First, we evaluated whether combining different foundation models yielded better performance than using any single model alone (Figure~2\textbf{C}). On the BRACS dataset (7 classes), ELF achieved a BA of 0.457, substantially higher than any individual foundation model, including GigaPath with a BA of 0.323 (the first panel in Figure~2\textbf{C}). Similarly, on EBRAINS (30 classes), ELF achieved a BA of 0.757, outperforming the best individual model, Virchow2, with a BA of 0.702 (\(P < 0.001\); the second panel in Figure~2\textbf{C}). Second, we compared ELF with three existing tile aggregation methods, including mean pooling\cite{ding2024multimodal}, ABMIL\cite{ilse2018attention}, and COBRAII\cite{lenz2025unsupervised}, a pretrained slide-level aggregator. ELF significantly outperformed all aggregation methods across five foundation models (Virchow2, CONCHV1.5, UNI, GigaPath, and H0) on the BRACS and EBRAINS datasets (Table S1).

Further, we compared ELF with alternative strategies for the integration of foundation models. First, we evaluated simple yet efficient aggregation methods such as mean pooling and ABMIL (Table S1). On both evaluation datasets, ELF achieved higher performance, and in some instances, by a large margin (BA: 0.757 vs. 0.699 for ABMIL for EBRAINS, Table S1). Next, we compared ELF with more sophisticated model integration approaches through distillation\cite{ma2024towards} or concatenation\cite{lenz2025unsupervised} (Figure~2\textbf{D}). Again, ELF achieved the highest performance on BRACS (BA: 0.457 vs. 0.392 for GPFM) as well as EBRAINS (BA: 0.757 vs. 0.706 for COBRAII; \(P <\) 0.01; Figure~2\textbf{D}). Although both COBRAII and ELF integrate information from multiple foundation models, COBRAII learns attention weights by fusing features from multiple models and applies the same weights to tile-level features, whereas ELF jointly optimizes these weights under the supervision of all five models during pretraining (Table S2). To distinguish the effects of model design from those of pretraining dataset scale, we additionally retrained COBRAII on the same 50,000-slide dataset used for ELF. COBRAII\_50K did not close the performance gap with ELF across six classification tasks (average BA: 0.6881 vs. 0.7746; Table S3).

ELF integrates diverse pathology foundation models by leveraging their complementary representations of morphological features. A potential hypothesis is that each model may focus on distinct histological regions. To test this, we extracted tile-level embeddings from each individual foundation model and passed them through ELF's slide encoder to obtain attention scores for each tile. We then computed pairwise Pearson correlations between the attention maps generated by different models. The results revealed consistently low correlations, suggesting that different models indeed focus on non-overlapping morphological regions (Figure S2\textbf{A}). We further extracted slide-level embeddings from each individual foundation model's tile-level embeddings using ELF's slide encoder and measured their correlations (Figure S2\textbf{B}). The results indicated extremely low correlations, highlighting that different models encode patches in distinct ways. These findings highlighted ELF's ability to integrate diverse histological perspectives, consistent with the observed performance gains in complex disease classification and subtyping tasks.

We also investigated scaling laws in both data scaling and ensemble scaling aspects. Models were pre-trained with increasing amounts of data (5k, 10k, 20k, 30k, and full), while preserving consistent data distributions across tasks. As shown in Figure S3, performance improved with more pretraining data, but the marginal gains gradually diminished at larger data scales, reflecting a saturation effect commonly observed in large-scale representation learning\cite{filiot2025distilling,kaplan2025openmidnight}. Notably, challenging multi-class tasks such as EBRAINS (30 classes) continued to benefit from additional data, albeit with smaller incremental gains. We also examined ensemble scaling by increasing the ensemble size from 1 to 5, averaging results over all possible model combinations (Figure S4). Performance improved consistently with larger ensembles, with gains tapering off beyond four models. Together, these results demonstrated favorable but saturating scaling behavior with respect to both data volume and ensemble size, enabling strong performance without requiring excessively large training datasets.

\begin{figure*}[p]
\centering
\includegraphics[width=\textwidth,height=1.0\textheight,keepaspectratio]{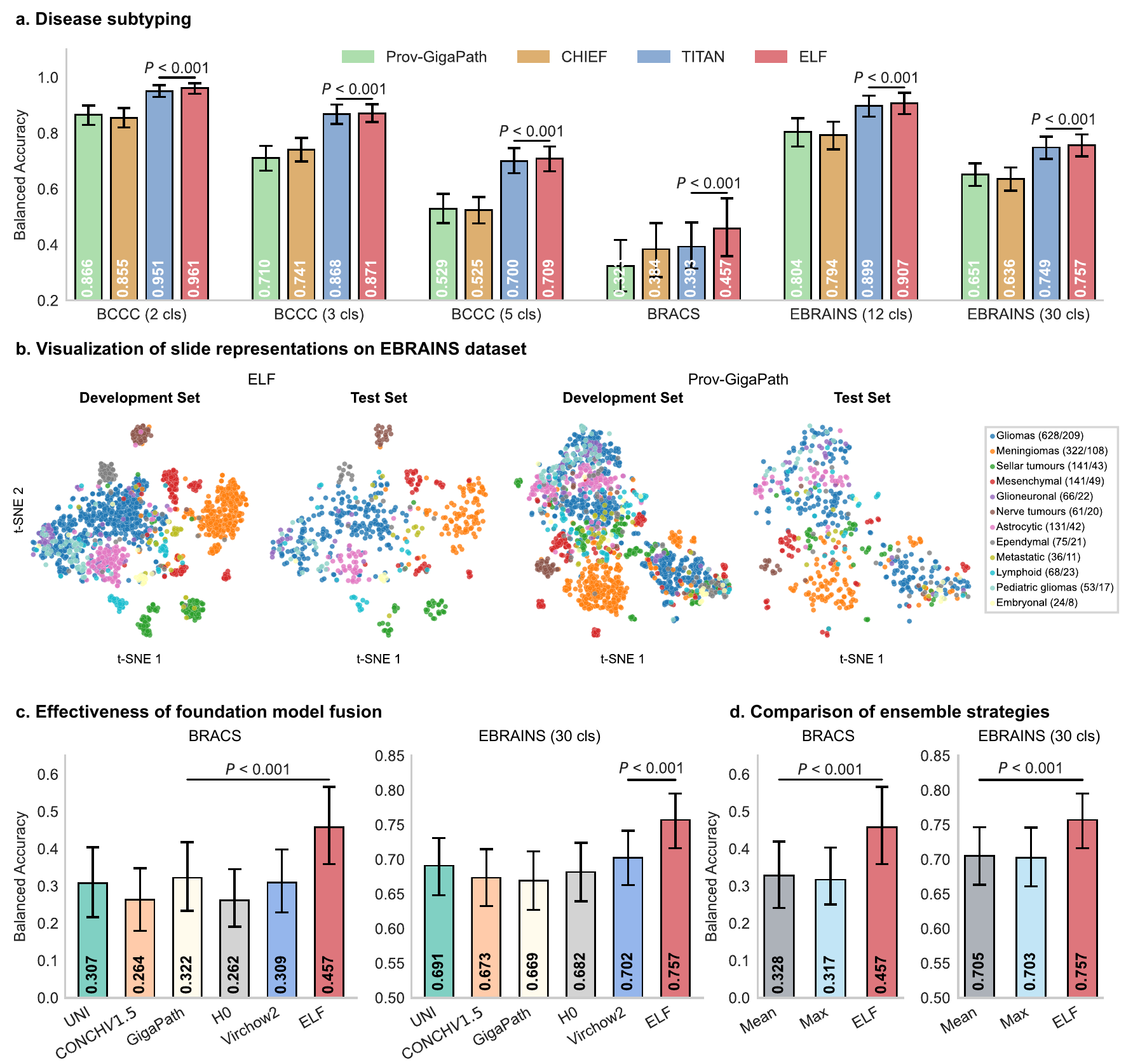}
\end{figure*}
\begin{figure*}[t]
\caption{\textbf{ELF identifies disease classes and tumor subtypes across benchmark datasets.}
\textbf{A,} Performance evaluation on six disease subtyping tasks using the held-out test sets of BCCC (n = 397), BRACS (n = 87), and EBRAINS (n = 573). ELF outperforms three state-of-the-art foundation models (Prov-GigaPath, CHIEF, TITAN) in the metric of balanced accuracy.
\textbf{B,} Visualization of slide-level feature representations. t-SNE maps show the slide-level feature representations of ELF and Prov-GigaPath on the EBRAINS dataset (12 cls) in the development (n = 1,746) and test (n = 573) sets.
\textbf{C,} ELF outperforms individual foundation models for disease subtyping on the BRACS (n = 87 WSIs) and EBRAINS (n = 573) test sets.
\textbf{D,} The ELF ensemble learning strategy outperforms alternative aggregation methods on the BRACS (n = 87) and EBRAINS (n = 573) test sets, including the distillation-based GPFM and the combination method COBRAII using model-agnostic attention to ensemble five foundation models. Statistical significance is shown for ELF compared with the next-best method.
In A, C and D, all bar plots show mean balanced accuracy with 95\% confidence intervals estimated by bootstrap resampling (1,000 replicates). The \(p\)-values were derived from a paired bootstrap procedure, with Bonferroni correction for the three prespecified pairwise comparisons.}
\label{fig:figure2}
\end{figure*}

\subsection{Molecular biomarker detection}\label{molecular-biomarker-detection}

Recent advances in computational pathology have demonstrated the feasibility of predicting therapeutically relevant biomarkers directly from routine H\&E-stained slides, which may accelerate treatment decisions\cite{prelaj2024artificial}. To systematically evaluate ELF's capability in this context, we conducted a large-scale study using the TCGA pan-cancer cohort, which includes 28 clinically actionable genetic alterations spanning 14 cancer types and approximately 6,500 patients\cite{weinstein2013cancer}. On the TCGA, ELF achieved a mean area under the receiver operating characteristic curve (AUC) of 0.719 \(\pm\) 0.142 across 84 biomarker-indication combinations, significantly surpassing prior foundation models, including TITAN (0.701 \(\pm\) 0.154), CHIEF (0.676~\(\pm\)~0.162), and Prov-GigaPath (0.678 \(\pm\) 0.149) (\(P < 0.05\); Figure~3\textbf{A}). A heatmap visualization further illustrates ELF's biomarker prediction performance for individual tasks, with a high accuracy (AUC \(>\) 0.85) for clinically actionable biomarkers, such as \emph{BRAF} mutation in thyroid and colorectal cancers, \emph{KRAS} and \emph{ROS1} mutations in lung cancer, \emph{IDH1} and \emph{FGFR3} mutations in gliomas (Figure~3\textbf{B} and Figure S5).

Identifying MSI as well as \emph{BRAF and KRAS mutation status} represents a fundamental part of the diagnostic workup and treatment of metastatic colorectal cancer (CRC). To further evaluate generalizability across diverse patient populations in this clinically relevant domain, we conducted external validation of ELF for predicting \emph{BRAF}, \emph{KRAS} mutations, and MSI status using four CRC cohorts: SR386\cite{myles2025surgen}, SR1482\cite{myles2025surgen}, MCO\cite{ward2015mco,jonnagaddala2016integration}, and TCGA-CRC\cite{weinstein2013cancer}, comprising 2,711 unique patients after accounting for overlap across cohorts. Among them, SR386, a recently released real-world dataset, was used for internal cross-validation, while SR1482, MCO, and TCGA-CRC served as external validation cohorts. For \emph{BRAF} mutation prediction, ELF consistently outperformed competing models across four cohorts with an average AUC of 0.806 \(\pm\) 0.021. Specifically, the performance on three external cohorts was 0.777 \(\pm\) 0.033 (SR1482), 0.842 \(\pm\) 0.014 (MCO), and 0.763 \(\pm\) 0.031 (TCGA-CRC), respectively. These represented notable AUC improvements ranging up to 13\% compared to the existing model CHIEF\cite{wang2024pathology} (\emph{P}\(<\)0.01; Figure~3\textbf{C}, top row). For MSI status prediction, ELF achieved excellent performance with an average AUC of 0.886 \(\pm\) 0.015 across the four cohorts. Specifically, the results on three external cohorts were 0.869 \(\pm\) 0.042 (SR1482), 0.897 \(\pm\) 0.013 (MCO), 0.855 \(\pm\) 0.028 (TCGA-CRC), respectively, consistently exceeding the second-best model (CHIEF) up to 7\% (Figure~3\textbf{C}, bottom row). Similar patterns were observed for \emph{KRAS} mutation prediction (Table S4). To assess the clinical utility of these results, we computed the positive predictive value at high negative predictive value (i.e., rule-out) for biomarker detection (Table S4). In the largest MCO external validation dataset, ELF achieved positive predictive values (PPV) of 0.35 and 0.55 for \emph{BRAF} mutation and MSI detection at the negative predictive value (NPV) of 0.95, which is higher than alternative models.

In addition to mutation status prediction, we further evaluated ELF on a continuous regression task for predicting aneuploidy scores. Aneuploidy, when cells harbor abnormal numbers of chromosomes, is a near-universal characteristic of tumor cells with significant clinical implications and for which reliable biomarkers are lacking. Aneuploidy is also inversely correlated with the expression of immune-related genes and a marker of immune evasion and resistance to immunotherapy\cite{taylor2018genomic}. We evaluated this task using the pan-cancer TCGA dataset, comprising 8,819 patients across 32 cancer types, with aneuploidy scores obtained from a prior study\cite{taylor2018genomic}. The results showed that ELF achieved better performance than three existing methods across three regression evaluation metrics (Figure~3\textbf{D}, first three panels). Specifically, ELF achieved the highest Pearson correlation (0.645 \(\pm\) 0.014), the lowest mean squared error (40.308 \(\pm\) 1.687), and the greatest coefficient of determination (\(R^{2}\) = 0.414 \(\pm\) 0.018).

We also evaluated ELF for the detection of whole-genome doubling\cite{taylor2018genomic}, an early oncogenic event with prognostic and therapeutic implications as well as a driver of chromosomal instability\cite{quinton2021whole,lambuta2023whole} (Figure~3\textbf{D}, right panels). ELF achieved an AUC of 0.845 \(\pm\) 0.010 (all \emph{P}\(<\)0.001), compared with 0.822 \(\pm\) 0.010 (TITAN), 0.806 \(\pm\) 0.011 (CHIEF) and 0.815 \(\pm\) 0.010 (Prov-GigaPath). Ablation studies on the largest pan-cancer regression and classification tasks confirmed the effectiveness of integrating diverse foundation models and using the ensemble strategy (Figure S6). Finally, we evaluated the performance of ELF for the prediction of tumor mutational burden (TMB), a clinically actionable biomarker with predictive and therapeutic implications as well as an FDA-recognized determinant of immunotherapy response\cite{yarchoan2017tumor,marcus2021fda} (Figure S7). The results showed that ELF predicted TMB status accurately with an AUC \(>\) 0.86. Moreover, ELF achieved the highest PPV (0.266) among the evaluated models at NPV\(>\)0.95, highlighting its potential clinical utility. These results demonstrated ELF's enhanced capability to capture chromosomal instability--related histomorphological patterns.

\begin{figure*}[p]
\centering
\includegraphics[width=\textwidth,height=1.0\textheight,keepaspectratio]{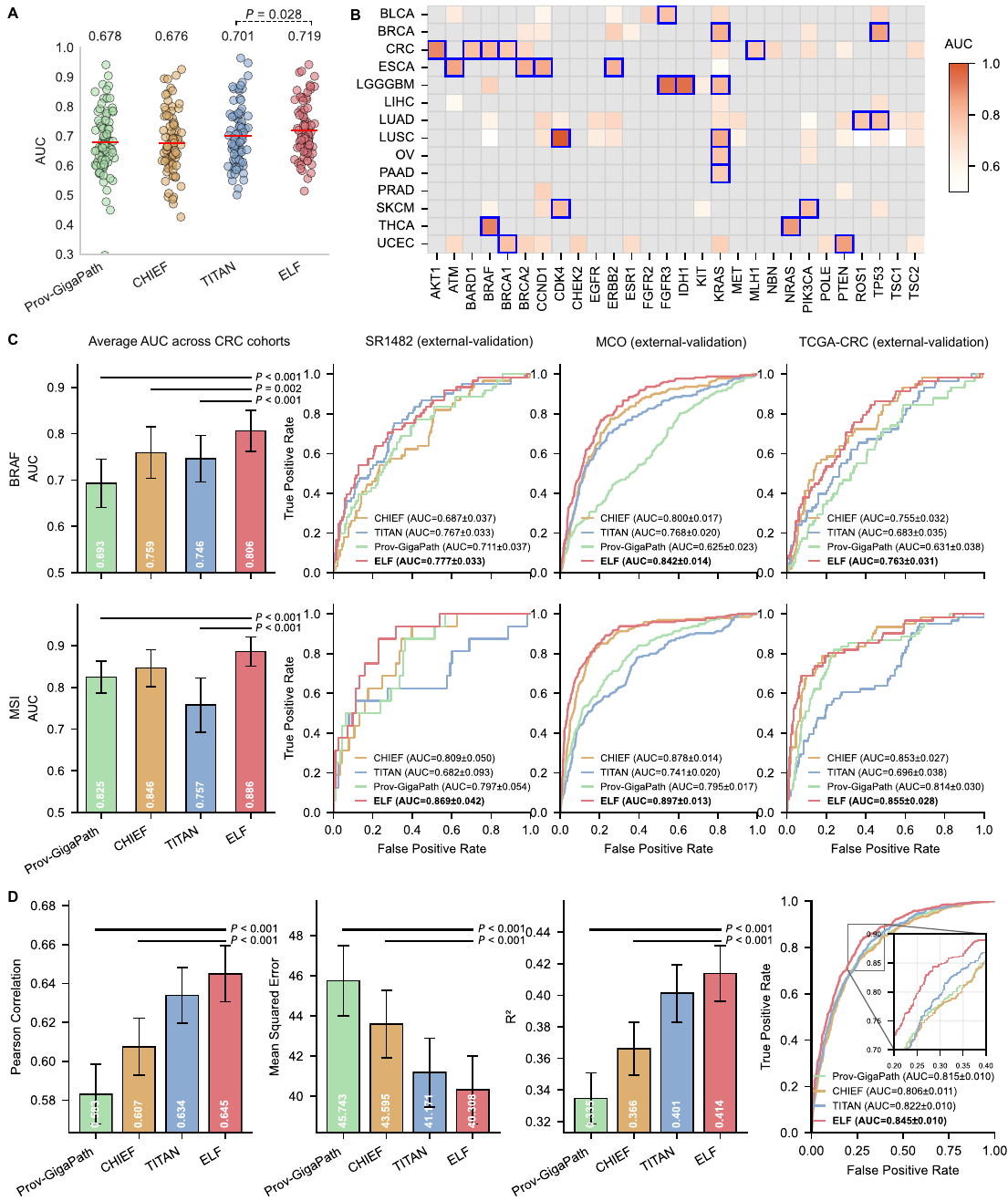}
\end{figure*}
\begin{figure*}[t]
\caption{\textbf{ELF predicts molecular biomarkers across cancer types and patient cohorts.}
\textbf{A,} Overall performance comparison of biomarker detection across 84 biomarker-indication combinations in the TCGA dataset, including 14 cancer types and 28 biomarkers (n = 6,510 patients represented by 7,721 WSIs). ELF significantly outperforms Prov-GigaPath, CHIEF, and TITAN in AUC (\emph{P} \(<\) 0.05). The \(p\)-values above the overall performance bars were derived from two-sided paired Wilcoxon signed-rank tests across the 84 biomarker--cancer combinations, with Bonferroni correction for the three prespecified pairwise comparisons.
\textbf{B,} Heatmap showing ELF's AUC performance for individual biomarker prediction tasks in the TCGA dataset. Blue-outlined boxes highlight biomarkers with AUC \(>\) 0.75 and at least 5 positive samples.
\textbf{C,} ELF achieves higher performance for detecting BRAF mutation and MSI status across one internal and three external colorectal cancer (CRC) cohorts. For BRAF, sample sizes were n = 400, 338, 1,488, and 501 patients in SR386, SR1482, MCO, and TCGA-CRC, respectively; for MSI, the corresponding sample sizes were n = 400, 129, 1,488, and 429 patients. Left bar plots show average performance across cohorts; ROC curves display external validation results. Bar plots show mean AUC with standard deviation across the four CRC cohorts. \(p\)-values were obtained by two-sided DeLong tests for paired ROC curves within each external cohort (ELF vs baseline), then combined across cohorts using inverse-variance fixed-effect meta-analysis, with Bonferroni correction for the three prespecified pairwise comparisons.
\textbf{D,} Performance comparison on the aneuploidy score regression (the left three panels) and whole-genome doubling classification (the right panel) tasks in the held-out test set (n = 1,767). ELF surpasses all baseline models across Pearson correlation, mean squared error (MSE), \(R^{2}\), and AUC (based on whole-genome doubling positive \emph{vs} negative\cite{taylor2018genomic}). Statistical significance for Pearson correlation, MSE, and coefficient of determination was assessed using paired bootstrap tests, whereas AUC comparisons were assessed using two-sided DeLong tests, with Bonferroni correction for the three prespecified pairwise comparisons.
In C and D, error bars represent the standard deviation derived from separate univariate patient-level bootstrap resampling for each model (1,000 replicates).}
\label{fig:figure3}
\end{figure*}

\subsection{Anticancer therapy response prediction}\label{anticancer-therapy-response-prediction}

Accurate prediction of therapy response is crucial for personalized treatment but remains an unmet need due to inherent complexity and challenges\cite{passaro2024cancer,hoang2024deep}. To address this, we investigated ELF's ability to predict patient response across multiple anticancer therapy settings, including both chemotherapy-based and targeted treatments, spanning two cancer types and eight cohorts with a total of 679 patients. These datasets represent diverse treatment regimens and contexts, including platinum-based chemotherapy and trastuzumab- and bevacizumab-based targeted therapy in neoadjuvant, adjuvant, and advanced settings.

For chemotherapy response prediction, ELF demonstrated strong performance in metastatic breast cancer (Metastatic breast (Platinum), \(n = 77\))\cite{bergstrom2024deep} and high-grade serous ovarian cancer (Ovary (Platinum), \(n = 158\))\cite{chowdhury2023proteogenomic}, achieving AUCs of 0.710 \(\pm\) 0.032 and 0.761 \(\pm\) 0.018, respectively (Figure~4\textbf{A}). For targeted therapy, ELF achieved similarly high performance in Breast (Trastuzumab, Yale)\cite{farahmand2022her2} cohort treated with trastuzumab (AUC = 0.749 \(\pm\) 0.040, \(n = 85\)) and Ovary (Bevacizumab)\cite{wang2022histopathological} patients treated with bevacizumab (AUC = 0.817 \(\pm\) 0.076, \(n = 36\)). Across all four cohorts, ELF achieved an overall mean AUC of 0.759, outperforming two existing slide-level foundation models by 7\%--9\% (Prov-GigaPath and CHIEF, Figure~4\textbf{A}).

To assess whether ELF predictions reflect clinically meaningful outcomes, we performed Kaplan--Meier survival analyses using the cohort-specific time-to-event endpoints. ELF risk scores successfully stratified patients into high-risk and low-risk groups, with significant hazard ratios: Metastatic breast (Platinum) in progression-free survival (PFS; HR = 1.64; 95\% CI, 1.02--2.63; P = 0.04) and Ovary (Bevacizumab) in recurrence-free survival (RFS; HR = 3.70; 95\% CI, 1.31--10.45; \emph{P} = 0.0095), further supporting the clinical relevance of ELF predictions (Figure~4\textbf{B}).

To evaluate model generalizability, we conducted cross-institutional validation using two additional breast cancer studies (TransNEO\cite{sammut2022multi} and IMPRESS\cite{huang2023artificial}). In the external validation cohort (IMPRESS, \(n = 64\)), ELF achieved an AUC of 0.668 (95\% CI: 0.515-0.806) for predicting pathologic complete response (pCR) to neoadjuvant chemotherapy, surpassing previous foundation models by 3--12\% (Figure~4\textbf{C}). In another breast cohort, ELF reached an AUC of 0.662 (0.530--0.790) for predicting pCR to neoadjuvant trastuzumab in the external validation cohort (IMPRESS, \(n = 62\)), again outperforming baseline models by 8\%--16\% (Figure~4\textbf{C}). Combining H\&E-based ELF-predicted score with CD8\cite{huang2023artificial} improved the AUC to 0.743 and 0.759 for predicting response to neoadjuvant chemotherapy and trastuzumab, respectively. Additionally, the ablation study investigated the performance of integrating diverse foundation models and the ensemble strategy on two external cohorts, also demonstrating the effectiveness of the ensemble strategy (Figure S8).

To aid interpretation of the model, we generated attention heatmaps overlaid onto WSIs for two representative HER2-positive cases with pathological complete response (pCR) and non-pCR outcomes, respectively (Figure~4\textbf{D}; a high-resolution version is provided in Figure S9). In the non-pCR case, pathologist-delineated regions of interest (ROIs) within high-attention areas corresponded to fibrotic stromal reaction surrounding the tumor, sparse peritumoral lymphocytic infiltrate, and evidence of lobular differentiation. In contrast, the pCR case demonstrated markedly distinct histological features within high-attention regions, characterized by abundant tumor-infiltrating lymphocytes (TILs), high-grade nuclear pleomorphism with marked cytological atypia, and the presence of tumor necrosis. We further examined the top-ranked attention patches after excluding border patches and found that the high-attention regions reflected the case-specific histological features described above, consistent with the heatmap-level observations. Additional representative examples and morphological interpretations for TNBC and ovarian cancer patients treated with Bevacizumab are presented in Figure S10 and Figure S11, respectively. Collectively, these observations indicated that the model's attention aligned with histological features relevant to treatment response across the cancer types examined.

\begin{figure*}[p]
\centering
\includegraphics[width=\textwidth,height=1.0\textheight,keepaspectratio]{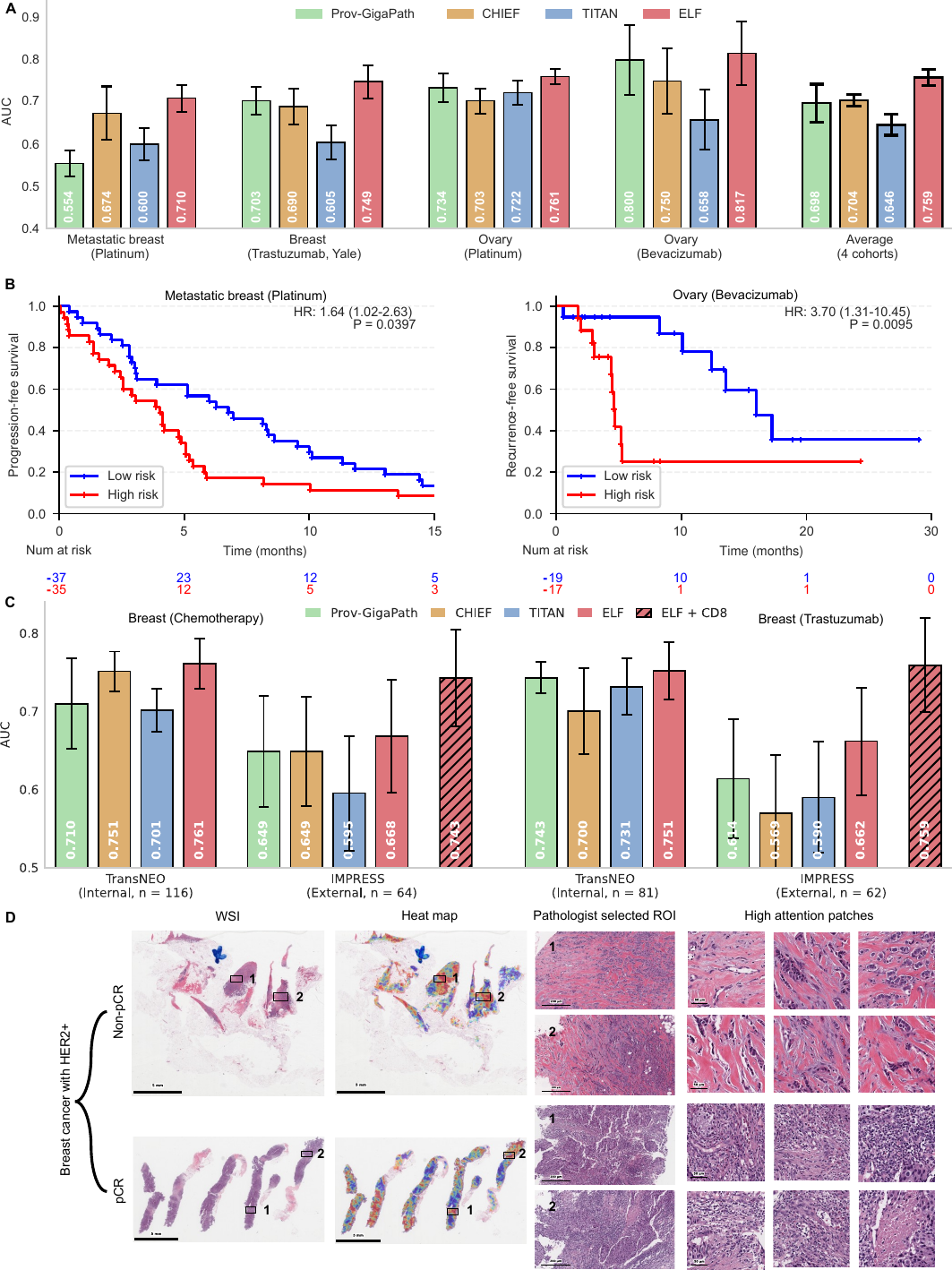}
\end{figure*}
\begin{figure*}[t]
\caption{\textbf{ELF predicts anticancer therapy response and stratifies progression-free and recurrence-free survival.}
\textbf{A,} Performance comparison of ELF and baseline models (Prov-GigaPath, CHIEF, TITAN) across four cohorts in predicting therapy response: Metastatic breast (Platinum) (n = 77), Ovary (Platinum) (n = 158), Breast (Trastuzumab, Yale) (n = 85), and Ovary (Bevacizumab) (n = 36). These comprise chemotherapy cohorts (Metastatic breast (Platinum) and Ovary (Platinum)) and molecularly targeted therapy cohorts (Breast (Trastuzumab, Yale) and Ovary (Bevacizumab)). ELF consistently achieves the highest AUC across all cohorts.
\textbf{B,} Kaplan--Meier analyses based on ELF-derived risk scores stratify progression-free survival (PFS) in the Metastatic breast (Platinum) cohort (n = 72) and recurrence-free survival (RFS) in the Ovary (Bevacizumab) cohort (n = 36). A two-sided log-rank test was used to compare survival outcomes between the high-risk and low-risk groups.
\textbf{C,} Generalizability of ELF in neoadjuvant chemotherapy and trastuzumab response prediction using internal TransNEO cohorts (chemotherapy, n = 116; trastuzumab, n = 81) and external IMPRESS cohorts (chemotherapy, n = 64; trastuzumab, n = 62). A combination of ELF-predicted risk score with CD8 further improves response prediction accuracy.
\textbf{D,} Representative H\&E-stained whole-slide images from two HER2-positive breast cancer cases with non-pathological complete response (non-pCR; top) and pathological complete response (pCR; bottom). For each case, the original H\&E-stained WSI, the corresponding attention heatmap overlaid on the WSI, pathologist-selected regions of interest (ROIs) from high-attention areas, and top-ranked high-attention patches are shown from left to right. Scale bars, 5 mm for the WSIs and attention heatmaps, 200 \(\mu\text{m}\) for the pathologist-selected ROIs, and 50 \(\mu\text{m}\) for the high-attention patches.
In \textbf{A} and \textbf{C}, bar plots show mean AUC with error bars representing standard error across five folds for cross-validation cohorts and standard deviation estimated from 1,000 bootstrap resamples for external validation cohorts.}
\label{fig:figure4}
\end{figure*}

\subsection{Immunotherapy response prediction}\label{immunotherapy-response-prediction}

Immunotherapy with immune checkpoint inhibitors (ICIs) has reshaped cancer treatment and is a standard of care in most tumor types. However, clinical response remains heterogeneous across patients, and reliable predictive biomarkers are lacking\cite{cristescu2018pan,samstein2019tumor}. To address this, we evaluated ELF's ability to predict durable clinical response to ICIs using pretreatment H\&E-stained images in a real-world dataset consisting of 13 independent cohorts of 1,057 patients across eight cancer types. Durable response was defined as PFS \(\geq\) 6 months for advanced-disease cohorts and PFS \(\geq\) 24 months for non-advanced-disease cohorts\cite{nabet2020noninvasive,louie2024molecular}. For the breast neoadjuvant cohort, response was instead defined as pathological complete response (pCR) at surgery. Among these, 11 cohorts were derived from patients treated at Stanford Medicine, one cohort from the Cancer Moonshot Project, and one additional cohort from Memorial Sloan Kettering Cancer Center (MSKCC).

First, we analyzed all 13 cohorts spanning eight cancer types, including bladder cancer (localized, \(n = 47\) and advanced, \(n = 46\)), breast cancer (neoadjuvant, \(n = 32\) and advanced, \(n = 33\)), endometrial cancer (\(n = 96\)), gastroesophageal adenocarcinoma (\(n = 106\)), head and neck cancer (\(n = 47\)), melanoma (non-metastatic, \(n = 63\) and metastatic, \(n = 78\)), non-small cell lung cancer (NSCLC; Stanford, \(n = 148\), Cancer Moonshot Project, \(n = 34\), and MSKCC, \(n = 275\)), and renal cell carcinoma (\(n = 52\)). Across all eight cancer types, ELF achieved a mean AUC of 0.724, with a higher mean AUC than the evaluated foundation models (Prov-GigaPath, CHIEF, and TITAN) with mean AUCs ranging from 0.612 to 0.664 (\emph{P} \(<\) 0.05; Figure~5\textbf{A} and Table S5).

We further investigated the predictive performance and potential clinical utility of ELF using two large NSCLC cohorts: an internal validation cohort, Stanford and an independent external validation cohort, MSKCC. In a subset of the Stanford cohort with PD-L1 expression data available, combining ELF's prediction with PD-L1 and TMB further improved predictive performance to AUC = 0.773, compared to ELF (AUC = 0.732), PD-L1 (AUC = 0.683), and TMB (AUC = 0.594) alone. In the independent MSKCC external validation cohort, ELF (AUC = 0.640) also outperformed PD-L1 (AUC = 0.603) and TMB (AUC = 0.506), and the composite model further improved predictive performance to AUC = 0.722 (DeLong test \(P <\) 0.01). In multivariable analysis adjusting for age, sex, smoking history, TMB, and PD-L1 status, the ELF risk score remained the strongest independent predictor of durable response in both Stanford and MSKCC cohorts (\(P <\) 0.01; Figures~5\textbf{B} and 5\textbf{C}). Moreover, the ablation study investigated the performance of integrations of diverse foundation models and the ensemble strategy on lung and breast external cohorts, where these performances confirmed the effectiveness of the ensemble strategy (Figure S12\textbf{).}

To assess clinical utility at a meaningful operating point, we applied a locked threshold of 0.441, selected in the Stanford internal cohort by minimizing the two-sided log-rank P value. At this threshold, the composite model achieved a sensitivity of 0.86, specificity of 0.44, PPV of 0.48, and NPV of 0.84 in the MSKCC validation cohort (Figure~5\textbf{D} and Table S6). Compared with PD-L1 alone, the composite model showed higher sensitivity, PPV, and NPV, although specificity was lower. We also performed decision curve analysis, which confirmed that the locked threshold falls within the net benefit region (0.20--0.75, Figure S13).

We further evaluated ELF's ability to stratify patients by PFS using the same locked threshold. In the Stanford cohort, ELF significantly separated high- and low-risk groups (HR = 1.84; 95\% CI: 1.27--2.67; \(P = 0.0011\)), with median PFS of 4.6 versus 15.1 months, respectively. This stratification was independently validated in the MSKCC cohort, where the same locked threshold significantly separated the high- and low-risk groups (HR = 1.98; 95\% CI: 1.47--2.67; \(P < 0.0001\)), with median PFS of 2.7 versus 8.6 months, respectively (Figure S14). Furthermore, ELF maintained consistent stratification performance across clinically relevant subgroups in the external validation cohort (MSKCC), including lines of immunotherapy and ICI regimen type (monotherapy \emph{vs.} chemo--ICI combination, Figure S15).

Furthermore, in the strictly matched subset of MSKCC cohort with available PD-L1 data, compared with PD-L1 alone, the ELF-based composite model increased the durable response rate among predicted responders from 53\% to 59\%, while simultaneously reducing the predicted non-responder proportion by half (32\% to 16\%) (Figure 5\textbf{E}). Beyond binary stratification, three-tier ELF risk groupings demonstrated significant stepwise PFS separation in the full MSKCC cohort with available PFS data (\(P < 0.0001\)), whereas PD-L1 TPS groupings were evaluated among patients with available PD-L1 data (Figure 5\textbf{F,} \emph{P} \(= 0.0232\)). Consistent stratification of PFS was observed across a broad range of cancer types beyond NSCLC (Figure 5\textbf{G} and Figure S14).

To provide interpretability for the ELF model predictions, we performed attention-based visualization to identify the important regions for predicting immunotherapy response. Specifically, we overlaid ELF's attention maps on representative images and compared morphological patterns between responders and non-responders (Figures S16, S17 and S18). In responder cases from breast cancer cohorts, the model's high-attention regions correspond to syncytial tumour cell nests with marked nuclear pleomorphism and dense intra- and peri-tumoural lymphoid infiltrates. By contrast, in non-responders, the high-attention areas were dominated by densely collagenized stroma with scant lymphocytic presence. Similarly, in the head and neck cohort, high-attention regions in responders highlighted brisk T cell aggregates intermingled with pleomorphic tumour cells, whereas non-responder heat maps focused on desmoplastic stroma devoid of inflammatory cells. Quantitative analysis of high-attention patches in the MSKCC validation cohort further corroborated this pattern: patches related to the tumor immune microenvironment consistently outnumbered tumor patches across all attention thresholds (\(\sim\)25--30\% vs.~\(\sim\)13--16\%) (Figure S19). This indicated that the ELF model's attention was primarily focused on immune-rich and fibrotic regions that have been previously implicated in ICI efficacy\cite{bagaev2021conserved}.

Together, these results showed that ELF-based risk stratification provided information complementary to PD-L1 expression across the evaluated cohorts, supporting further evaluation for ICI response stratification.

\begin{figure*}[p]
\centering
\includegraphics[width=\textwidth,height=1.0\textheight,keepaspectratio]{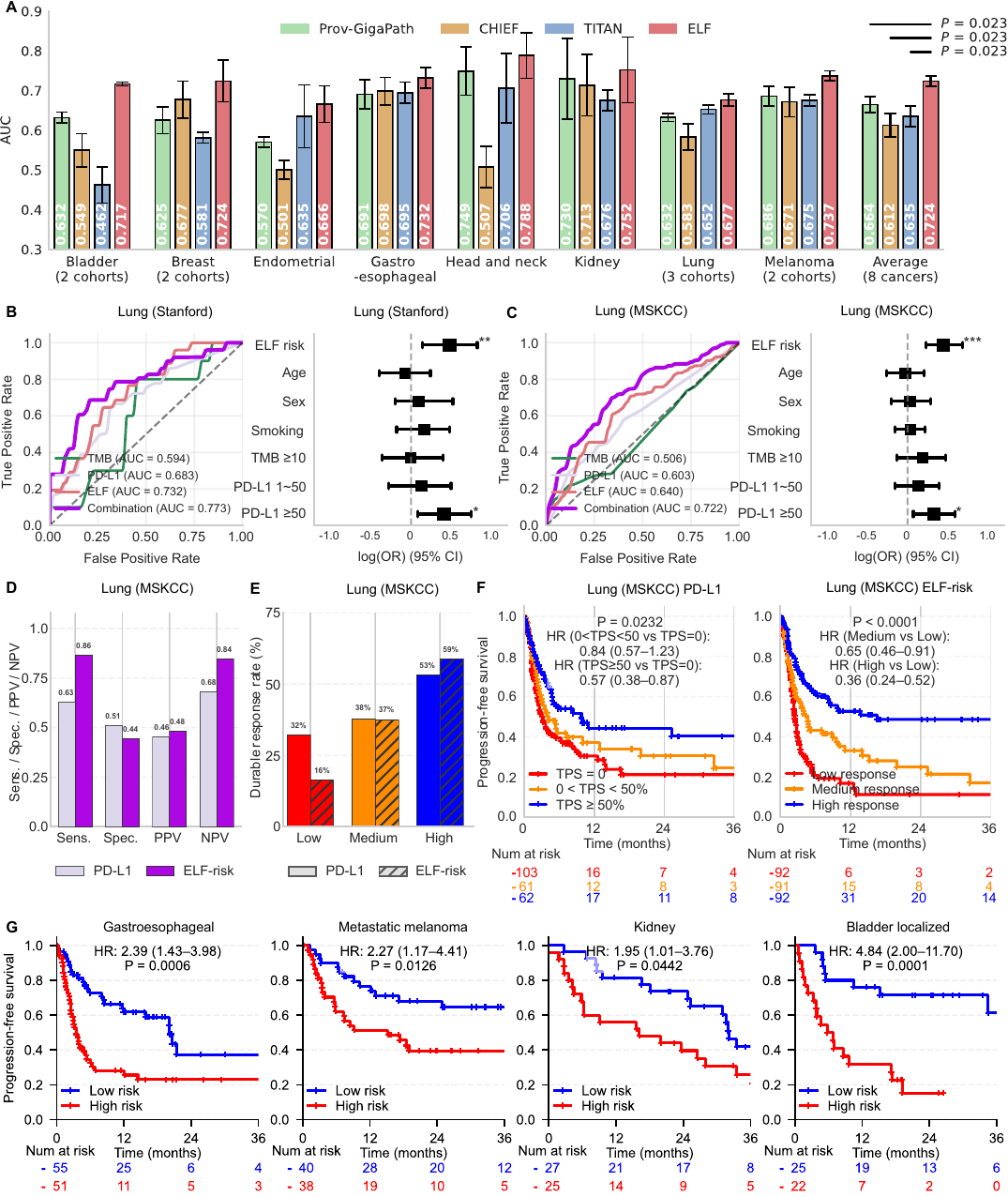}
\end{figure*}
\begin{figure*}[t]
\caption{\textbf{ELF predicts immunotherapy response and stratifies progression-free survival across cancers.}
\textbf{A,} Evaluation of durable response prediction to immune checkpoint inhibitors (ICIs) across eight cancer types and 13 cohorts (n = 1,057): advanced bladder (n = 46), localized bladder (n = 47), kidney (n = 52), non-metastatic melanoma (n = 63), metastatic melanoma (n = 78), head and neck (n = 47), gastroesophageal (n = 106), endometrial (n = 96), Stanford lung (n = 148), CMB lung (n = 34), MSKCC lung (n = 275), neoadjuvant breast (n = 32), and advanced breast (n = 33). Bar plots show mean AUC with error bars representing standard deviation estimated by bootstrap resampling (1,000 replicates) for external validation cohorts and mean with standard error across five folds for cross-validation cohorts. The \(p\)-values above the Average (8 cancers) bars were derived from two-sided paired Wilcoxon signed-rank tests across eight matched cancer-type-level AUC values with Bonferroni correction for three prespecified pairwise comparisons. Since ELF outperforms each comparator in all 8 cancer types, every pairwise comparison attained the minimum \(p\)-value achievable under this design (raw \(p = 2/2^{8} = 0.0078\), Bonferroni-corrected to \(0.023\)); the identical adjusted \(p = 0.023\) thus represents the strongest, not a moderate, result the test can yield.
\textbf{B,} ROC curves for durable response prediction and multivariable logistic regression forest plots for patients with available PD-L1 status in the Stanford NSCLC cohort (n = 127), adjusting for age, sex, smoking history, TMB, and PD-L1 status. Squares indicate odds ratios, horizontal error bars indicate 95\% confidence intervals, and the vertical dashed line indicates an odds ratio of 1. P values were obtained using two-sided Wald tests for the regression coefficients (\(***\ P < 0.001\), \(**\ P < 0.01\), \(*\ P < 0.05\)).
\textbf{C,} ROC curves for durable response prediction and multivariable logistic regression forest plots for patients with available PD-L1 data in the MSKCC NSCLC cohort (n = 226), with the same covariates, graphical conventions, and statistical analysis as in B.
\textbf{D,} Sensitivity, specificity, PPV, and NPV of the ELF-based combination model and PD-L1 TPS for durable response prediction in the strictly matched subset of MSKCC NSCLC patients with available PD-L1 data (n = 226).
\textbf{E,} Durable response rates across ELF-based combination risk groups and PD-L1 TPS subgroups in the same strictly matched subset of MSKCC NSCLC patients with available PD-L1 data (n = 226).
\textbf{F,} Kaplan--Meier analyses of progression-free survival (PFS) according to PD-L1 TPS groupings among MSKCC NSCLC patients with available PD-L1 data (left; n = 226) and according to ELF-based combination risk groups in the full MSKCC cohort with available PFS data (right; n = 275); survival differences were assessed using two-sided log-rank tests.
\textbf{G,} Kaplan--Meier analyses of PFS stratified by ELF-derived binary risk scores (low vs.~high) in gastroesophageal (n = 106), metastatic melanoma (n = 78), kidney (n = 52), and localized bladder cancer (n = 47) cohorts; survival differences were assessed by a two-sided log-rank test.}
\label{fig:figure5}
\end{figure*}

\section{DISCUSSION}\label{discussion-1}

In this study, we introduced an ensemble approach for developing pathology foundation models, ELF, that harnesses the complementary representational strengths of independently trained foundation models. By employing an ensemble learning approach, ELF integrated five state-of-the-art foundation models through a unified slide-level encoder, enabling the generation of robust and generalizable features. Evaluated across 125 diverse tasks, including disease subtyping, biomarker detection, and response prediction for both targeted therapies and immunotherapies, ELF outperformed individual foundation models as well as existing slide-level approaches. These findings support ensemble-based integration as an approach for combining complementary pathology foundation-model representations.

Most existing pathology foundation models are trained as tile-level models. For downstream applications, separate models must be trained to combine features from all the tiles in a slide. And since each WSI may contain thousands of tiles, this requires a large labeled dataset for training purposes. By contrast, ELF is a pretrained slide-level foundation model that directly encodes the entire WSI into compact representations. This design reduces the amount of task-specific training data required relative to traditional tile-level approaches and may therefore be useful in data-limited settings such as therapeutic response prediction.

Several slide-level foundation models have been developed\cite{xu2024whole,wang2024pathology,ding2024multimodal}. However, they are all tightly coupled to one specific tile-level model\cite{xu2024whole,chen2024towards,wang2022transformer}. Compared with these existing slide-level models, ELF improved the AUC by 8-18\% for predicting anti-cancer therapy response; and by 10-17\% for predicting immunotherapy response across all evaluation datasets. This improvement is attributable to the ensemble strategy of ELF that leverages the complementary strengths of the diverse set of pretrained foundation models that exist today without the need to train new tile-level foundation models from scratch and also reducing environmental impact. It is straightforward to further extend ELF to incorporate additional foundation models, although this will come at increased computational cost.

Although recent studies have attempted to combine multiple foundation models, the optimal approach remains unclear. One study GPFM\cite{ma2024towards} employed model distillation to learn from the knowledge of several expert models for a new tile-level foundation model. By smoothing out the differences across models, this approach does not fully exploit inter-model complementarity and did not outperform existing foundation models in benchmark studies\cite{ma2025pathbench}. Direct comparison between ELF and GPFM further showed that ensembling multiple foundation models yields greater benefit than distilling them into a single one. In another study, COBRAII\cite{lenz2025unsupervised} learns attention weights by fusing features from multiple models and then applies the same attention weights to tile-level features. While COBRAII could in principle be extended to support ensembling of multiple foundation models, this capability was not systematically validated in the original study. By contrast, ELF jointly optimizes these weights under the supervision of all five models during pretraining, producing a more coherent and ensemble-aware slide encoder. This design allows ELF to preserve model-specific representations and integrate them synergistically. Empirical analyses revealed low correlation among the image features and attention maps generated from different base models. By aligning and combining these diverse features, ELF constructs richer slide-level representations that enhance predictive performance across diverse clinical applications.

The ability to detect therapeutically relevant biomarkers using routine H\&E-stained slides has important implications for targeted therapies. When evaluated on detecting 28 clinically actionable genetic alterations across 14 cancer types, ELF achieved higher AUCs than the evaluated foundation models, exceeding 0.85 for specific gene mutations such as \emph{BRAF} in thyroid carcinoma and \emph{KRAS} in lung adenocarcinoma. The model maintained generalizability across diverse patient populations for detecting actionable biomarkers in CRC. In external validation, our model achieved PPV of 0.35 and 0.55 for \emph{BRAF} mutation and \emph{MSI} detection respectively, significantly higher than alternative models (0.14-0.28 for \emph{BRAF}; 0.19-0.46 for \emph{MSI}), at a high NPV of 0.95. Such a model could be used in a triage setting as a rule-out test to reduce unnecessary and costly molecular assays. Alternatively, the model may also be used as a prioritization tool (process model-positive cases first) rather than a hard rule-out. In addition to individual genes, ELF can also accurately detect whole-genome doubling, TMB and high aneuploidy, which are broadly shared tumor cell traits with significant prognostic implications and for which therapeutic targeting is currently under development\cite{taylor2018genomic,lambuta2023whole,yarchoan2017tumor}.

Beyond detecting known biomarkers, we further evaluated ELF for predicting therapeutic responses and outcomes directly from pre-treatment H\&E slides. Compared with existing pathology foundation models, ELF achieved higher predictive performance than the evaluated pathology foundation models across cancer types treated with a range of anticancer therapies including cytotoxic chemotherapy, molecularly targeted therapy, and immunotherapy. These results support further evaluation of ELF across tumor types and treatment modalities, particularly in settings with limited training data.

For patients with NSCLC, the ELF-based composite model significantly improved the accuracy of immunotherapy response prediction compared with standard biomarker PD-L1. The model increased the durable response rate among predicted responders while reducing the predicted non-responder proportion by half in an external validation cohort. These findings support further evaluation of ELF as a clinical decision-support approach for identifying responders and non-responders.

In conclusion, this study presents an ensemble strategy for building foundation models by integrating the complementary strengths of existing pretrained models. Our ensemble learning-based approach enables the creation of robust and generalizable representations without requiring access to proprietary training data. The resulting ensemble foundation model demonstrates strong performance in disease classification, biomarker detection, and treatment response prediction across a wide range of cancers. Together, these findings support further evaluation of ELF as a scalable framework for pathology applications in oncology.

\subsection{Limitations of the study}\label{limitations-of-the-study}

This study has several limitations. First, while ELF integrates five state-of-the-art tile-level foundation models, the choice of which models to include was guided by their availability and benchmark performance at the time of development; future foundation models may offer further complementary value. Second, H\&E histology-based predictions alone may be insufficient to guide clinical treatment decisions, and integration with other modalities (genomics, imaging, and clinical records) may further improve performance and warrants further investigation. Third, although ELF was evaluated across multiple cancer types and treatment modalities, many results are based on retrospective and single-institution cohorts, with some notable exceptions; prospective trials in diverse patient populations will be essential to validate these findings and support clinical translation.

\clearpage
\section{RESOURCE AVAILABILITY}\label{resource-availability}

\subsection{Lead contact}\label{lead-contact}

Requests for further information and resources should be directed to and will be fulfilled by the lead contact, Dr. Ruijiang Li (rli2@stanford.edu).

\subsection{Materials availability}\label{materials-availability}

This study did not generate new materials.

\subsection{Data and code availability}\label{data-and-code-availability}

\begin{itemize}
\item
  The histopathology images used for pretraining are publicly available: ACROBAT (\url{https://researchdata.se/en/catalogue/dataset/2022-190-1}), BCNB (\url{https://bupt-ai-cz.github.io/BCNB}), COBRA (\url{https://registry.opendata.aws/cobra}), DHMC (\url{https://bmirds.github.io/}), GTEx (\url{https://www.gtexportal.org/home}), IMP-CRS (\url{https://rdm.inesctec.pt/dataset/nis-2023-008}), IPD-Brain (https://figshare.com/articles/dataset/IPD\_Brain/27186087), NMI-Bladder (\url{https://figshare.com/projects/nmi-wsi-diagnosis/61973}), PAIP (\url{http://www.wisepaip.org/}), PANDA (\url{https://www.kaggle.com/c/prostate-cancer-grade-assessment/data}), and TCGA (\url{https://portal.gdc.cancer.gov}). Downstream evaluation datasets: BCCC (\url{https://datahub.aida.scilifelab.se/10.23698/aida/bccc}), BRACS (\url{https://www.bracs.icar.cnr.it/}), EBRAINS (\url{https://doi.org/10.25493/WQ48-ZGX}), MCO (\url{https://www.sredhconsortium.org/sredh-datasets/mco-study-whole-slide-image-dataset}), SurGen (\url{https://www.ebi.ac.uk/biostudies/studies/S-BIAD1285}), Metastatic breast/Platinum (\url{https://www.synapse.org/Synapse:syn59490671}), Breast/Trastuzumab/Yale (\url{https://www.cancerimagingarchive.net/collection/her2-tumor-rois}), Ovary/Platinum (\url{https://www.cancerimagingarchive.net/collection/ptrc-hgsoc}), Ovary/Bevacizumab (\url{https://www.cancerimagingarchive.net/collection/ovarian-bevacizumab-response}), TransNEO (\url{https://zenodo.org/record/6337925}), and IMPRESS (\url{https://tinyurl.com/IMPRESS-DATA}). Data for the Stanford and MSKCC immunotherapy cohorts are not publicly available due to patient privacy considerations. Additionally, restrictions imposed by the IRB and data use agreements with participating institutions limit redistribution of individual-level clinical data. De-identified data may be made available to qualified researchers upon reasonable request and subject to a data use agreement, pending approval from the corresponding institution\textquotesingle s ethics review board.
\item
  All original code is publicly available at \url{https://github.com/lilab-stanford/ELF}, including pretraining, inference, and evaluation code. The model weights can be downloaded from \url{https://drive.google.com/file/d/1t79jlus2ZfVe6RGS4bqX1oJhmypdxyGb/view?usp=sharing}.
\item
  Any additional information required to reanalyze the data reported in this paper is available from the lead contact upon request.
\end{itemize}

\section{ACKNOWLEDGMENTS}\label{acknowledgments}

This study was supported in part by the National Institutes of Health under research grants R01CA269559, R01CA285456, and R01CA290715 from the National Cancer Institute. We acknowledge the ACROBAT\cite{weitz2024acrobat}, BCNB\cite{xu2021predicting}, COBRA\cite{geijs2024detection}, DHMC\cite{zhu2021development,wei2019pathologist}, GTEx\cite{lonsdale2013genotype}, IMP-CRS\cite{neto2022imil4path,neto2024interpretable}, IPD-Brain\cite{chauhan2024ipd}, NMI-Bladder\cite{zhang2019pathologist}, PAIP\cite{kim2021paip}, PANDA\cite{bulten2022artificial}, TCGA\cite{weinstein2013cancer}, BCCC\cite{yacob2023weakly}, BRACS\cite{brancati2022bracs}, EBRAINS\cite{roetzer2022digital}, MCO\cite{ward2015mco,jonnagaddala2016integration}, SurGen\cite{myles2025surgen}, Metastatic breast (Platinum)\cite{bergstrom2024deep}, Breast (Trastuzumab, Yale)\cite{farahmand2022her2}, Ovary (Platinum)\cite{chowdhury2023proteogenomic}, Ovary (Bevacizumab)\cite{wang2022histopathological}, TransNEO\cite{sammut2022multi}, and IMPRESS\cite{huang2023artificial} consortia for making their datasets publicly available. We also thank the research team\cite{taylor2018genomic} for providing aneuploidy scores and WGD/TMB data, our colleagues at Stanford Medicine\cite{callahan2023stanford} and Memorial Sloan Kettering Cancer Center for contributions to the immunotherapy patient cohort, and the Marlowe\cite{kapfer2025marlowe} computational infrastructure at Stanford for GPU support.

\section{AUTHOR CONTRIBUTIONS}\label{author-contributions}

Conceptualization, R.L., X.L., and X.W.; methodology, X.L., X.W., J.X., Y.J., Z.L., and Y.C.; investigation, X.L., X.W., F.E., X.Z., S.Y., Y.L., R.Q., J.X., Y.J., Z.L., Y.C., C.B., T.K., F.M.O., K.Y., M.A., A.H., S.W., S.R., R.W., J.N., and M.D.; data acquisition, R.L., Y.L., X.Z., F.E., J. G., S. C., C.B., T.K., F.M.O., K.Y., M.A., A.H., S.W., S.R., R.W., J.N., A. S., M.D. and C. V.; writing--original draft, R.L., X.L., X.W., S.Y., X.Z., and R.Q.; writing--review \& editing, all authors; funding acquisition, R.L.; supervision, R.L.

\section{DECLARATION OF INTERESTS}\label{declaration-of-interests}

The authors declare no competing interests.

\section{DECLARATION OF GENERATIVE AI AND AI-ASSISTED TECHNOLOGIES IN THE WRITING PROCESS}\label{declaration-of-generative-ai-and-ai-assisted-technologies-in-the-writing-process}

During the preparation of this work, the authors used ChatGPT (OpenAI) and Claude (Anthropic) for coding assistance and to improve the language and readability of the manuscript. After using these tools, the authors carefully reviewed, verified, and edited the generated content and code as needed and take full responsibility for the content of the publication.

\section{SUPPLEMENTAL INFORMATION INDEX}\label{supplemental-information-index}

\begin{itemize}
\item
  Figure S1 - Figure S19, Table S1 - Table S17 and their legends in a PDF.
\end{itemize}

\section{STAR METHODS}\label{star-methods-1}

\subsection{Key resources table}\label{key-resources-table}

{\def\LTcaptype{none} 
\begin{longtable}[]{@{}
  >{\raggedright\arraybackslash}p{(\linewidth - 4\tabcolsep) * \real{0.2036}}
  >{\raggedright\arraybackslash}p{(\linewidth - 4\tabcolsep) * \real{0.1421}}
  >{\raggedright\arraybackslash}p{(\linewidth - 4\tabcolsep) * \real{0.6542}}@{}}
\toprule\noalign{}
\begin{minipage}[b]{\linewidth}\raggedright
\textbf{REAGENT or RESOURCE}
\end{minipage} & \begin{minipage}[b]{\linewidth}\raggedright
\textbf{SOURCE}
\end{minipage} & \begin{minipage}[b]{\linewidth}\raggedright
\textbf{IDENTIFIER}
\end{minipage} \\
\midrule\noalign{}
\endhead
\bottomrule\noalign{}
\endlastfoot
& & \\
ELF pretrained model weights & This paper & \url{https://drive.google.com/file/d/1t79jlus2ZfVe6RGS4bqX1oJhmypdxyGb} \\
ELF source code (pretraining, inference, evaluation) & This paper & \url{https://github.com/lilab-stanford/ELF} \\
\multicolumn{3}{@{}>{\raggedright\arraybackslash}p{(\linewidth - 4\tabcolsep) * \real{1.0000} + 4\tabcolsep}@{}}{%
\textbf{Software and algorithms}} \\
Python 3.10.12 & Python Software Foundation & \url{https://www.python.org} \\
PyTorch 2.6.0 (CUDA 12.4) & Meta AI & \url{https://pytorch.org} \\
Scikit-learn 1.6.1 & N/A & \url{https://scikit-learn.org} \\
XGBoost 3.0.2 & Chen and Guestrin\cite{chen2016xgboost} & \url{https://xgboost.readthedocs.io} \\
OpenSlide 4.3.1 & N/A & \url{https://openslide.org} \\
CLAM & Lu et al.\cite{lu2021data} & \url{https://github.com/mahmoodlab/CLAM} \\
UNI (tile-level foundation model) & Chen et al.\cite{chen2024towards} & \url{https://huggingface.co/MahmoodLab/UNI} \\
CONCHV1.5 (tile-level foundation model) & Lu et al.\cite{lu2024visual,ding2024multimodal} & \url{https://huggingface.co/MahmoodLab/conchv1_5} \\
GigaPath (tile-level foundation model) & Xu et al.\cite{xu2024whole} & \url{https://huggingface.co/prov-gigapath/prov-gigapath} \\
Virchow2 (tile-level foundation model) & Zimmermann et al.\cite{zimmermann2024virchow2} & \url{https://huggingface.co/paige-ai/Virchow2} \\
H-Optimus-0 (tile-level foundation model) & Bioptimus\cite{hoptimus0} & \url{https://huggingface.co/bioptimus/H-optimus-0} \\
CHIEF (slide-level foundation model) & Wang et al.\cite{wang2024pathology} & \url{https://github.com/hms-dbmi/CHIEF} \\
Prov-GigaPath (slide-level foundation model) & Xu et al.\cite{xu2024whole} & \url{https://huggingface.co/prov-gigapath/prov-gigapath} \\
TITAN (slide-level foundation model) & Ding et al.\cite{ding2024multimodal} & \url{https://huggingface.co/MahmoodLab/TITAN} \\
COBRAII (slide-level foundation model) & Lenz et al.\cite{lenz2025unsupervised} & \url{https://github.com/KatherLab/COBRA} \\
Scikit-survival 0.24.0 & N/A & \url{https://scikit-survival.readthedocs.io} \\
lifelines 0.30.0 & N/A & \url{https://lifelines.readthedocs.io} \\
Matplotlib 3.10.0 & N/A & \url{https://matplotlib.org} \\
\multicolumn{3}{@{}>{\raggedright\arraybackslash}p{(\linewidth - 4\tabcolsep) * \real{1.0000} + 4\tabcolsep}@{}}{%
\textbf{Other (key datasets)}} \\
ACROBAT (breast cancer) & Weitz et al.\cite{weitz2024acrobat} & \url{https://researchdata.se/en/catalogue/dataset/2022-190-1} \\
BCNB (breast cancer) & Xu et al.\cite{xu2021predicting} & \url{https://bupt-ai-cz.github.io/BCNB} \\
COBRA (skin cancer) & Geijs et al.\cite{geijs2024detection} & \url{https://registry.opendata.aws/cobra} \\
DHMC (renal/lung cancer) & Zhu et al.\cite{zhu2021development}; Wei et al.\cite{wei2019pathologist} & \url{https://bmirds.github.io/} \\
GTEx (normal tissues) & Lonsdale et al.\cite{lonsdale2013genotype} & \url{https://www.gtexportal.org/home} \\
IMP-CRS (CRC) & Neto et al.\cite{neto2022imil4path,neto2024interpretable} & \url{https://rdm.inesctec.pt/dataset/nis-2023-008} \\
IPD-Brain (brain tumor) & Chauhan et al.\cite{chauhan2024ipd} & \url{https://figshare.com/articles/dataset/IPD\_Brain/27186087} \\
NMI-Bladder (bladder cancer) & Zhang et al.\cite{zhang2019pathologist} & \url{https://figshare.com/projects/nmi-wsi-diagnosis/61973} \\
PAIP (liver cancer) & Kim et al.\cite{kim2021paip} & \url{http://www.wisepaip.org/} \\
PANDA (prostate cancer) & Bulten et al.\cite{bulten2022artificial} & \url{https://www.kaggle.com/c/prostate-cancer-grade-assessment/data} \\
TCGA pan-cancer dataset & Weinstein et al.\cite{weinstein2013cancer} & \url{https://portal.gdc.cancer.gov} \\
BCCC (skin cancer subtyping) & Yacob et al.\cite{yacob2023weakly} & \url{https://datahub.aida.scilifelab.se/10.23698/aida/bccc} \\
BRACS (breast cancer subtyping) & Brancati et al.\cite{brancati2022bracs} & \url{https://www.bracs.icar.cnr.it/} \\
EBRAINS (brain tumor subtyping) & Roetzer-Pejrimovsky et al.\cite{roetzer2022digital} & \url{https://doi.org/10.25493/WQ48-ZGX} \\
SurGen (CRC cohorts SR386/SR1482) & Myles et al.\cite{myles2025surgen} & \url{https://www.ebi.ac.uk/biostudies/studies/S-BIAD1285} \\
MCO (CRC cohort) & Ward et al.\cite{ward2015mco} & \url{https://www.sredhconsortium.org/sredh-datasets/mco-study-whole-slide-image-dataset} \\
Metastatic breast/Platinum & Bergstrom et al.\cite{bergstrom2024deep} & \url{https://www.synapse.org/Synapse:syn59490671} \\
Breast/Trastuzumab (Yale) & Farahmand et al.\cite{farahmand2022her2} & \url{https://www.cancerimagingarchive.net/collection/her2-tumor-rois} \\
Ovary/Platinum (HGSOC) & Chowdhury et al.\cite{chowdhury2023proteogenomic} & \url{https://www.cancerimagingarchive.net/collection/ptrc-hgsoc} \\
Ovary/Bevacizumab & Wang et al.\cite{wang2022histopathological} & \url{https://www.cancerimagingarchive.net/collection/ovarian-bevacizumab-response} \\
TransNEO & Sammut et al.\cite{sammut2022multi} & \url{https://zenodo.org/record/6337925} \\
IMPRESS & Huang et al.\cite{huang2023artificial} & \url{https://tinyurl.com/IMPRESS-DATA} \\
CMB lung immunotherapy cohort & Cancer Moonshot Biobank\cite{cancerMoonshot2022} & \url{https://doi.org/10.7937/3CX3-S132} \\
\end{longtable}
}

\subsection{Experimental model and study participant details}\label{experimental-model-and-study-participant-details}

Eligible cases were required to have an H\&E-stained WSI and the clinical endpoint or molecular label required for the corresponding analysis. For treatment-response analyses, an eligible pretreatment H\&E-stained WSI was required. Cases with missing primary endpoints or required molecular labels, as well as WSIs failing quality control, were excluded from the corresponding analysis. Publicly available, de-identified datasets were analyzed in accordance with their respective data-use terms and the ethical approvals reported in the original studies. The retrospective Stanford cohorts were approved by the Stanford University Institutional Review Board (protocol 67432), with a waiver of informed consent. The de-identified MSKCC cohort was provided by MSKCC collaborators under an MSKCC Institutional Review Board-approved research framework (protocol 18-013) with a waiver of informed consent and was analyzed under formal institutional collaboration and data-use arrangements. Detailed cohort-specific eligibility criteria, cancer types, treatments, endpoints, and data sources are provided in the corresponding cohort subsections of the Method details. Cohort composition and sample distributions are summarized in Table S7 for pretraining; Table S10 for disease classification and subtyping; Tables S11--S14 for biomarker, aneuploidy, WGD, and TMB analyses; Table S15 for anticancer therapy-response cohorts; Table S16 for immunotherapy-response cohorts; and Table S17 for the baseline clinical characteristics of the Stanford and MSKCC NSCLC cohorts.

\subsection{Method details}\label{method-details}

\subsubsection{Data curation for pretraining}\label{data-curation-for-pretraining}

For pretraining of the ELF foundation model, we collected 53,699 whole-slide images (WSIs) and their corresponding clinical information (such as tissue of origin and tissue type: cancer vs normal) from 11 publicly available datasets, including ACROBAT\cite{weitz2024acrobat}, BCNB\cite{xu2021predicting}, COBRA\cite{geijs2024detection}, DHMC\cite{zhu2021development,wei2019pathologist}, GTEx\cite{lonsdale2013genotype}, IMP-CRS\cite{neto2022imil4path,neto2024interpretable}, IPD-Brain\cite{chauhan2024ipd}, NMI-Bladder\cite{zhang2019pathologist}, PAIP\cite{kim2021paip}, PANDA\cite{bulten2022artificial}, TCGA\cite{weinstein2013cancer}. In the TCGA, we only used four cohorts: TCGA-THCA, TCGA-CESC, TCGA-ESCA, and TCGA-UCEC, since these anatomical sites are not represented in the other public datasets. All images of this training set are formalin-fixed paraffin-embedded tissues stained with haematoxylin and eosin (H\&E) and come from 20 anatomic sites including adrenal, bladder, brain, breast, cervical, colorectal, esophagus, heart, kidney, liver, lung, ovary, pancreas, prostate, skin, soft tissue, stomach, testis, thyroid, and uterus. Figure 1\textbf{B} presents an overview of the data distribution across these anatomic sites. Detailed data distribution is summarized in Table S7.

\subsubsection{Model design and pretraining}\label{model-design-and-pretraining}

To harness the complementary strengths of multiple pathology foundation models, we developed ELF, a unified slide-level ensemble framework. Instead of directly concatenating high-dimensional features from different models, which typically demands large-scale labelled data for training the multiple-instance classifier and may introduce training instability, ELF integrates information from multiple foundation models through a compact, learnable architecture. Specifically, it employs a multi--head attention--based multi-instance learning model as a patch aggregator\cite{ilse2018attention}, trained using cross-model and cross-view contrastive learning together with auxiliary weak supervision. This enables ELF to generate a robust and generalized slide-level representation. The overall ELF pipeline is shown in Figure 1\textbf{A}.

In the ELF framework, each WSI is encoded into a unified slide-level representation through three main steps: (1) tissue segmentation and patching; (2) patch-level feature embedding using five pretrained pathology foundation models; (3) patch feature aggregation and ensemble representation synthesis. For tissue segmentation and patching, we used the CLAM toolbox\cite{lu2021data} to extract non-overlapping patches of size 1024 \(\times\) 1024, 512 \(\times\) 512 or 256 \(\times\) 256 pixels at 40\(\times\), 20\(\times\) or 10\(\times\) magnification (\textasciitilde0.25 um/pixel, \textasciitilde0.50 um/pixel or \textasciitilde1.0 um/pixel) and resize to a patch size suitable for each of the tile-level foundation models\cite{el2025whole} (ablation study in Table S8). To capture diverse morphological patterns, patch-level features were extracted using five state-of-the-art foundation models with their default settings: UNI\cite{chen2024towards}, CONCHV1.5\cite{lu2024visual,ding2024multimodal}, GigaPath\cite{xu2024whole}, H-optimus-0\cite{hoptimus0}, and Virchow2\cite{zimmermann2024virchow2}. These models are pretrained on distinct large-scale pathology datasets and have demonstrated strong performance in downstream tasks\cite{xiang2025vision,el2025whole}. These heterogeneous embeddings were then fed into a unified ensemble architecture, which employs attention-based multiple instance learning to independently encode each set of features and integrate them into a compact slide-level representation that leverages the complementary value of the constituent models.

\textbf{Slide encoder.} ELF employs the attention-based multiple instance learning (ABMIL) model\cite{ilse2018attention,lu2021data} with multiple attention heads as a backbone\cite{wang2024pathology,lenz2025unsupervised,vaidya2025molecular}, because many recent studies\cite{neidlinger2024benchmarking,shao2025mil,luo2025nnmil} have demonstrated that ABMIL is a broadly applicable and effective method for slide-level aggregation, despite being less complex than variants (TransMIL\cite{shao2021transmil} and VarMIL\cite{schirris2022deepsmile} etc.). For preprocessing, input patch embeddings from different foundation models with varying dimensions (\(X \in \{\mathbb{R}^{N \times 768},\mathbb{R}^{N \times 1024},\mathbb{R}^{N \times 1280},\mathbb{R}^{N \times 1536}\}\)) are dynamically resized to a unified dimensional space (\(X^{768} \in \mathbb{R}^{N \times 768}\)) using linear interpolation, which ensures consistent feature representation across different patch encoders (ablation study in Table S9). Then, they are normalized with a LayerNorm\cite{ba2016layer} layer before the attention computation.

The slide encoder architecture utilizes a fixed 8-head attention mechanism where input features \(X^{768} \in \mathbb{R}^{N \times 768}\) are reshaped into \(\mathbb{R}^{N \times 96 \times 8}\), with each attention head processing 96-dimensional feature slices independently. Each head employs a separate ABMIL network with hidden dimensions of 96, following the gated attention formulation with two parallel branches (\(a\) and \(b\)). The attention weights \(\alpha_{i}\) for each head \(i\) are computed as:

\[\alpha_{i} = c_{i}(tanh(a_{i}X_{i}^{768}) \odot \sigma(b_{i}X_{i}^{768}))\]

where \(\odot\) denotes element-wise multiplication, \(\tanh\) and \(\sigma\) represent the hyperbolic tangent and sigmoid functions, respectively, and \(X_{i}^{768}\) represents the feature slice for head \(i\). The individual attention scores from all 8 heads are aggregated by averaging across the head dimension, followed by transpose and softmax normalization. The final slide-level features are computed by weighted aggregation using both processed and original features.

\[s_{i}^{768} = \text{softmax}(\bar{\alpha})^{T}X_{i}^{768},\quad s_{i} = \text{softmax}(\bar{\alpha})^{T}X_{i}\]

where \(\bar{\alpha}\) represents the mean-aggregated attention weights across all 8 heads, \(s_{i}^{768}\) denotes the processed slide embedding with unified 768 dimensions, and \(s_{i}\) represents the raw slide embedding that preserves the original patch encoder dimensions. A unified dimension for slide embeddings facilitates model training, while \(s_{i}\) maintains the native characteristics of different foundation models, which enables consistent attention computation and leverages model-specific feature representations. These slide embeddings are fused by concatenation as the final slide representation.

ELF contains only 14.3 million (M) parameters, fewer than the other evaluated slide encoders, including COBRAII\cite{lenz2025unsupervised} (15.5 M), TITAN\cite{ding2024multimodal} (48.5 M), and Prov-GigaPath\cite{xu2024whole} (86.3 M). In terms of computational cost, the dominant bottleneck lies in tile-level feature extraction, which may take several minutes per whole-slide image depending on the image size. Notably, all features in our experiments are extracted at 10\(\times\) magnification, which substantially reduces computation compared to 20\(\times\) or 40\(\times\) settings, yielding approximately 4\(\times\) and 16\(\times\) speedups, respectively, while maintaining strong downstream performance. We further quantified the computational cost on the EBRAINS test set (573 WSIs) using an NVIDIA L40S GPU. Per-slide feature extraction times for each foundation model were as follows: GigaPath 152.4\,s, UNI 133.8\,s, Virchow2 183.6\,s, H-Optimus-0 177.6\,s, and CONCH v1.5 201.5\,s. ELF's slide-level encoding introduces a negligible overhead of 0.1\,s per WSI. In a sequential deployment setting, the total end-to-end inference time is 849s (approximately 14 minutes) per WSI. In a parallel deployment setting where all five feature extractors run simultaneously on separate GPUs, the effective wall-clock time reduces to 202s (approximately 3.5 minutes) per WSI, determined by the slowest extractor. The computational overhead is well within clinical constraints given that the turnaround time for standard H\&E and IHC staining is on the order of multiple hours to days.

\textbf{Contrastive pretraining.} We trained ELF with the momentum contrastive learning framework (MoCoV3)\cite{chen2021empirical}, which consists of a base encoder and a momentum encoder to learn discriminative slide-level representations. The momentum encoder parameters \(\theta_{k}\) are updated using exponential moving averages of the base encoder parameters \(\theta_{q}\):

\[\theta_{k} \leftarrow m\theta_{k} + (1 - m)\theta_{q}\]

where \(m \in \lbrack 0,1)\) denotes the momentum coefficient set to 0.99.

To capture both cross-foundation-model consistency and multi-view coherence, we designed a dual positive sampling strategy complemented by discriminative negative sampling. For positive pair construction, we employed two complementary strategies. First, embeddings extracted from the same tissue region using different foundation models were treated as positive pairs, enabling the model to learn model-invariant representations across diverse foundation models. Second, embeddings from two spatially overlapping tissue regions processed by the same foundation model were regarded as another type of positive pair, promoting cross-view consistency within the embedding space. To further enhance discriminability, we constructed negative pairs from embeddings corresponding to non-overlapping tissue regions. Specifically, cross-model negatives were generated by pairing embeddings from distinct regions processed by different foundation models, encouraging the model to differentiate histologically unrelated regions while mitigating model-specific bias. In addition, intra-model spatial negatives were formed by pairing distant regions within the same model to prevent representational collapse and strengthen spatial discriminability. Together, this joint positive--negative sampling scheme ensured that the learned representations achieved both cross-model alignment and spatially aware separation, reflecting biologically meaningful relationships inherent in histopathological structures. The contrastive loss follows the InfoNCE formulation\cite{oord2018representation}, where for a given query \(q\), the objective maximizes agreement with its corresponding positive key \(k^{+}\) while minimizing similarity to negative samples:

\[\mathcal{L}_{cont} = - log\frac{exp(\text{sim}(q,k^{+})/\tau)}{\sum_{i = 1}^{N}\exp(\text{sim}(q,k_{i})/\tau)}\]

where \(\text{sim}( \cdot , \cdot )\) denotes cosine similarity, \(\tau = 0.2\) is the temperature parameter, and \(N\) represents the batch size. To prevent representational collapse, queries and keys were generated by distinct encoders with the momentum-updated key encoder providing more stable targets during training.

\textbf{Weakly-supervised pretraining.} We further incorporated an auxiliary weakly supervised signal through classification supervision that leverages readily available clinical metadata. Specifically, we employed two classification heads: a binary cancer classification head that distinguishes between cancer and non-cancer samples, and a multi-class organ classification head that predicts the anatomical tissue of origin across 20 different organs. The combined classification loss is formulated as:

\[\mathcal{L}_{cls} = \frac{1}{4}\lbrack\mathcal{L}_{CE}(p_{1}^{cancer},y^{cancer}) + \mathcal{L}_{CE}(p_{2}^{cancer},y^{cancer}) + \mathcal{L}_{CE}(p_{1}^{organ},y^{organ}) + \mathcal{L}_{CE}(p_{2}^{organ},y^{organ})\rbrack\]

where \(\mathcal{L}_{CE}\) denotes cross-entropy loss, \(p_{i}\) represents the predictions from the \(i\)-th augmented view, and \(y\) denotes the corresponding ground truth labels. This weakly supervised approach provides semantic guidance during representation learning while requiring only coarse-grained annotations that are routinely collected in clinical practice. The total training objective combines both contrastive and classification losses:

\[\mathcal{L}_{total} = \mathcal{L}_{cont} + \lambda\mathcal{L}_{cls}\]

where \(\lambda\) balances the contribution of supervised and self-supervised components. This hybrid training strategy enables the model to learn both semantically meaningful and discriminative representations suitable for diverse downstream tasks\cite{wang2024pathology,zhaosuperclip}.

\textbf{Pretraining settings.} We pretrained ELF with 53,699 WSIs (90\% for training and 10\% for validation, Table S7) and the combination objective function \(\mathcal{L}_{total}\) for 300 epochs using distributed training across 8 NVIDIA H100 GPUs with a total effective batch size of 768 samples. Each sample consisted of up to 4,096 patch embeddings extracted from different foundation models (CONCHV1.5\cite{lu2024visual,ding2024multimodal}, GigaPath\cite{xu2024whole}, H-Optimus-0\cite{hoptimus0}, UNI\cite{chen2024towards}, and Virchow2\cite{zimmermann2024virchow2}), with feature dimensions varying from 768 to 1,536 depending on the patch encoder architecture. The training employed the AdamW optimizer with \(\beta_{1} = 0.9\), \(\beta_{2} = 0.95\), and \(\epsilon = 1 \times 10^{- 8}\). We implemented a cosine learning rate decay scheduler with a peak learning rate of \(1 \times 10^{- 4}\), linear warm-up over 10 epochs, and weight decay of 0.05. The momentum coefficient for the key encoder was dynamically adjusted using a cosine schedule from 0.996 to 0.999 throughout training. Mixed-precision training with automatic mixed precision (AMP) was employed to reduce memory consumption and accelerate training while maintaining numerical stability. The total loss combined contrastive and classification objectives with equal weighting (\(\lambda\)=1). We validated the trained model every 2 epochs and \emph{selected the highest-performing checkpoint} across tissue of origin and cancer vs normal classification for downstream tasks. During inference, all available patches were used to construct the final slide-level representation, ensuring no tissue information is discarded at test time. The attention-based aggregation module applies softmax normalization over all patches, making the slide-level representation robust to variations in patch count across WSIs\cite{shao2025mil,luo2025nnmil}.

\subsubsection{Model evaluation and datasets}\label{model-evaluation-and-datasets}

We evaluated the effectiveness and generalizability of the pretrained ELF model on four major clinical applications in oncology: disease classification and subtyping, therapeutic biomarker detection, anticancer therapy response prediction, and immunotherapy response prediction. As in prior works\cite{ding2024multimodal,vaidya2025molecular}, we adopted a linear probing strategy (logistic regression)\cite{he2022masked} for all downstream tasks with the frozen ELF slide encoder. Additionally, we also evaluated the performance of ELF in the aneuploidy regression task using XGB\cite{chen2016xgboost} with the same linear probing strategy. In all evaluation experiments, data were split using patient-level label stratification or according to the official dataset partitions. Each WSI was encoded into a slide embedding using the pretrained ELF slide encoder. For comparison, we benchmarked ELF against three state-of-the-art slide-level foundation models (CHIEF\cite{wang2024pathology}, Prov-GigaPath\cite{xu2024whole}, and TITAN\cite{ding2024multimodal}) under the same evaluation settings.

\textbf{Sample size and data split.} No prospective sample-size calculation was performed; cohort sizes reflected the availability of eligible cases in the retrospective or publicly available datasets. Treatment allocation was not randomized. Data splitting was performed at the patient level using official dataset partitions or stratified cross-validation, and external validation cohorts were not used for model training or threshold selection. Cohort-specific details are listed below.

\textbf{Disease classification and subtyping.} We evaluated ELF's performance on disease classification and subtyping tasks based on WSIs. To evaluate its generalizability across different organs and subtyping complexities, we used three large-scale publicly available benchmark datasets: the Basal Cell Carcinoma Classification dataset (BCCC; skin cancer, 2/3/5 classes), the BReAst Carcinoma Subtyping dataset (BRACS; breast cancer, 7 cls), and the EBRAINS dataset (brain tumors, 12/30 cls). All evaluations followed the official data splits provided by the original studies. Detailed data distributions are summarized in Table S10. The details of each dataset and evaluation are described below.

\textbf{\ul{BCCC}.} This dataset originally consisted of 1,832 WSIs from excision specimens of cutaneous basal cell carcinomas collected at the Department of Pathology at Sahlgrenska University Hospital\cite{yacob2023weakly}. It has three different classification tasks (binary, three, and five classes). The official split included training and testing sets with 1,435 and 397 WSIs, respectively. One WSI was removed from the training set due to its low image quality, resulting in 1,434 training WSIs and 397 testing WSIs (1,831 WSIs in total) used in the analysis. We conducted five-fold cross-validation on the training set and ensembled their prediction on the testing set as final predictions.

\textbf{\ul{BRACS}.} This is a breast carcinoma subtyping dataset with 6 different subtypes of lesions (including pathological benign, usual ductal hyperplasia, flat epithelial atypia, atypical ductal hyperplasia, ductal carcinoma in situ, and invasive carcinoma) and normal tissue samples (7 cls totally)\cite{brancati2022bracs}. A total of 547 WSIs is split into training, validation, and testing sets with 395, 65, and 87 WSIs, respectively.

\textbf{\ul{EBRAINS}.} This is an established brain tumor subtyping dataset for pathology model evaluation\cite{roetzer2022digital}. It includes 2,319 WSIs with coarse (12 cls) and fine-grained (30 cls) classification labels. The dataset is split\cite{chen2024towards,ding2024multimodal,vaidya2025molecular} into training, validation, and testing sets with 1,151, 595, and 573 WSIs, respectively.

\textbf{Molecular biomarker detection.} To evaluate ELF's capacity to predict clinically relevant molecular biomarkers from histopathology slides, we performed a comprehensive assessment across multiple biomarkers and cancer types. Specifically, we evaluated ELF on (1) 84 binary biomarker classification tasks from The Cancer Genome Atlas (TCGA); (2) three key biomarkers (BRAF/KRAS mutations and MSI status) across four colorectal cancer (CRC) cohorts to test generalizability; and (3) a pan-cancer aneuploidy score regression task; (4) whole genome doubling classification task and (5) tumor mutational burden classification task to evaluate ELF's ability to capture chromosomal instability--related morphological patterns. In all experiments, we compared ELF with state-of-the-art slide-level foundation models (CHIEF\cite{wang2024pathology}, Prov-GigaPath\cite{xu2024whole}, and TITAN\cite{ding2024multimodal}) under the same evaluation settings.

\textbf{\ul{TCGA biomarkers}.} We assessed ELF's ability to predict a broad panel of therapeutic biomarkers across cancers in TCGA dataset\cite{weinstein2013cancer}. We started from 54 candidate biomarkers\cite{wang2024screen} curated from My Cancer Genome (MCG)\cite{swanton2012my} and OncoKB\cite{suehnholz2020oncokb,chakravarty2017oncokb}, where some of them are associated with a clinically proven targeted therapy and others are recommended by the National Comprehensive Cancer Network (NCCN). We then applied a minimum threshold of five altered cases per biomarker--cancer pair to ensure statistical robustness. This filtering yielded a final evaluation set of 28 biomarkers across 14 cancer types, comprising 7,721 WSIs from 6,510 patients (Table S11). The included cancer types were BLCA, BRCA, CRC (COAD and READ), ESCA, LGG/GBM, LIHC, LUAD, LUSC, OV, PAAD, PRAD, SKCM, THCA, and UCEC. The biomarker panel consist of \emph{AKT1, ATM, BARD1, BRAF, BRCA1, BRCA2, CCND1, CDK4, CHEK2, EGFR, ERBB2, ESR1, FGFR2, FGFR3, IDH1, KIT, KRAS, MET, MLH1, NBN, NRAS, PIK3CA, POLE, PTEN, ROS1, TP53, TSC1}, and \emph{TSC2}. The definition of alteration status follows the curated variant annotations with therapeutic levels of evidence (Level 1--4) in the OncoKB database\cite{suehnholz2020oncokb,chakravarty2017oncokb}, which include clinically actionable somatic mutations, copy number alterations, and gene fusions where applicable. Only genes covered by the MSK-IMPACT targeted sequencing panel\cite{cheng2015memorial} were included in the TCGA biomarker benchmarking analyses. Detailed alteration definitions and class distributions for each biomarker are provided in Table S11. The performance of ELF for each biomarker was evaluated using five-fold cross-validation under a linear probing setting, with stratified sampling according to alteration status (implemented via the scikit-learn package). All data splits were performed at the patient level to prevent data leakage. For patients with multiple slides, the slide-level predictions were averaged to obtain the final patient-level prediction. The same setting was employed in all the following studies.

\textbf{\ul{\emph{BRAF}, \emph{KRAS}, and MSI biomarkers in CRC}.} To further evaluate ELF's generalizability in clinically actionable biomarker prediction, we focused on three key biomarkers (\emph{BRAF/KRAS} mutations and MSI status), across four large CRC cohorts, comprising 2,711 unique patients after accounting for overlap across cohorts. These included the Molecular and Cellular Oncology study (MCO)\cite{ward2015mco,jonnagaddala2016integration} (1,490 patients), SR386\cite{myles2025surgen} (400 patients), SR1482\cite{myles2025surgen} (399 patients), and TCGA-CRC\cite{weinstein2013cancer} (501 patients) cohorts. SR386 (primary CRC), a recently released real-world dataset, was used as the internal development cohort with five-fold cross-validation to train five independent models. These models were then applied to the three external validation cohorts: SR1482 (metastatic and primary CRC), MCO (archival resection cases from 1994--2010), and TCGA-CRC (primary CRC with genomic characterizations). For each external WSI, the final prediction was averaged across the five models. Detailed cohort statistics are provided in Table S12.

\textbf{\ul{Aneuploidy regression}.} We applied the ELF model to predict the continuous Aneuploidy Score from routine histopathology images. We evaluated the performance using the TCGA pan-cancer dataset, which included a total of 8,819 patients and 10,854 WSIs. All patients were split into training, validation, and test sets at a ratio of 7:1:2. For this regression task, we used three metrics to evaluate performance, including Pearson correlation, Mean Squared Error, and Coefficient of determination (\(R^{2}\)). Detailed data distributions are summarized in Table S13.

\textbf{\ul{Whole genome doubling classification}.} We also applied the ELF model to classify the whole genome doubling (WGD) from routine histopathology images\cite{taylor2018genomic}. We evaluated the performance using the TCGA pan-cancer dataset with the same data split as the Aneuploidy regression task. We simplified this task as the prediction of WGD positive (includes samples with 1 or 2 WGD) versus WGD negative (0 WGD) like the existing clinical study\cite{taylor2018genomic}. Detailed data distributions are summarized in Table S14.

\textbf{\ul{Tumor mutational burden classification}.} We further used the ELF model to classify the tumor mutational burden (TMB) from routine histopathology images\cite{taylor2018genomic}. We simplified this task as the positive versus negative prediction using the clinically recommended cutoff with 10 mut/Mb\cite{marcus2021fda}. We evaluated the performance using the TCGA pan-cancer dataset through five-fold cross-validation, as the total number of positive patients is too limited. Detailed data distributions are summarized in Table S14.

\textbf{Anticancer therapy response prediction.} To assess ELF's ability to predict therapeutic response directly from pre-treatment pathology slides, we curated eight cohorts comprising 679 patients across two cancer types and distinct treatment strategies. These included molecularly targeted therapy cohorts in breast and ovarian cancers, including Breast (Trastuzumab, Yale), Breast (Trastuzumab, TransNEO), Breast (Trastuzumab, IMPRESS), and Ovary (Bevacizumab); chemotherapy cohorts in breast and ovarian cancers, including Metastatic breast (Platinum), Ovary (Platinum), Breast (Chemotherapy, TransNEO), and Breast (Chemotherapy, IMPRESS). Five-fold cross-validation was performed on Metastatic breast (Platinum), Breast (Trastuzumab, Yale), Breast (Chemotherapy, TransNEO), Breast (Trastuzumab, TransNEO), Ovary (Platinum), and Ovary (Bevacizumab). Generalization ability was further evaluated using the external IMPRESS cohort\cite{huang2023artificial}, which was split into two sub-cohorts, Breast (Chemotherapy, IMPRESS) and Breast (Trastuzumab, IMPRESS), corresponding to the chemotherapy and trastuzumab treatment arms, respectively. Detailed data distributions are summarized in Table S15.

\textbf{\ul{Metastatic breast (Platinum)}.} The metastatic breast cancer cohort includes 77 patients (99 WSIs) treated sequentially with platinum, and a subset of 54 who additionally received taxane\cite{bergstrom2024deep}. Radiographic response and progression-free survival (PFS) were assessed every 2--3 months by RECIST 1.1 criteria\cite{eisenhauer2009new}.

\textbf{\ul{Breast (Trastuzumab, Yale)}.} Eighty-five patients with HER2-positive invasive breast carcinoma were retrospectively identified via the Yale Pathology database\cite{farahmand2022her2}. All patients had a pretreatment core-needle biopsy confirming HER2 overexpression or amplification by ASCO/CAP criteria and subsequently received neoadjuvant trastuzumab \(\pm\) pertuzumab before definitive surgery. Pathologic response was determined on the resection specimen and dichotomized: responders (n = 36) exhibited no residual invasive carcinoma, lymphovascular invasion, or metastasis, whereas non-responders (n = 49) had any residual invasive, lymphovascular, or metastatic tumor.

\textbf{\ul{Ovary (Platinum)}.} The dataset comprises 348 WSIs from tumor biopsies in 158 patients with high-grade serous ovarian carcinoma (HGSOC)\cite{chowdhury2023proteogenomic}. Each patient was categorised as either refractory or sensitive to platinum-based chemotherapy, based on their response to treatment. A patient was considered refractory if the disease progressed during treatment or within four weeks of completing therapy, while a patient was considered sensitive if there was no progression throughout therapy.

\textbf{\ul{Ovary (Bevacizumab)}.} We used the Ovary (Bevacizumab) dataset\cite{wang2022histopathological} that comprised 288 WSIs from 78 patients treated with first-line bevacizumab alongside platinum-based chemotherapy. Following previous work\cite{vaidya2025molecular}, non-response was defined by either radiologically confirmed tumor regrowth or a CA-125 elevation to more than twice the normal upper limit during therapy. After quality control, 85 slides from 36 patients were included in our analysis. Recurrence-free survival (RFS) was used as the time-to-event endpoint for the Kaplan--Meier analysis.

\ul{\textbf{Breast (Chemotherapy, TransNEO)} and \textbf{Breast (Trastuzumab, TransNEO)}.} This publicly available dataset (\textbf{TransNEO}) originally comprises 168 pre-treatment breast tumour biopsies from patients who received neoadjuvant chemotherapy with or without HER2-targeted therapy, each profiled by digital pathology, whole-exome sequencing, copy-number analysis and RNA sequencing\cite{sammut2022multi}. Surgical pathology was used to determine treatment response (pathological complete response versus residual disease), which were then correlated with baseline multi-omic features. Additionally, an independent cohort of 75 patients were similarly profiled and used to validate a multi-omic machine-learning model for pathological complete response. Among 203 patients who had paired WSIs and treatment response information, a total of 197 patients (116 HER2- breast cancer patients treated with chemotherapy and 81 HER2+ patients treated with trastuzumab plus chemotherapy) passed quality control and were included in this analysis.

\ul{\textbf{Breast (Chemotherapy, IMPRESS)} and \textbf{Breast (Trastuzumab, IMPRESS)}}. This dataset includes 126 breast cancer patients who received neoadjuvant therapy and subsequently underwent surgical resection from the public IMPRESS dataset\cite{huang2023artificial}. Among these, 62 HER2+ patients were treated with trastuzumab plus chemotherapy, and 64 triple-negative breast cancer patients (TNBC) were treated with chemotherapy. Pathologic complete response (pCR) was defined as the absence of residual invasive carcinoma in the breast and lymph nodes, while any remaining invasive disease constituted an incomplete response.

\textbf{Immunotherapy response prediction.} We assessed ELF's ability to predict durable response to immune checkpoint inhibitors (ICIs) across eight tumor types using 13 retrospective cohorts encompassing 1,057 patients. These cohorts included both localized and advanced disease settings, covering lung, gastrointestinal, kidney, bladder, melanoma, head and neck, endometrial, and breast cancers. ELF was evaluated using pre-treatment H\&E-stained biopsies to predict clinical benefit from ICIs. We used five-fold cross-validation on internal datasets and independent testing for external cohorts to assess generalization. Detailed data distributions are summarized in Table S16.

\textbf{\ul{Stanford immunotherapy cohorts}.} Eleven cohorts were curated from Stanford University Medical Center with IRB approval and waiver of consent, encompassing eight tumor types: NSCLC (\textbf{Lung (Stanford)}, n = 148), Gastroesophageal adenocarcinoma (\textbf{Gastro-esophageal}, n = 106), advanced endometrial cancer (\textbf{Endometrial}, n = 96), renal cell carcinoma (\textbf{Kidney}, n = 52), head and neck carcinoma (\textbf{Head and neck}, n = 47), melanoma (\textbf{Non-metastatic melanoma}, n = 63 and \textbf{Metastatic melanoma}, n = 78), bladder carcinoma (\textbf{Bladder localized}, n = 47 and \textbf{Bladder advanced}, n = 46) and breast carcinoma (\textbf{Breast neoadjuvant}, n = 32 and \textbf{Breast advanced}, n = 33). Eligible patients had localized disease or advanced (metastatic or recurrent) disease treated with anti--PD-1/PD-L1 ICB (with or without chemotherapy) between 2018 and 2023 and an available pretreatment H\&E-stained core-needle or surgical biopsy. Clinical and imaging records were extracted from the Stanford Research Repository\cite{callahan2023stanford}. Progression-free survival (PFS) was measured from treatment initiation to documented disease progression or death, with patients without an event censored at the last follow-up. Durable response was defined as PFS \(\geq\) 6 months for advanced-disease cohorts and PFS \(\geq\) 24 months for non-advanced-disease cohorts; patients who did not meet the corresponding threshold were classified as non-responders. For the breast neoadjuvant cohort, response was instead defined as pathological complete response (pCR) at surgery. Patients with missing response endpoints or insufficient follow-up to determine response status were excluded from the corresponding response analysis. Notably, models trained on the \textbf{Breast neoadjuvant} cohort (endpoint: pathological complete response) were applied to the \textbf{Breast advanced} cohort (endpoint: durable clinical benefit), providing a cross-stage external validation of the model. The detailed baseline clinical characteristics for the two larger internal and external NSCLC cohorts are summarized in Table S17.

\textbf{\ul{CMB-Lung immunotherapy cohort}.} To further evaluate generalizability, we incorporated one independent external cohort, \textbf{Lung (CMB)}, consisting of 40 lung cancer patients treated with anti--PD-1/PD-L1 ICIs\cite{cancerMoonshot2022}. After excluding six patients without follow-up data, 34 patients were included for testing. We applied models trained on Stanford lung cohort to predict immunotherapy response in this cohort.

\textbf{\ul{MSKCC Lung immunotherapy cohort}.} We further collected a larger scale NSCLC dataset from Memorial Sloan Kettering Cancer Center (MSKCC) as an independent external cohort. The \textbf{MSKCC} consists of 275 biopsy WSIs from NSCLC patients treated with anti--PD-1/PD-L1 ICIs (the baseline clinical characteristics provided in Table S17). We applied models trained on Stanford lung cohort to predict immunotherapy response in this cohort.

\textbf{Handling of missing data.} To ensure comparability, analyses were restricted to matched cases with complete data for the primary clinical variables required for each analysis, including PD-L1 where applicable. Among the remaining covariates, only TMB values in the Stanford cohort were incomplete and were imputed using the cohort median. No other variables were imputed.

\subsubsection{Model visualization}\label{model-visualization-1}

To enhance the interpretability of the model, we visualized attention maps to highlight the spatial regions driving the ELF predictions. First, WSIs were tiled with 80\% overlap to produce a fine-grained scalar attention weight. Raw scores were min--max normalized to the {[}0, 1{]} and mapped to a continuous colormap and composited over the original high-resolution WSI with a semi-transparent overlay, so that warmer hues pinpoint the most influential regions for model prediction.

\textbf{CellViT-based cellular composition analysis.} To characterize the cellular composition of regions prioritized by ELF, we applied the pretrained CellViT-SAM-H model\cite{horst2024cellvit} using its default inference settings to patches from all 275 patients in the MSKCC external validation cohort. For each patient, patches were ranked according to the attention weights generated by the trained ELF model, and top-attention patches were selected at thresholds ranging from the top 5\% to the top 50\%. CellViT-derived cell labels were used to calculate the proportions of neoplastic and inflammatory cells within each patch. Patches containing \(\geq\)50\% neoplastic cells were classified as Tumor, whereas patches containing \textless50\% neoplastic cells and \(\geq\)10\% inflammatory cells were classified as tumor microenvironment (TME). All remaining patches, predominantly representing stromal or normal tissue, were classified as Other. At each attention threshold, the proportion of selected patches assigned to each tissue category was first calculated separately for each patient and then averaged across the 275 patients. This analysis was descriptive, and no inferential statistical test was performed.

\subsubsection{Computing hardware and software}\label{computing-hardware-and-software}

We used Python (version 3.10.12), PyTorch (version 2.6.0, CUDA 12.4) (\url{https://pytorch.org}) and other necessary open-source packages for all experiments and analyses. For pretraining, we employed eight 80 GB NVIDIA H100 GPUs for multi-GPU and mixed-precision training using distributed data-parallel (DDP). We used OpenSlide (version 4.3.1), openslide-python (version 1.4.2) and CLAM\cite{lu2021data} and their required packages to process all WSIs. All downstream experiments were conducted on a cluster server with 8~\(\times\)~48 GB NVIDIA L40 GPUs. We used several public machine learning packages for linear-probe evaluation, analysis and visualization, including Scikit-learn (version 1.6.1), Scikit-survival (version 0.24.0), lifelines (version 0.30.0), XGBoost (version 3.0.2) and Matplotlib (version 3.10.0). We used five tile-level foundation models for feature extraction; these pretrained models are publicly accessible: UNI (\url{https://huggingface.co/MahmoodLab/UNI}), CONCHV1.5 (\url{https://huggingface.co/MahmoodLab/conchv1_5}), Gigapath (\url{https://huggingface.co/prov-gigapath/prov-gigapath}), Virchow2 (\url{https://huggingface.co/paige-ai/Virchow2}) and H-optimus-0 (\url{https://huggingface.co/bioptimus/H-optimus-0}). The implementations and pretrained models of comparison methods (slide encoders) were listed here: CHIEF (), Prov-GigaPath (\url{https://huggingface.co/prov-gigapath/prov-gigapath}), TITAN (\url{https://huggingface.co/MahmoodLab/TITAN}) and the second version of COBRAII (\url{https://github.com/KatherLab/COBRA}).

\subsection{Quantification and statistical analysis}\label{quantification-and-statistical-analysis}

We used balanced accuracy (BA) as the primary metric for disease classification and subtyping tasks, and the area under the receiver operating characteristic curve (AUC) for biomarker detection, anticancer therapy response prediction, and immunotherapy response prediction. For regression tasks, model performance was assessed using Pearson correlation coefficient, mean squared error (MSE), and coefficient of determination (\(R^{2}\)).

For tasks with an independent external test set, AUC confidence intervals and paired comparisons between models were assessed using the DeLong method, which accounts for correlated predictions on the same patients. For all other metrics, paired patient-level bootstrap resampling was used (B = 1,000 replicates). At each iteration, a resampling index of length n was drawn with replacement and applied to both models, and the paired performance difference (ELF minus comparator; \(\Delta\)) was recorded. The 95\% confidence interval was obtained from the percentile distribution of \(\Delta\), and a two-sided empirical P value was calculated as:

\[p = 2 \times min\left( mean(\Delta \leq 0),\ mean(\Delta \geq 0) \right),\]

capped at 1. For paired bootstrap comparisons, statistical significance was declared when the 95\% confidence interval of \(\Delta\) excluded zero. Bootstrap confidence intervals were obtained empirically and did not require a normal approximation. Bootstrap resamples containing only one class were assigned the chance-level value for the corresponding metric. For Figures 3D and S6, error bars represent the standard deviation from separate univariate patient-level bootstrap resampling for each model (1,000 replicates).

For aggregate comparisons across biomarker tasks, cohorts, or cancer types, two-sided paired Wilcoxon signed-rank tests were applied to matched task-, cohort-, or cancer-type-level performance values. For the multi-cohort comparisons in Figure 3C, cohort-specific results were combined using inverse-variance fixed-effect meta-analysis. For tasks evaluated using five-fold cross-validation, patients were stratified and partitioned at the patient level into five non-overlapping folds, with each fold used once as the held-out test set. Performance metrics were calculated separately using the held-out predictions from each fold and summarized descriptively as the mean and standard error across the five folds. This patient-level design prevented information leakage between the training and test sets while allowing all eligible patients to contribute to out-of-sample performance estimation. P values for the three prespecified comparisons of ELF with individual competitors were adjusted using the Bonferroni method, and adjusted P values below 0.05 were considered statistically significant. No additional formal tests of distributional assumptions were performed.

For survival analysis, Kaplan--Meier curves were used to visualize cohort-specific time-to-event outcomes, including progression-free survival (PFS) and recurrence-free survival (RFS), according to model-predicted risk groups or PD-L1 TPS categories. Differences between two groups were assessed using two-sided log-rank tests, whereas comparisons involving three groups were assessed using two-sided multivariate log-rank tests. Hazard ratios (HRs) and 95\% confidence intervals were estimated using univariable Cox proportional-hazards models implemented with CoxPHFitter in the Python lifelines package. For binary analyses, the low-risk group was used as the reference category. For three-group analyses, separate pairwise Cox models were fitted after restricting the data to the reference and comparison groups. Specifically, PD-L1 categories of 0 \textless{} TPS \textless{} 50\% and TPS \(\geq\) 50\% were each compared with TPS = 0, and the Medium response and High response groups were each compared with the Low response group. The above three-tier response groups were defined using tertiles of the predicted risk score. P values displayed on the Kaplan--Meier plots were derived from the log-rank tests. For cross-validation cohorts, the median predicted probability was used as the dichotomization threshold. For external validation cohort, the threshold was selected in the internal training cohort by minimizing the two-sided log-rank P value and was subsequently applied unchanged to the validation cohorts.

Multivariable logistic regression models were adjusted for age, sex, smoking history, TMB, and PD-L1 status. Odds ratios and 95\% confidence intervals were obtained by exponentiating the regression coefficients and their confidence limits, and two-sided Wald tests were used for inference.

Decision curve analysis was performed in the MSKCC external validation cohort among patients with available PD-L1 data to evaluate the clinical net benefit of the combination model relative to PD-L1 alone and the treat-all and treat-none strategies across a range of threshold probabilities. Net benefit was calculated as TP/n - FP/n \(\times\) pt / (1 - pt), where TP and FP denote the numbers of true- and false-positive classifications, respectively, n is the number of patients, and pt is the threshold probability.

\clearpage
\renewcommand{\refname}{REFERENCES}
\bibliography{resources/bibs/main,resources/bibs/dataset}

\input{supplementary-content.tex}
\end{document}

%% file: supplementary-content.tex
\clearpage
\section*{SUPPLEMENTAL INFORMATION}

\setcounter{figure}{0}
\renewcommand{\figurename}{Figure}
\renewcommand{\thefigure}{S\arabic{figure}}
\renewcommand{\theHfigure}{S\arabic{figure}}

\input{resources/latex/figures/supp-figure.tex}

\clearpage
\setcounter{table}{0}
\renewcommand{\tablename}{Table}
\renewcommand{\thetable}{S\arabic{table}}
\renewcommand{\theHtable}{S\arabic{table}}

\input{resources/latex/tables/supp-table.tex}

%% file: resources/latex/figures/supp-figure.tex
\begin{figure}
    \centering
    \includegraphics[width=1.0\textwidth]{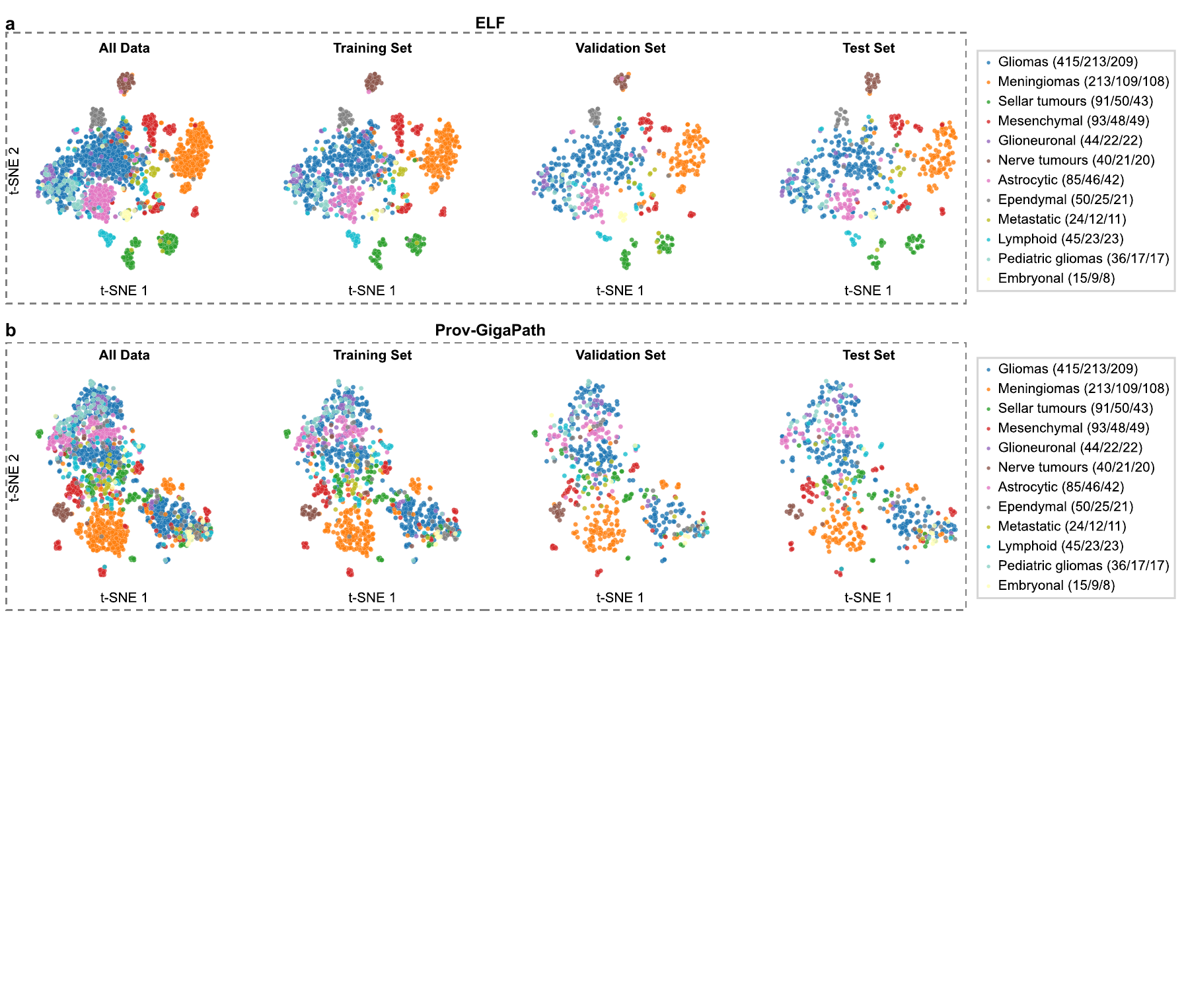}
    \caption{\textbf{Comparison of clustering capability.} Detailed comparisons of two-dimensional t-distributed stochastic neighbor embedding (t-SNE) representations in the EBRAINS dataset with 12 classes are shown for the training ($n=1{,}151$ WSIs), validation ($n=595$ WSIs), test ($n=573$ WSIs), and complete datasets ($n=2{,}319$ WSIs). \textbf{A} shows ELF ensemble representations, \textbf{B} shows single-model representations generated using the complete EBRAINS dataset ($n=2{,}319$ WSIs), and \textbf{C} shows Prov-GigaPath representations.
}

    \label{ext-fig:ext-fig1}
\end{figure}

\begin{figure}
    \centering
    \includegraphics[width=0.88\textwidth]{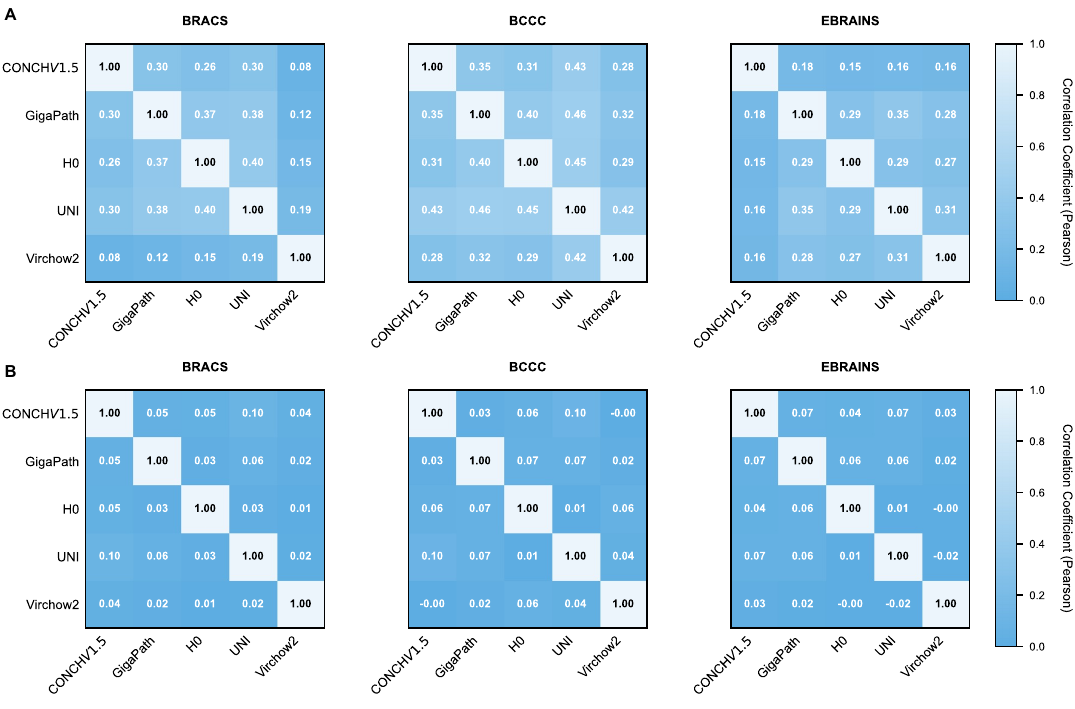}
    \caption{
\textbf{Pearson correlation analysis of attention scores and slide-level embeddings across foundation models.}
\textbf{A,} Pearson correlation analysis of attention scores across different foundation models in the BRACS ($n=547$ WSIs), BCCC ($n=1{,}831$ WSIs after quality control), and EBRAINS ($n=2{,}319$ WSIs) datasets. All attention scores were generated using the pretrained ELF slide encoder.
\textbf{B,} Pearson correlation analysis of slide-level embeddings across foundation models in the same three datasets. The pretrained slide encoder was used to generate unified 768-dimensional slide-level embeddings for the correlation analysis.}
    \label{ext-fig:ext-fig1-sub2}
\end{figure}

\begin{figure}
    \centering
    \includegraphics[width=1.0\textwidth]{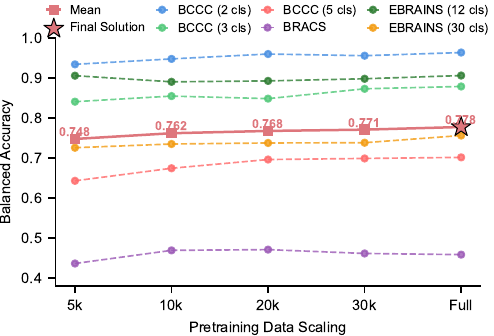}
    \caption{
\textbf{Scaling behavior of ELF across six disease-subtyping tasks.}
Balanced accuracy is shown for ELF models pretrained using 5{,}000, 10{,}000, 20{,}000, 30{,}000, or the complete set of 53{,}699 WSIs. Each pretrained model was evaluated on six fixed held-out disease-subtyping tasks: the two-, three-, and five-class BCCC tasks ($n=397$ test WSIs), the seven-class BRACS task ($n=87$ test WSIs), and the 12- and 30-class EBRAINS tasks ($n=573$ test WSIs). The mean represents the average balanced accuracy across the six tasks.
}
    \label{ext-fig:ext-scaling-train}
\end{figure}

\begin{figure}
    \centering
    \includegraphics[width=1.0\textwidth]{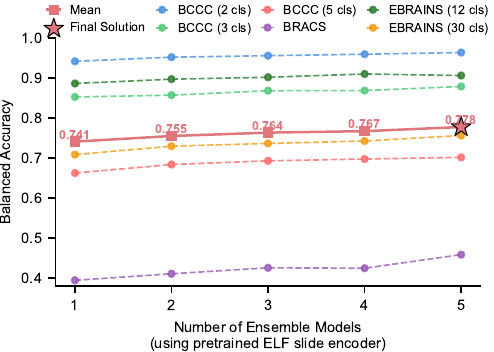}
    \caption{
\textbf{Ablation analysis of the number of foundation models included in ELF.} Balanced accuracy is shown for ensembles containing one to five of the foundation models incorporated in ELF. All configurations used the same pretrained ELF slide encoder for slide-level aggregation, thereby isolating the effect of ensemble size. For ensemble sizes of one to four models, performance was averaged over all possible model combinations (5, 10, 10, and 5 combinations, respectively); the five-model configuration represents the final ELF model. All configurations were evaluated on the same six held-out disease-subtyping tasks: the two-, three-, and five-class BCCC tasks ($n=397$ test WSIs), the seven-class BRACS task ($n=87$ test WSIs), and the 12- and 30-class EBRAINS tasks ($n=573$ test WSIs). The mean represents the average balanced accuracy across the six tasks.}
    \label{ext-fig:ext-scaling-model}
\end{figure}

\begin{figure}[p]
  \centering
  \includegraphics[width=\textwidth]{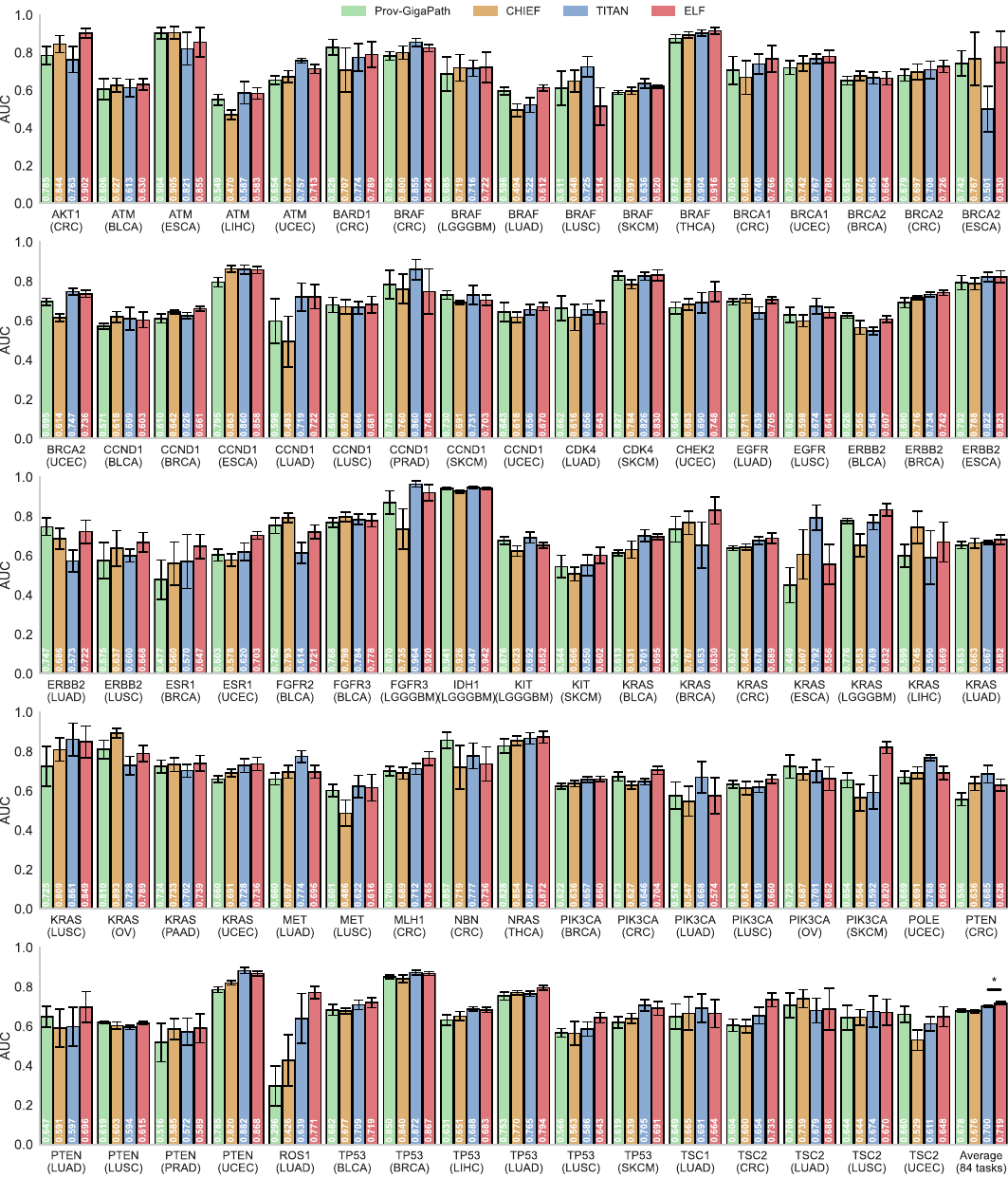}
  \caption{
\textbf{Detailed performance comparison of biomarker prediction in the TCGA dataset between ELF and other models.} Performance was evaluated across 84 biomarker--cancer combinations comprising 28 biomarkers and 14 cancer types ($n=6{,}510$ patients represented by 7{,}721 WSIs). Results are reported as mean $\pm$ standard error across five-fold cross-validation performed at the patient level. For patients represented by multiple WSIs, slide-level predictions were averaged to obtain a single patient-level prediction. The statistical results above the Average (84 tasks) bars were derived from two-sided paired Wilcoxon signed-rank tests across the 84 biomarker--cancer combinations, with Bonferroni correction for the three prespecified pairwise comparisons. Cross-validation folds were used to summarize performance and were not treated as independent units of inferential testing. \textsuperscript{*} indicates a Bonferroni-adjusted $p<0.05$.}
\end{figure}

\begin{figure}[t]
    \centering
    \includegraphics[width=0.88\textwidth]{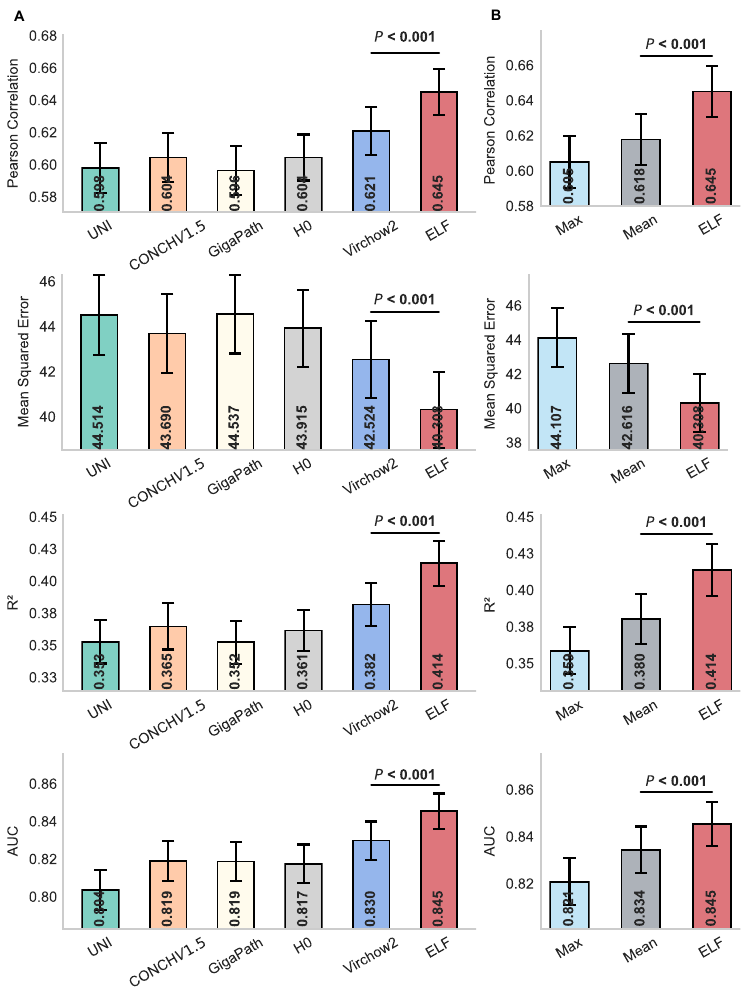}


  \caption{
\textbf{The performance of integrations of diverse foundation models and the ensemble strategy on the aneuploidy score regression and WGD classification tasks.}
\textbf{A,} ELF significantly outperforms individual foundation models on the regression and classification tasks across four metrics ($P<0.001$).
\textbf{B,} The ELF ensemble-learning strategy significantly outperforms alternative aggregation methods using the same five tile-level foundation models ($P<0.001$).
Analyses were performed in the held-out TCGA test set ($n=1{,}767$ patients). Error bars represent the standard deviation derived from separate univariate patient-level bootstrap resampling for each model (1,000 replicates). The $p$-values for the first three metrics (Pearson correlation, MSE, and $R^2$) were derived from two-sided paired bootstrap tests; AUC comparisons for WGD classification were assessed using two-sided paired DeLong tests.
}
\label{ext-fig:ext-fig7}
\end{figure}

\begin{figure}[t]
    \centering
    \includegraphics[width=1.00\textwidth]{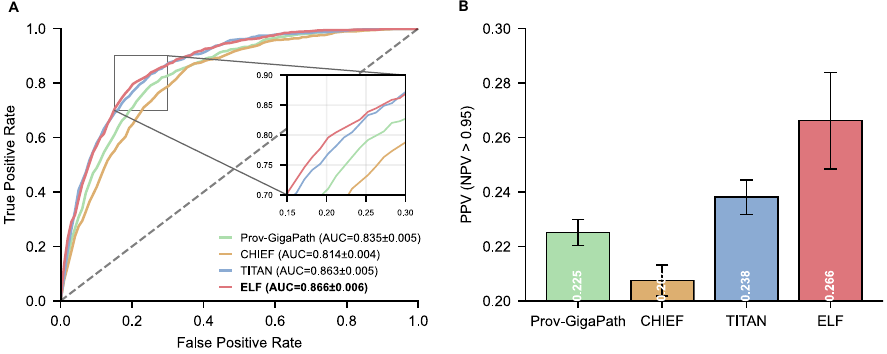}
    \caption{
\textbf{Detailed performance comparison of TMB status prediction in the TCGA pan-cancer cohort between ELF and other models.}
ELF outperforms all baseline models in terms of AUC (\textbf{A}) and PPV at NPV $>0.95$ (\textbf{B}). Analyses were performed in the TCGA pan-cancer cohort ($n=8{,}200$ patients represented by 10{,}102 WSIs), comprising 701 TMB-high and 7{,}499 TMB-low patients, using five-fold cross-validation at the patient level. For patients represented by multiple WSIs, slide-level predictions were averaged to obtain a single patient-level prediction. Error bars represent the mean $\pm$ standard error across the five cross-validation folds.
}

    \label{ext-fig:ext-fig10}
\end{figure}

\begin{figure}[t]
    \centering
    \includegraphics[width=0.8\textwidth]{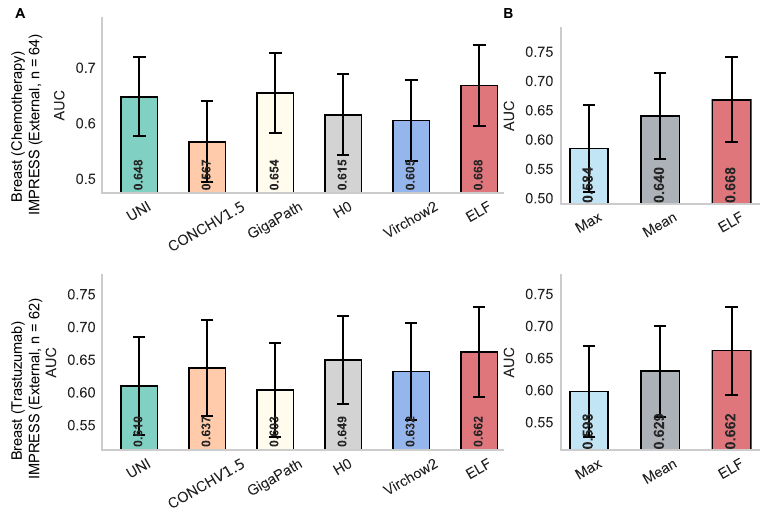}
  \caption{
\textbf{The performance of integrations of diverse foundation models and the ensemble strategy on the Breast (Chemotherapy) and Breast (Trastuzumab) cohorts.}
\textbf{A,} ELF outperforms individual foundation models for anticancer therapy response in the external Breast (Chemotherapy, IMPRESS) cohort ($n=64$ patients) and Breast (Trastuzumab, IMPRESS) cohort ($n=62$ patients).
\textbf{B,} The ELF ensemble learning strategy outperforms alternative aggregation methods using the same five tile-level foundation models across the two external cohorts. Statistical significance was assessed using the two-sided DeLong test. All results are presented as mean AUCs with standard deviations estimated from 1{,}000 patient-level bootstrap replicates.
}
\label{ext-fig:ext-fig8}
\end{figure}

\begin{figure}
    \centering
    \includegraphics[width=1.00\textwidth]{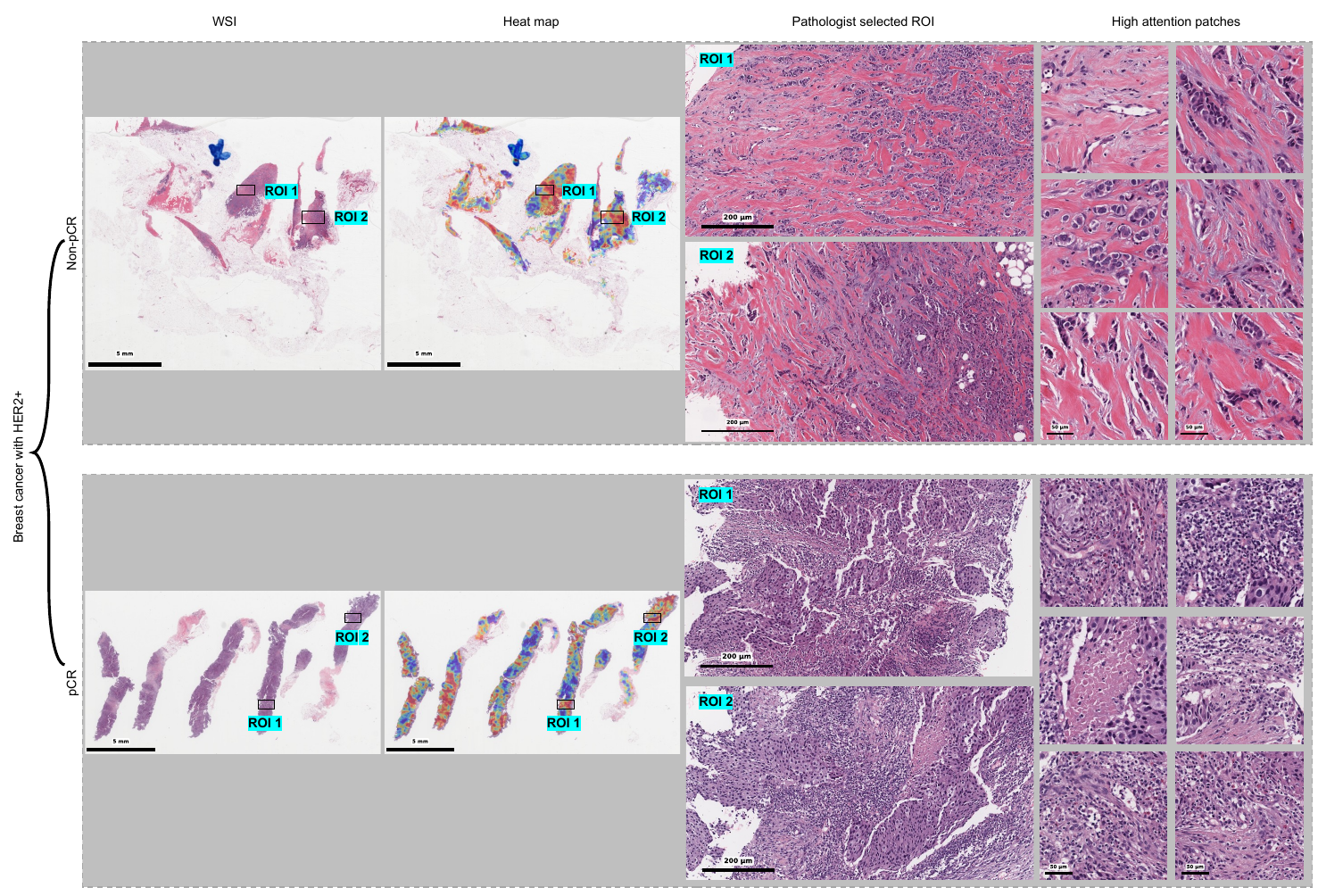}
    \caption{
\textbf{Visualization comparison between representative pCR and non-pCR cases with HER2-positive breast cancer.}
For each case, the original WSI, attention heatmap, two pathologist-selected regions of interest (ROIs) within high-attention areas, and representative high-attention patches are displayed from left to right. In the non-pCR case, the pathologist-selected ROIs correspond to a fibrotic stromal reaction surrounding the tumor, sparse peritumoral lymphocytic infiltration, and evidence of lobular differentiation. In contrast, the pCR case demonstrates markedly distinct histological features within the high-attention regions, characterized by abundant tumor-infiltrating lymphocytes (TILs), high-grade nuclear pleomorphism with marked cytological atypia, and tumor necrosis. The top-ranked attention patches were further examined after excluding border patches, showing consistency with the case-specific histological features described above and with the heatmap-level observations. Scale bars, 5~mm for the WSIs and attention heatmaps, 200~$\mu$m for the pathologist-selected ROIs, and 50~$\mu$m for the high-attention patches.
}
    \label{ext-fig:ext-HER2+-pCR}
\end{figure}

\begin{figure}
    \centering
    \includegraphics[width=1.00\textwidth]{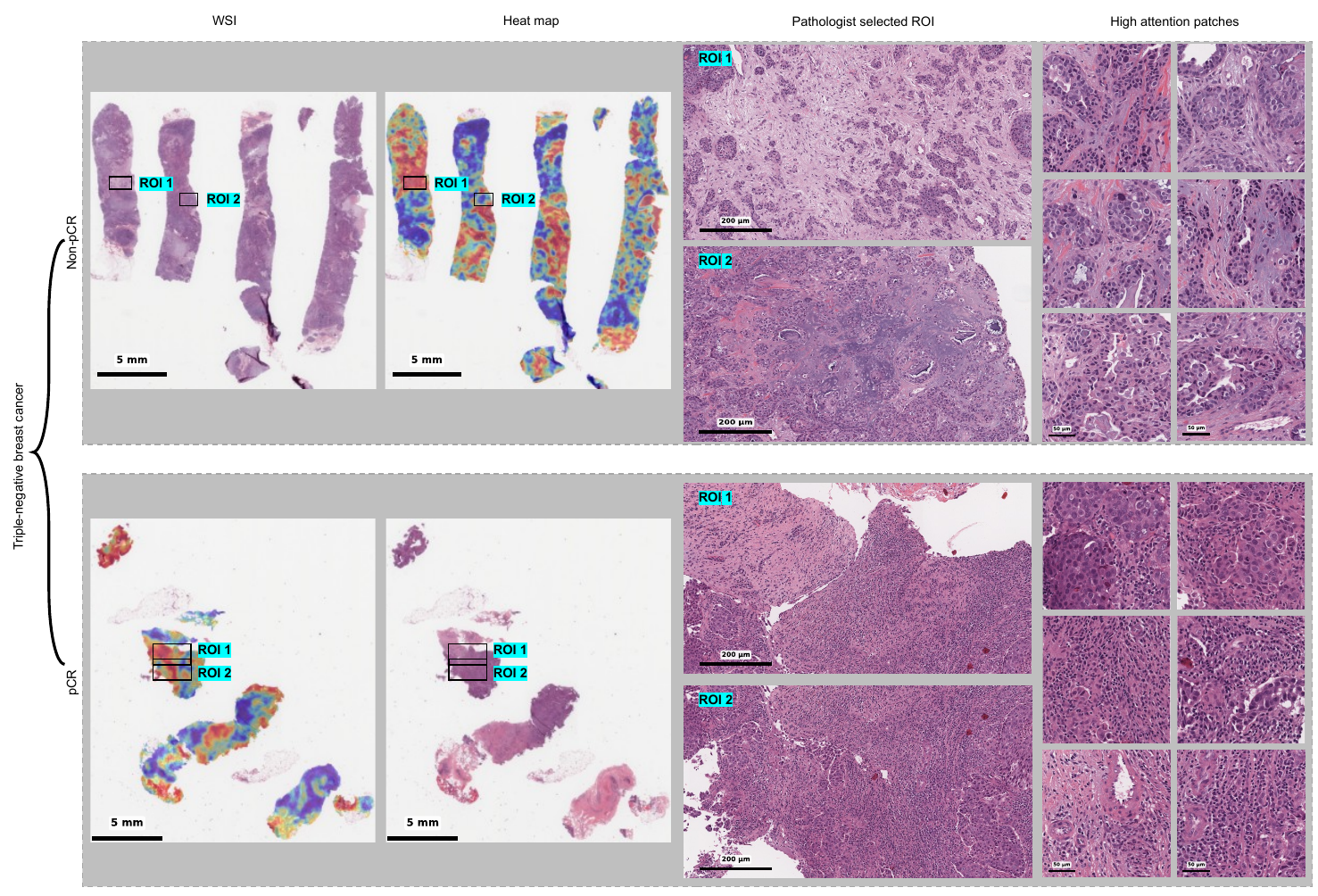}
    \caption{
\textbf{Visualization comparison between representative pCR and non-pCR cases with triple-negative breast cancer.}
For each case, the original WSI, attention heatmap, two pathologist-selected regions of interest (ROIs) within high-attention areas, and representative high-attention patches are displayed from left to right. In the non-pCR case, the pathologist-selected ROIs show a fibrotic stromal reaction surrounding the tumor with sparse peritumoral lymphocytic infiltration (ROI 1) and matrix-producing differentiation (ROI 2), indicative of an immunologically ``cold'' tumor microenvironment with limited immune engagement. In contrast, the pCR case shows prominent peritumoral lymphocytic infiltration (ROI 1) and abundant tumor-infiltrating lymphocytes accompanied by high-grade nuclear pleomorphism (ROI 2), reflecting an immune-rich tumor microenvironment associated with pathological complete response in this case. Scale bars, 5~mm for the WSIs and attention heatmaps, 200~$\mu$m for the pathologist-selected ROIs, and 50~$\mu$m for the high-attention patches.
}
    \label{ext-fig:ext-TNBC-pCR}
\end{figure}

\begin{figure}
    \centering
    \includegraphics[width=1.00\textwidth]{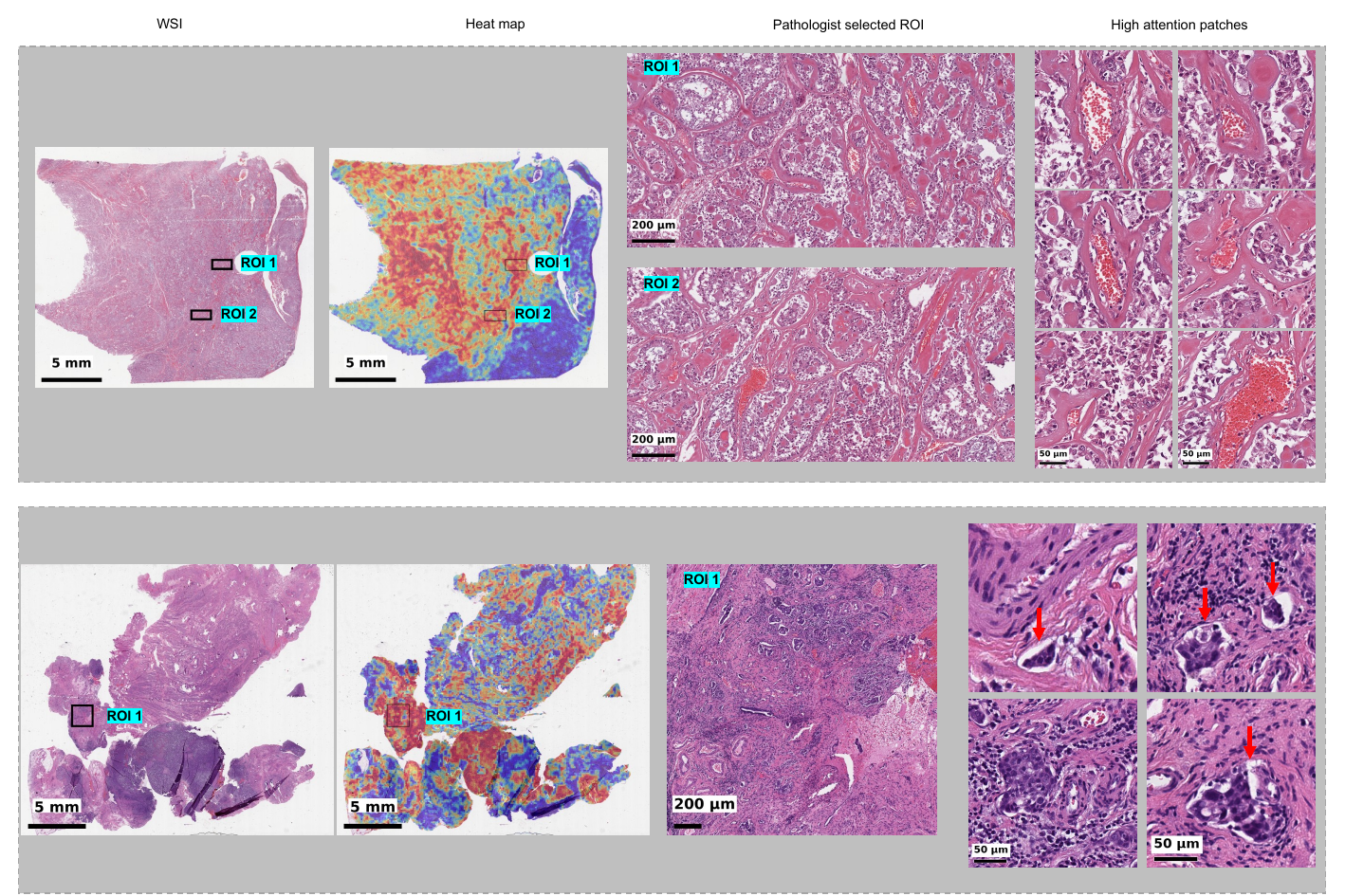}
    \caption{
\textbf{Visualization of attention maps in the Ovary (Bevacizumab) cohort.}
From left to right, each case shows the original WSI, attention heatmap, pathologist-selected regions of interest (ROIs), and representative high-attention patches. In the first case (top), the high-attention regions show prominent perivascular remodeling, characterized by thickened, hyalinized, and congested vessels with associated stromal fibrosis and tumor cells arranged around the remodeled vascular structures. In the second case (bottom), the pathologist-selected ROI and high-attention patches show increased numbers of small vessels with perivascular stromal remodeling; areas of vascular invasion are indicated by red arrows. Scale bars, 5~mm for the WSIs and attention heatmaps, 200~$\mu$m for the pathologist-selected ROIs, and 50~$\mu$m for the high-attention patches.
}
    \label{ext-fig:ext-ovary}
\end{figure}

\begin{figure}[t]
    \centering
    \includegraphics[width=0.8\textwidth]{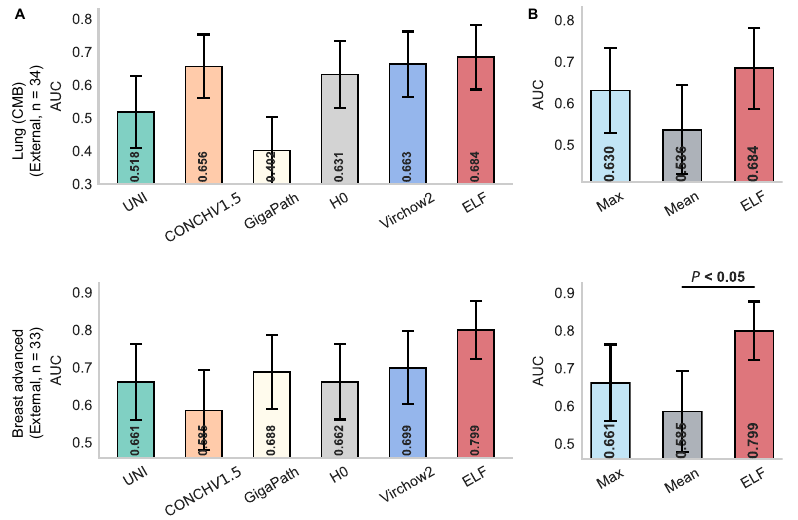}
  \caption{
\textbf{The performance of integrations of diverse foundation models and the ensemble strategy on the Lung (CMB) and Breast advanced immunotherapy cohorts.}
\textbf{A,} ELF outperformed individual foundation models for immunotherapy response prediction in the external Lung (CMB) cohort ($n=34$ patients) and Breast advanced cohort ($n=33$ patients).
\textbf{B,} The ELF ensemble learning strategy outperformed alternative aggregation methods using the same five tile-level foundation models across the two external cohorts. Statistical significance was assessed using the two-sided paired DeLong test. Results are presented as mean AUCs with standard deviations estimated from 1{,}000 patient-level bootstrap replicates.
}
\label{ext-fig:ext-fig9}
\end{figure}

\begin{figure}[t]
    \centering
    \includegraphics[width=0.9\textwidth]{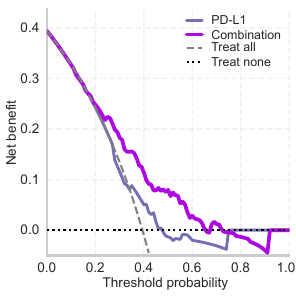}
\caption{
\textbf{Decision curve analysis (DCA) in the MSKCC external validation cohort.}
Decision curve analysis was performed in the strictly paired subset of patients with available PD-L1 results ($n=226$). Net benefit is plotted against threshold probability for PD-L1 expression alone, the combination model, treat-all, and treat-none strategies. The combination model demonstrates greater net benefit than PD-L1 alone and the treat-all strategy across threshold probabilities of 0.20--0.75.
}
\label{ext-fig:mskcc-dca}
\end{figure}

\begin{figure}
    \centering
\includegraphics[width=\textwidth]{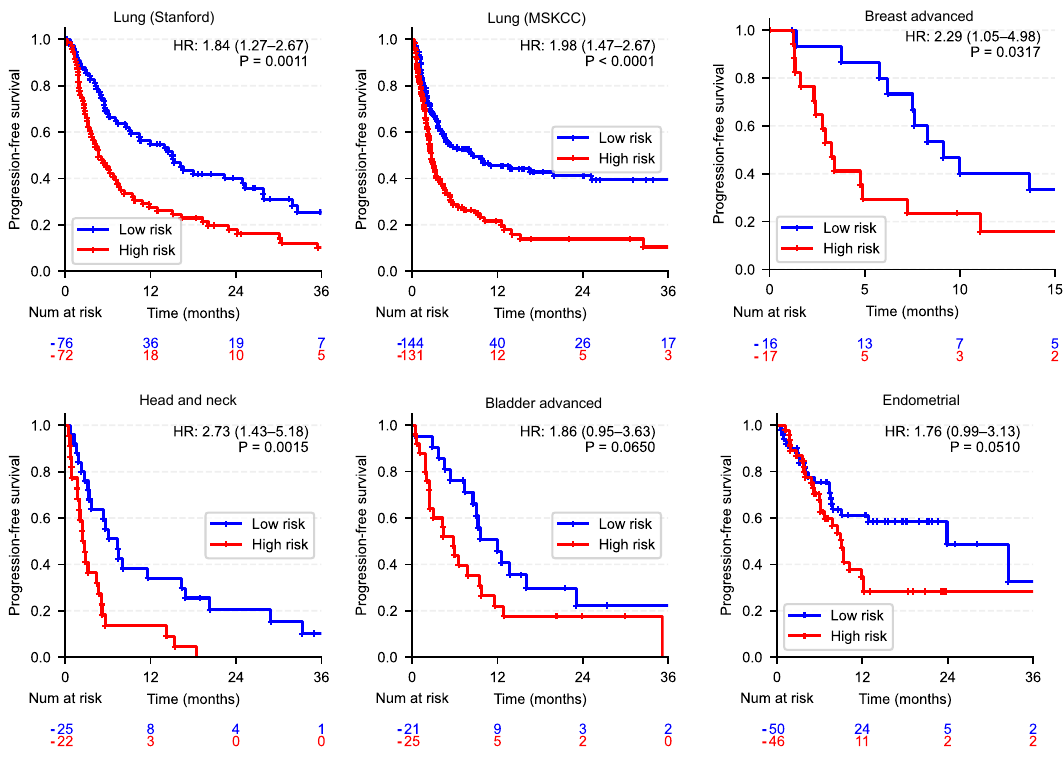}
    \caption{
\textbf{Kaplan--Meier analysis of progression-free survival across immunotherapy cohorts.}
Kaplan--Meier curves show progression-free survival in six immunotherapy cohorts: Lung (Stanford; $n=148$), Lung (MSKCC; $n=275$), Breast advanced ($n=33$), Head and neck ($n=47$), Bladder advanced ($n=46$), and Endometrial ($n=96$). For the two lung cohorts, patients were stratified using the fixed ELF risk-score threshold determined in the Stanford training cohort. For the other four cohorts, patients were stratified using the median ELF-predicted risk score within each cohort. ELF-predicted risk groups showed significant differences in progression-free survival in the Lung (Stanford), Lung (MSKCC), Breast advanced, and Head and neck cohorts, but not in the Bladder advanced or Endometrial cohorts. Hazard ratios were estimated using univariable Cox proportional hazards models, and survival differences between the high- and low-risk groups were assessed using two-sided log-rank tests.
}
    \label{ext-fig:ext-fig5}
\end{figure}

\begin{figure}
    \centering
\includegraphics[width=\textwidth]{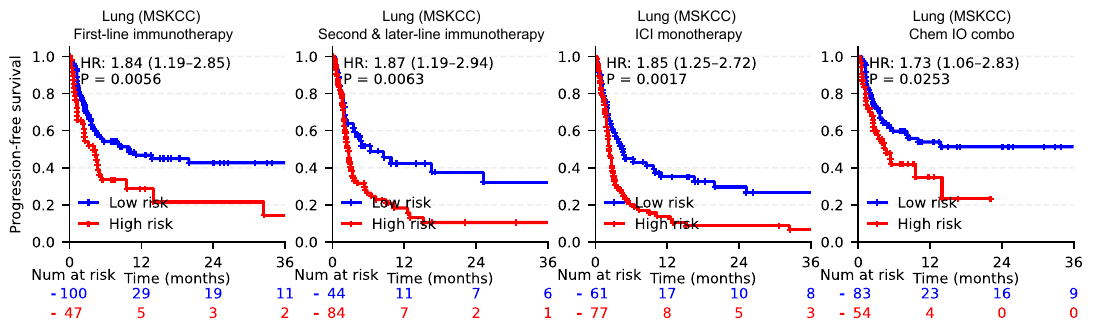}
    \caption{
\textbf{Kaplan--Meier analysis of progression-free survival in subgroups defined by line of therapy and treatment strategy in the external MSKCC validation cohort.}
Kaplan--Meier curves show progression-free survival separately for patients receiving first-line immunotherapy ($n=147$), second- or later-line immunotherapy ($n=128$), ICI monotherapy ($n=138$), or chemotherapy--immunotherapy combination therapy ($n=137$). Patients were dichotomized into high- and low-risk groups using the fixed ELF risk-score threshold determined in the Stanford development cohort. Hazard ratios were estimated using univariable Cox proportional hazards models, and survival differences between the high- and low-risk groups were assessed using two-sided log-rank tests.
}
    \label{ext-fig:line_treatment_msk}
\end{figure}

\begin{figure}
    \centering
    \includegraphics[width=1.00\textwidth]{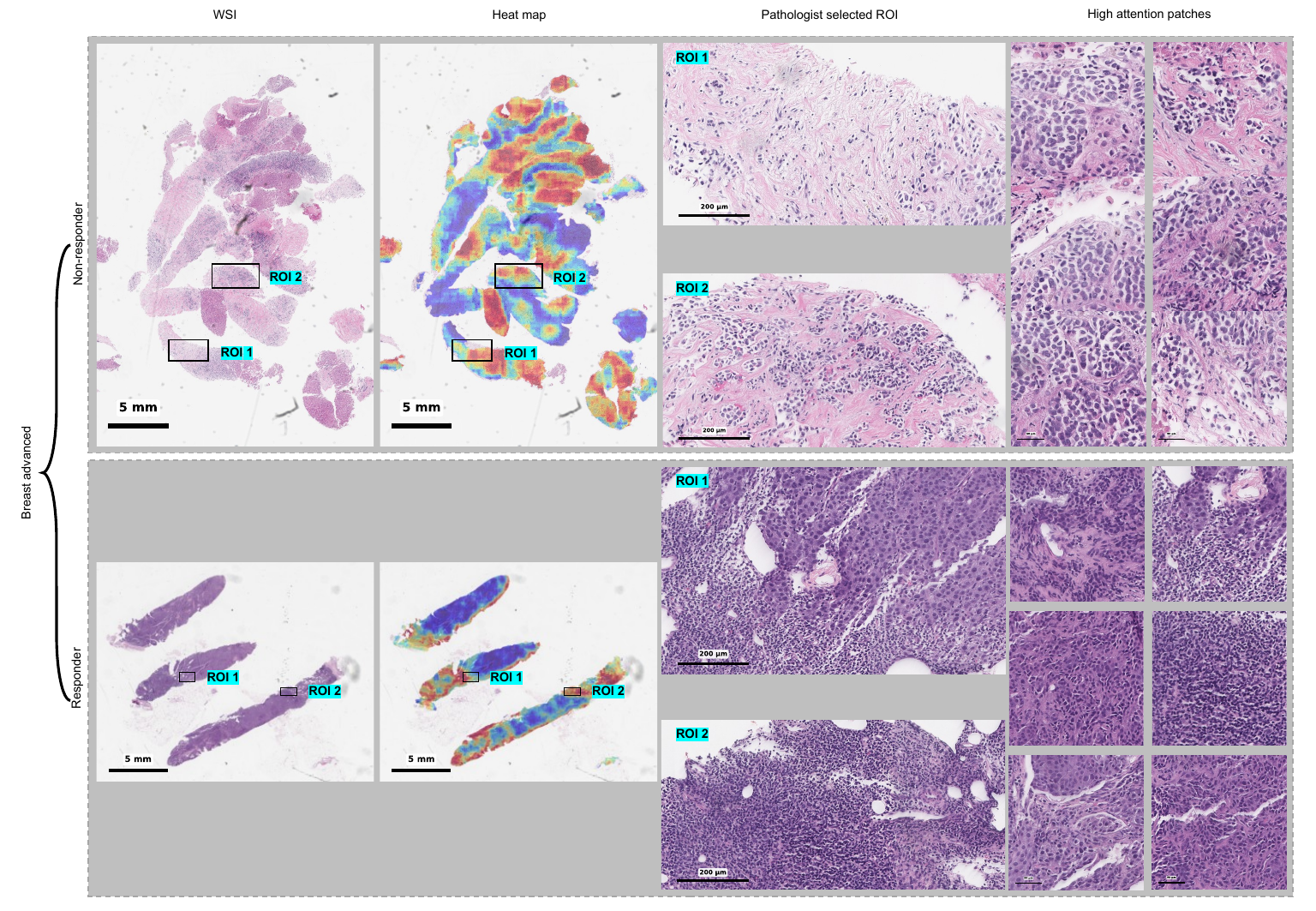}
    \caption{
\textbf{Visualization comparison between representative responder and non-responder cases in the Breast advanced cohort.}
For each case, the original WSI, attention heatmap, two pathologist-selected regions of interest (ROIs) within high-attention areas, and representative high-attention patches are displayed from left to right. In the non-responder case, the high-attention regions are dominated by densely collagenized stroma with scant lymphocytic infiltration. In contrast, the high-attention regions in the responder case show syncytial tumor cell nests with marked nuclear pleomorphism and dense intra- and peritumoral lymphocytic infiltration, reflecting an immune-rich tumor microenvironment associated with immunotherapy response in this case. Scale bars, 5~mm for the WSIs and attention heatmaps, 200~$\mu$m for the pathologist-selected ROIs, and 50~$\mu$m for the high-attention patches.
}
    \label{ext-fig:ext-BRCA-Adv}
\end{figure}

\begin{figure}
    \centering
    \includegraphics[width=1.00\textwidth]{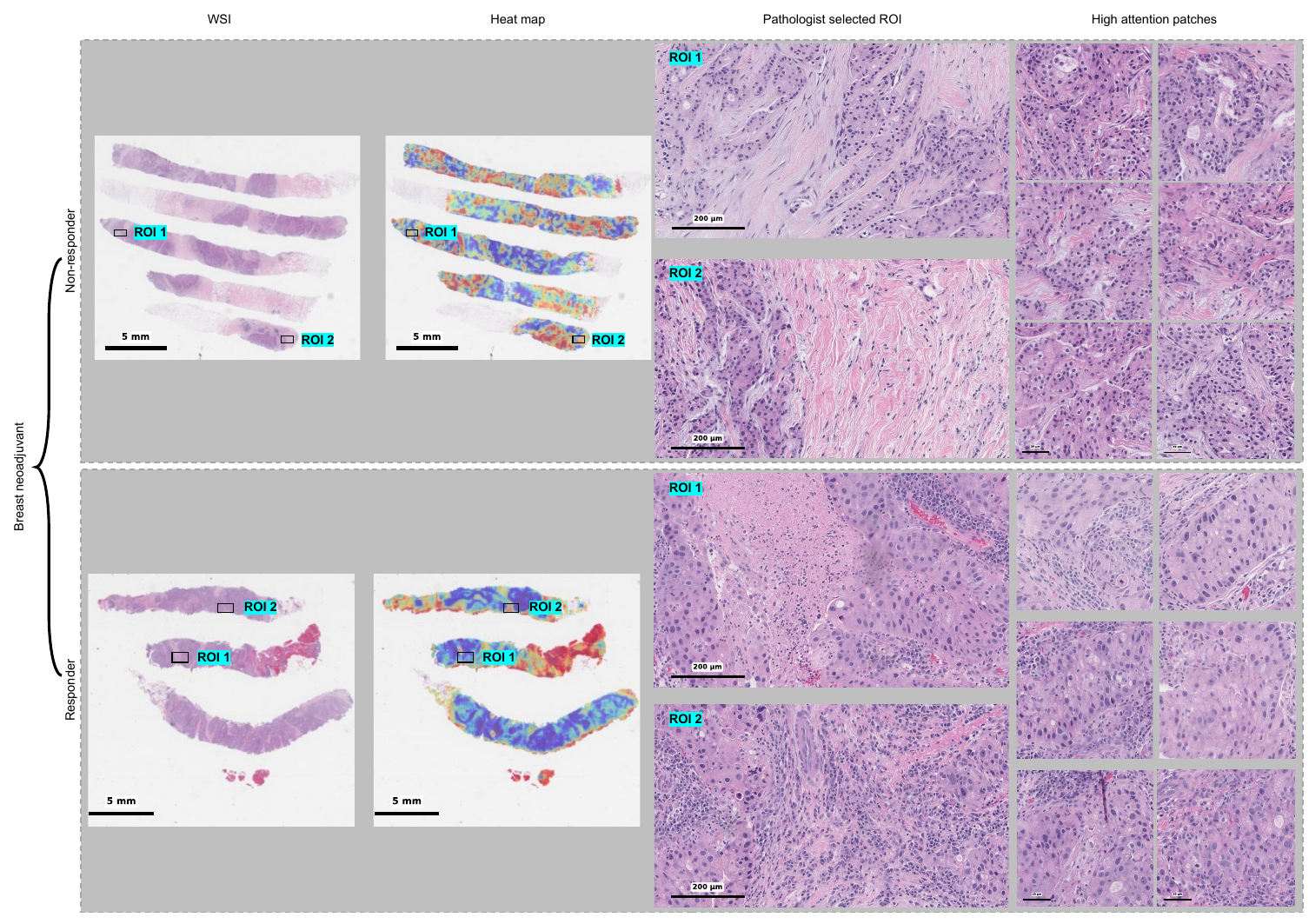}
    \caption{
\textbf{Visualization comparison between representative responder and non-responder cases in the Breast neoadjuvant cohort.}
For each case, the original WSI, attention heatmap, two pathologist-selected regions of interest (ROIs) within high-attention areas, and representative high-attention patches are displayed from left to right. In the responder case, the high-attention regions show prominent intratumoral and peritumoral lymphocytic infiltration, including high-grade tumor morphology with nuclear pleomorphism (ROI 1) and tumor necrosis with abundant inflammatory infiltration (ROI 2), reflecting an immune-rich tumor microenvironment associated with pathological complete response in this case. In contrast, the non-responder case shows residual tumor with lower-grade morphology and dense intratumoral fibrosis accompanied by minimal tumor-infiltrating lymphocytes (ROI 1), as well as dense peritumoral stromal fibrosis with minimal inflammation (ROI 2). Scale bars, 5~mm for the WSIs and attention heatmaps, 200~$\mu$m for the pathologist-selected ROIs, and 50~$\mu$m for the high-attention patches.
}
    \label{ext-fig:ext-BRCA-NEO}
\end{figure}

\begin{figure}
    \centering
    \includegraphics[width=1.00\textwidth]{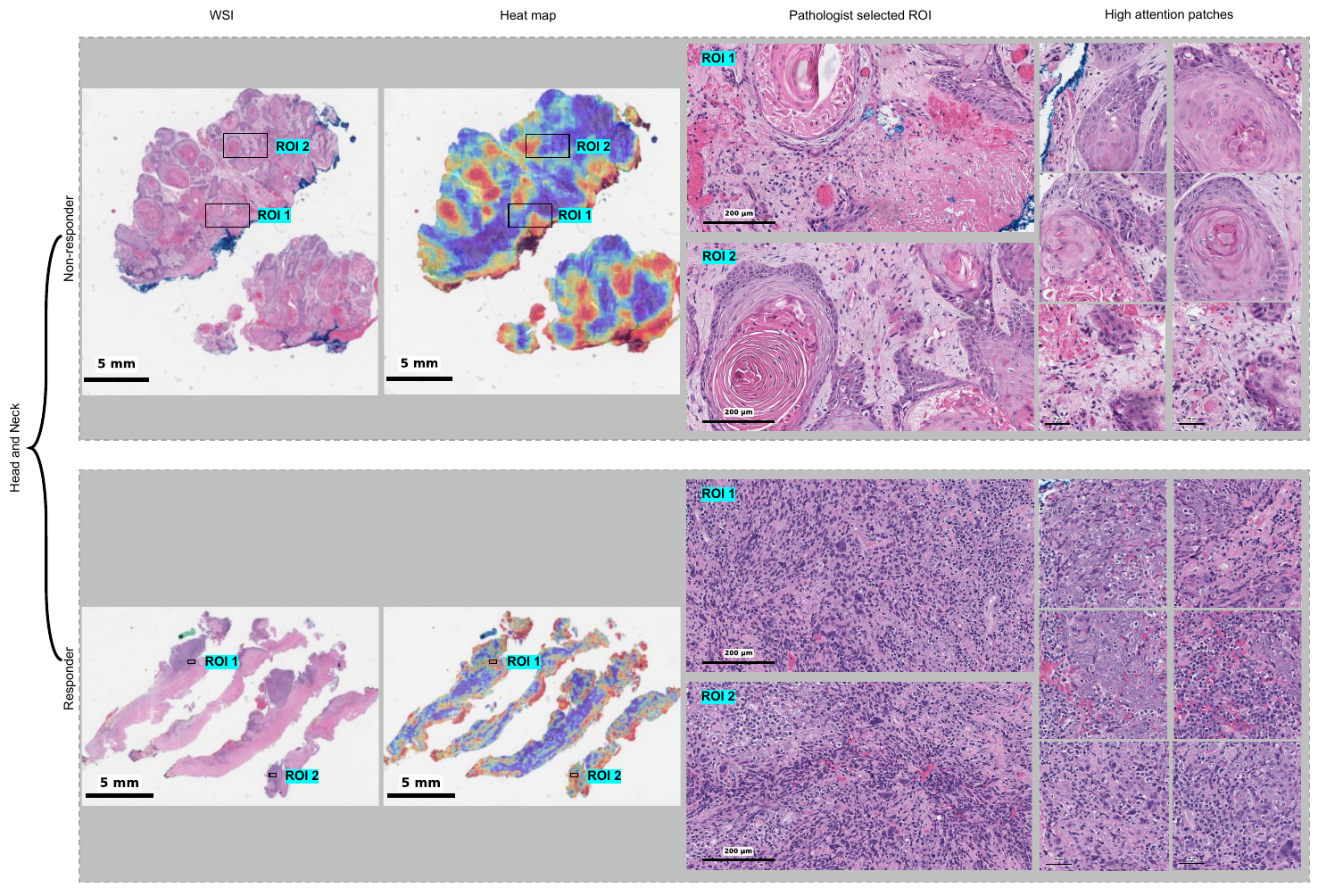}
    \caption{
\textbf{Visualization comparison between representative responder and non-responder cases in the Head and neck cohort.}
For each case, the original WSI, attention heatmap, two pathologist-selected regions of interest (ROIs) within high-attention areas, and representative high-attention patches are displayed from left to right. In the responder case, the high-attention regions show tumor cells with marked pleomorphism and bizarre nuclei accompanied by brisk tumor-infiltrating lymphocytic infiltration (ROIs 1 and 2), reflecting an immune-rich tumor microenvironment associated with immunotherapy response in this case. In contrast, the non-responder case shows well-differentiated tumor morphology and abundant intratumoral stroma containing dense collagen fibers with sparse lymphocytic infiltration (ROI 1), as well as marked fibrosis with dense collagen in the peritumoral stroma and similarly sparse lymphocytic infiltration (ROI 2). Scale bars, 5~mm for the WSIs and attention heatmaps, 200~$\mu$m for the pathologist-selected ROIs, and 50~$\mu$m for the high-attention patches.
}
    \label{ext-fig:ext-HaN}
\end{figure}

\begin{figure}
    \centering
    \includegraphics[width=1.00\textwidth]{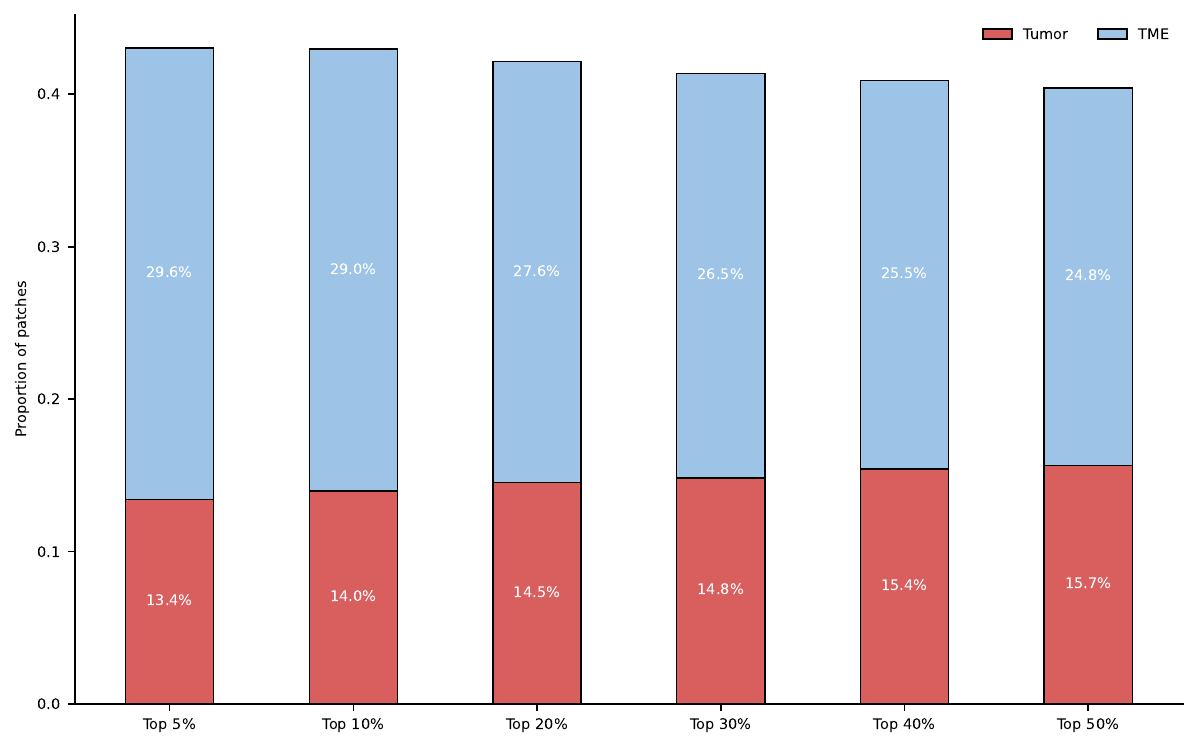}
    \caption{
\textbf{Proportions of Tumor and TME patches among high-attention patches in the MSKCC external validation cohort.}
Bar chart showing the mean patient-level proportions of patches classified as Tumor or TME among the top-attention patches at increasing attention thresholds (top 5\% to top 50\%) in the MSKCC cohort ($n = 275$ patients). Patch classification was based on cellular composition derived using CellViT-SAM-H cell segmentation~\cite{horst2024cellvit}. Tumor patches were defined as those containing $\geq 50\%$ neoplastic cells; TME patches were defined as those containing $<50\%$ neoplastic cells and $\geq 10\%$ inflammatory cells; and the remaining patches, predominantly representing stromal or normal tissue, were classified as Other. Other patches are not displayed in the bar chart. At each attention threshold, the proportions of selected patches classified as Tumor and TME were calculated separately for each patient and then averaged across patients. Attention weights were obtained from the trained ELF model. This analysis was descriptive, and no inferential statistical test was performed.
}
    \label{ext-fig:type_proportion_overall}
\end{figure}



\clearpage

%% file: resources/latex/tables/supp-table.tex
\input{resources/latex/tables/supplements/dataset.tex}

\input{resources/latex/tables/supplements/experiment.tex}

%% file: resources/latex/tables/supplements/dataset.tex
\begin{table}[htbp]
\centering
\caption{\textbf{Balanced Accuracy comparison of different aggregation methods and ensemble strategies on BRACS and EBRAINS datasets.} ABMIL needed to be trained independently for each foundation model (UNI, CONCH v1.5, GigaPath, H-optimus-0, and Virchow2) to aggregate their respective tile-level features for the final prediction. For the ensemble setting, the five sets of foundation models' features were combined by concatenating and feeding them into ABMIL for the final prediction. Mean pooling, COBRAII and ELF employed linear probing for evaluation with the same experimental settings. Values are reported as mean $\pm$ standard deviation from 1,000 bootstrap iterations, \textsuperscript{***} indicates statistical significance ($p < 0.001$) based on a paired bootstrap procedure between ELF and the second best one.}
\label{tab:ebrains_aggregation}
\scalebox{0.85}{\begin{tabular}{cclcccc}
\toprule
Dataset & Ensemble & Foundation Model & Mean Pooling & ABMIL & COBRAII & ELF \\
\hline
\multirow{7}{*}{\centering BRACS} & \multirow{5}{*}{\centering \XSolid} & Virchow2 & 0.309 $\pm$ 0.043 & {0.426 $\pm$ 0.051} & 0.370 $\pm$ 0.045 & -- \\
 & & CONCH V1.5 & 0.264 $\pm$ 0.043 & {0.407 $\pm$ 0.037} & 0.274 $\pm$ 0.046 & -- \\
 & & UNI & 0.307 $\pm$ 0.048 & {0.403 $\pm$ 0.049} & 0.296 $\pm$ 0.046 & -- \\
 & & GigaPath & 0.322 $\pm$ 0.046 & 0.385 $\pm$ 0.045 & 0.241 $\pm$ 0.041 & -- \\
 & & H0 & 0.262 $\pm$ 0.040 & 0.379 $\pm$ 0.045 & 0.264 $\pm$ 0.042 & -- \\
\cmidrule{2-7}
 & \multirow{2}{*}{\centering \Checkmark} & GPFM (3 FMs) & -- & 0.392 $\pm$ 0.050 & -- & -- \\
 & & Ensemble (5 FMs) & 0.328 $\pm$ 0.046 & 0.447 $\pm$ 0.052 & 0.273 $\pm$ 0.035 & \textbf{0.457 $\pm$ 0.051}\\
\midrule
\multirow{7}{*}{\centering EBRAINS} & \multirow{5}{*}{\centering \XSolid} & Virchow2 & 0.702 $\pm$ 0.020 & 0.690 $\pm$ 0.022 & 0.697 $\pm$ 0.020 & -- \\
 & & CONCH V1.5 & 0.673 $\pm$ 0.021 & 0.699 $\pm$ 0.022 & 0.671 $\pm$ 0.021 & -- \\
 & & UNI & 0.691 $\pm$ 0.021 & 0.679 $\pm$ 0.022 & 0.688 $\pm$ 0.021 & -- \\
 & & GigaPath & 0.669 $\pm$ 0.022 & 0.673 $\pm$ 0.022 & 0.673 $\pm$ 0.021 & -- \\
 & & H0 & 0.682 $\pm$ 0.021 & 0.682 $\pm$ 0.022 & 0.651 $\pm$ 0.022 & -- \\
\cmidrule{2-7}
 & \multirow{2}{*}{\centering \Checkmark} & GPFM (3 FMs) & -- & 0.664 $\pm$ 0.023 & -- & -- \\
 & & Ensemble (5 FMs) & 0.705 $\pm$ 0.021 & 0.699 $\pm$ 0.021 & 0.706 $\pm$ 0.021 & \textbf{0.757 $\pm$ 0.021}\textsuperscript{\textbf{***}} \\
\bottomrule
\end{tabular}}
\end{table}

\begin{table}
    \centering

    \caption{\textbf{{Methodological and experimental comparison of ELF against COBRAII.}} }
    \label{tab:elf_vs_cobraii}
    \resizebox{\textwidth}{!}{\begin{tabular}{lcc}
    \toprule
     & ELF & COBRAII \\
     \midrule
     Training Strategy & \makecell[c]{Contrastive learning\cite{chen2021empirical} \& \\ weakly-supervised learning\cite{wang2024pathology}} & Contrastive learning\cite{chen2021empirical} \\
      \midrule
     Slides & 53,699& 3,048\\
      \midrule
     Foundation models (FMs) & 5 & 4 \\
      \midrule
     Ensemble learning of multiple model embeddings& \checkmark & $\times$ \\
      \midrule
     Evaluation tasks & 125 & 15 \\
      \midrule
     Validation cancer types & 20+ & 4 \\
      \midrule
     \makecell[l]{Treatment Response \\ Prediction} & 1,736 patients across 21 independent cohorts& 0\\
    \bottomrule
    \end{tabular}}
    \end{table}

\begin{table}[htbp]
    \centering

    \caption{{\textbf{Comparison of COBRA II pretrained on TCGA, COBRA II\_50K retrained on our 50K-slide dataset, and ELF, evaluated by balanced accuracy.} For individual tasks, values are reported as mean $\pm$ standard deviation estimated from 1,000 patient-level bootstrap resamples. The Average row reports the mean $\pm$ standard deviation across the six task-level mean values.}}
    \label{tab:cobraii_50k_elf}
    \begin{tabular}{lccc}
    \toprule
    Dataset & COBRAII & COBRAII\_50K & ELF \\
    \midrule
    BCCC (2-class) & 0.8806 $\pm$ 0.0169 & 0.8779 $\pm$ 0.0170 & \textbf{0.9644 $\pm$ 0.0092} \\
    BCCC (3-class) & 0.7402 $\pm$ 0.0222 & 0.7790 $\pm$ 0.0211 & \textbf{0.8603 $\pm$ 0.0177} \\
    BCCC (5-class) & 0.5817 $\pm$ 0.0251 & 0.5729 $\pm$ 0.0267 & \textbf{0.7017 $\pm$ 0.0226} \\
    EBRAINS (coarse) & 0.8570 $\pm$ 0.0230 & 0.8797 $\pm$ 0.0216 & \textbf{0.9067 $\pm$ 0.0202} \\
    EBRAINS (fine) & 0.7056 $\pm$ 0.0209 & 0.7064 $\pm$ 0.0214 & \textbf{0.7565 $\pm$ 0.0205} \\
    BRACS & 0.2732 $\pm$ 0.0351 & 0.3129 $\pm$ 0.0403 & \textbf{0.4581 $\pm$ 0.0496} \\
    \midrule
    Average & 0.6731 $\pm$ 0.2239 & 0.6881 $\pm$ 0.2170 & \textbf{0.7746 $\pm$ 0.1826} \\
    \bottomrule
    \end{tabular}
\end{table}

\begin{table}
\centering
\caption{\textbf{Detailed performance of biomarker prediction across internal and external CRC cohorts.} PPV: Positive Predictive Value; NPV: Negative Predictive Value. All results are summarised at the patient level.}
\label{tab:crc-detailed-results}
\scalebox{0.78}{\begin{tabular}{cccccccccc}
\toprule
Dataset & Biomarker & Method & Evaluation& AUC & Accuracy & Sensitivity & Specificity & PPV & NPV \\
\midrule
SR386 & BRAF & ELF & Cross-validation & \textbf{0.843} & \textbf{0.805} & 0.756 & 0.811 & \textbf{0.414} & 0.963 \\
SR386 & BRAF & TITAN & Cross-validation & 0.765 & 0.678 & 0.778 & 0.665 & 0.385 & 0.960 \\
SR386 & BRAF & CHIEF & Cross-validation & 0.795 & 0.708 & 0.756 & 0.701 & 0.259 & 0.963 \\
SR386 & BRAF & Prov-GigaPath & Cross-validation & 0.804 & 0.740 & 0.511 & 0.769 & 0.241 & 0.936 \\
\hline
SR386 & KRAS & ELF & Cross-validation & \textbf{0.729} & 0.558 & 0.860 & 0.399 & \textbf{0.436} & 0.904 \\
SR386 & KRAS & TITAN & Cross-validation & 0.661 & 0.520 & 0.882 & 0.330 & 0.424 & 0.916 \\
SR386 & KRAS & CHIEF & Cross-validation & 0.697 & 0.497 & 0.870 & 0.305 & 0.421 & 0.918 \\
SR386 & KRAS & Prov-GigaPath & Cross-validation & 0.658 & \textbf{0.612} & 0.694 & 0.570 & 0.462 & 0.792 \\
\hline
SR386 & MSI & ELF & Cross-validation & \textbf{0.923} & 0.917 & 0.648 & 0.940 & 0.511 & 0.969 \\
SR386 & MSI & TITAN & Cross-validation & 0.911 & 0.840 & 0.614 & 0.859 & \textbf{0.533} & 0.967 \\
SR386 & MSI & CHIEF & Cross-validation & 0.844 & 0.780 & 0.800 & 0.777 & 0.288 & 0.983 \\
SR386 & MSI & Prov-GigaPath & Cross-validation & 0.892 & \textbf{0.920} & 0.548 & 0.951 & 0.513 & 0.962 \\
\hline
SR1482 & BRAF & ELF & External-validation & \textbf{0.776} & 0.524 & 0.918 & 0.437 & 0.264 & 0.960 \\
SR1482 & BRAF & TITAN & External-validation & 0.765 & \textbf{0.609} & 0.869 & 0.552 & \textbf{0.299} & 0.950 \\
SR1482 & BRAF & CHIEF & External-validation & 0.688 & 0.405 & 0.951 & 0.285 & 0.227 & 0.963 \\
SR1482 & BRAF & Prov-GigaPath & External-validation & 0.711 & 0.293 & 1.000 & 0.137 & 0.203 & 1.000 \\
\hline
SR1482 & KRAS & ELF & External-validation & \textbf{0.649} & 0.474 & 1.000 & 0.054 & \textbf{0.457} & 1.000 \\
SR1482 & KRAS & TITAN & External-validation & 0.528 & 0.449 & 1.000 & 0.009 & 0.446 & 1.000 \\
SR1482 & KRAS & CHIEF & External-validation & 0.586 & 0.451 & 1.000 & 0.014 & 0.447 & 1.000 \\
SR1482 & KRAS & Prov-GigaPath & External-validation & 0.619 & \textbf{0.584} & 0.542 & 0.617 & 0.530 & 0.628 \\
\hline
SR1482 & MSI & ELF & External-validation & \textbf{0.871} & \textbf{0.845} & 0.688 & 0.867 & \textbf{0.423} & 0.951 \\
SR1482 & MSI & TITAN & External-validation & 0.688 & 0.791 & 0.562 & 0.823 & 0.310 & 0.930 \\
SR1482 & MSI & CHIEF & External-validation & 0.811 & 0.682 & 0.875 & 0.655 & 0.264 & 0.974 \\
SR1482 & MSI & Prov-GigaPath & External-validation & 0.798 & 0.667 & 0.875 & 0.637 & 0.255 & 0.973 \\
\hline
MCO & BRAF & ELF & External-validation & \textbf{0.843} & \textbf{0.813} & 0.688 & 0.830 & \textbf{0.347} & 0.953 \\
MCO & BRAF & TITAN & External-validation & 0.769 & 0.749 & 0.705 & 0.755 & 0.275 & 0.951 \\
MCO & BRAF & CHIEF & External-validation & 0.800 & 0.706 & 0.792 & 0.694 & 0.254 & 0.962 \\
MCO & BRAF & Prov-GigaPath & External-validation & 0.625 & \textbf{0.880} & 0.006 & 0.995 & 0.143 & 0.884 \\
\hline
MCO & KRAS & ELF & External-validation & \textbf{0.636} & 0.328 & 0.998 & 0.034 & \textbf{0.312} & 0.972 \\
MCO & KRAS & TITAN & External-validation & 0.558 & 0.308 & 1.000 & 0.004 & 0.306 & 1.000 \\
MCO & KRAS & CHIEF & External-validation & 0.592 & 0.306 & 1.000 & 0.001 & 0.306 & 1.000 \\
MCO & KRAS & Prov-GigaPath & External-validation & 0.557 & \textbf{0.587} & 0.396 & 0.671 & 0.345 & 0.716 \\
\hline
MCO & MSI & ELF & External-validation & \textbf{0.897} & \textbf{0.886} & 0.682 & 0.916 & \textbf{0.552} & 0.950 \\
MCO & MSI & TITAN & External-validation & 0.742 & 0.517 & 0.846 & 0.468 & 0.193 & 0.953 \\
MCO & MSI & CHIEF & External-validation & 0.878 & 0.852 & 0.738 & 0.869 & 0.460 & 0.957 \\
MCO & MSI & Prov-GigaPath & External-validation & 0.795 & 0.691 & 0.764 & 0.680 & 0.265 & 0.950 \\
\hline
TCGA-CRC & BRAF & ELF & External-validation & \textbf{0.762} & \textbf{0.677} & 0.741 & 0.668 & \textbf{0.226} & 0.952 \\
TCGA-CRC & BRAF & TITAN & External-validation & 0.684 & 0.397 & 0.914 & 0.330 & 0.151 & 0.967 \\
TCGA-CRC & BRAF & CHIEF & External-validation & 0.754 & 0.547 & 0.810 & 0.512 & 0.179 & 0.954 \\
TCGA-CRC & BRAF & Prov-GigaPath & External-validation & 0.632 & 0.453 & 0.845 & 0.402 & 0.156 & 0.952 \\
\hline
TCGA-CRC & KRAS & ELF & External-validation & \textbf{0.622} & 0.415 & 1.000 & 0.003 & 0.414 & 1.000 \\
TCGA-CRC & KRAS & TITAN & External-validation & 0.564 & 0.417 & 1.000 & 0.007 & \textbf{0.415} & 1.000 \\
TCGA-CRC & KRAS & CHIEF & External-validation & 0.585 & \textbf{0.419} & 0.986 & 0.020 & 0.415 & 0.667 \\
TCGA-CRC & KRAS & Prov-GigaPath & External-validation & 0.592 & 0.415 & 1.000 & 0.003 & 0.414 & 1.000 \\
\hline
TCGA-CRC & MSI & ELF & External-validation & \textbf{0.857} & \textbf{0.841} & 0.738 & 0.859 & \textbf{0.464} & 0.952 \\
TCGA-CRC & MSI & TITAN & External-validation & 0.698 & 0.438 & 0.902 & 0.361 & 0.190 & 0.957 \\
TCGA-CRC & MSI & CHIEF & External-validation & 0.854 & 0.830 & 0.738 & 0.845 & 0.441 & 0.951 \\
TCGA-CRC & MSI & Prov-GigaPath & External-validation & 0.815 & 0.774 & 0.787 & 0.772 & 0.364 & 0.956 \\
\bottomrule
\end{tabular}}
\end{table}

\begin{table}[htbp]
\centering

\caption{
\textbf{Immunotherapy response prediction performance across cohorts.}
For five-fold cross-validation cohorts, results (AUC) denote the mean $\pm$ standard error across folds. For external cohorts, results report AUC with 95\% bootstrap confidence intervals based on 1,000 patient-level resamples. Cross-validation folds were used only to summarize performance and were not treated as independent units of inferential testing. \textsuperscript{*} indicates statistical significance (adjusted $P<0.05$) based on a two-sided paired Wilcoxon signed-rank test across the eight cancer-type-level AUCs, with Bonferroni correction for the three prespecified pairwise comparisons.}
\label{tab:io_auroc_by_subcohort}
\resizebox{\textwidth}{!}{\begin{tabular}{@{}lcccc@{}}
\toprule
Cohort & Prov-GigaPath & CHIEF & TITAN & ELF \\
\midrule
Kidney & $0.730 \pm 0.102$ & $0.713 \pm 0.077$ & $0.676 \pm 0.026$ & \textbf{\boldmath$0.752 \pm 0.083$} \\
Melanoma (non-metastatic) & $0.721 \pm 0.059$ & $0.724 \pm 0.055$ & $0.696 \pm 0.032$ & \textbf{\boldmath$0.754 \pm 0.038$} \\
Melanoma (metastatic) & $0.651 \pm 0.036$ & $0.619 \pm 0.044$ & $0.654 \pm 0.023$ & \textbf{\boldmath$0.719 \pm 0.019$} \\
Head and neck & $0.749 \pm 0.061$ & $0.507 \pm 0.052$ & $0.706 \pm 0.088$ & \textbf{\boldmath$0.788 \pm 0.058$} \\
Gastro-esophageal & $0.691 \pm 0.036$ & $0.698 \pm 0.035$ & $0.695 \pm 0.027$ & \textbf{\boldmath$0.732 \pm 0.026$} \\
Bladder (localized) & $0.652 \pm 0.084$ & $0.490 \pm 0.108$ & $0.526 \pm 0.066$ & \textbf{\boldmath$0.723 \pm 0.039$} \\
Bladder (advanced) & $0.612 \pm 0.071$ & $0.609 \pm 0.043$ & $0.398 \pm 0.098$ & \textbf{\boldmath$0.710 \pm 0.023$} \\
Endometrial & $0.570 \pm 0.013$ & $0.501 \pm 0.024$ & $0.635 \pm 0.079$ & \textbf{\boldmath$0.666 \pm 0.046$} \\
Breast (neoadjuvant) & $0.578 \pm 0.058$ & $0.611 \pm 0.131$ & $0.600 \pm 0.060$ & \textbf{\boldmath$0.650 \pm 0.073$} \\
Breast (advanced) & $0.673~(0.467\text{--}0.857)$ & $0.743~(0.556\text{--}0.904)$ & $0.562~(0.361\text{--}0.748)$ & \textbf{\boldmath$0.798~(0.635\text{--}0.936)$} \\
Lung (Stanford internal) & $0.644 \pm 0.041$ & $0.662 \pm 0.046$ & $0.666 \pm 0.028$ & \textbf{\boldmath$0.705 \pm 0.027$} \\
Lung (CMB external) & $0.646~(0.443\text{--}0.824)$ & $0.531~(0.333\text{--}0.723)$ & $0.667~(0.458\text{--}0.846)$ & \textbf{\boldmath$0.684~(0.481\text{--}0.864)$} \\
Lung (MSKCC external) & $0.607~(0.536\text{--}0.675)$ & $0.555~(0.482\text{--}0.625)$ & $0.625~(0.557\text{--}0.688)$ & \textbf{\boldmath$0.640~(0.573\text{--}0.708)$} \\
\midrule
Average (8 cancers) & $0.664 \pm 0.020$ & $0.612 \pm 0.029$ & $0.635 \pm 0.026$ & \textbf{\boldmath$0.724 \pm 0.013$}\textsuperscript{\textbf{*}} \\
\bottomrule
\end{tabular}}
\end{table}

\begin{table}[htbp]
\centering

\caption{{\textbf{Operating characteristics of the ELF combination model versus 
PD-L1 and TMB in the external MSKCC cohort.} The combination model is evaluated at a fixed threshold 
$t = 0.441$ locked on the internal Stanford cohort. PD-L1 positivity: 
TPS $\geq 1\%$; TMB positivity: $\geq 10$\,mut/Mb. Sensitivity, 
specificity, PPV, and NPV are reported with 95\% CI; 
AUC is for the continuous score within each stratum. Bold: best-performing 
biomarker per metric per stratum.}}
\label{tab:msk_biomarker_by_treatment}
\resizebox{\textwidth}{!}{\begin{tabular}{llccccc}
\toprule
\textbf{Biomarker} & \textbf{Threshold} & \textbf{Sensitivity} & \textbf{Specificity} & \textbf{PPV} & \textbf{NPV} & \textbf{AUC} \\
\midrule
Combination (ELF + clinical) & $\geq 0.441$ & $\mathbf{0.86\ (0.78\text{--}0.92)}$ & $0.44\ (0.37\text{--}0.52)$ & $0.48\ (0.41\text{--}0.56)$ & $\mathbf{0.84\ (0.75\text{--}0.91)}$ & $\mathbf{0.72\ (0.66\text{--}0.78)}$ \\
 PD-L1 (TPS\,\%) & $\geq 1\%$ & $0.63\ (0.52\text{--}0.73)$ & $0.51\ (0.42\text{--}0.60)$ & $0.46\ (0.37\text{--}0.55)$ & $0.68\ (0.58\text{--}0.77)$ & $0.60\ (0.53\text{--}0.68)$ \\
 TMB & $\geq 10$\,mut/Mb & $0.22\ (0.15\text{--}0.32)$ & $\mathbf{0.88\ (0.82\text{--}0.92)}$ & $0.52\ (0.37\text{--}0.68)$ & $0.65\ (0.59\text{--}0.71)$ & $0.51\ (0.43\text{--}0.58)$ \\
\bottomrule
\end{tabular}}
\end{table}

\begin{table}
\centering
\caption{\textbf{Data distribution used for ELF pre-training.} The model development dataset consists of 53,699 WSIs across 20 main organs from multiple public datasets, such as GTEx, PANDA, etc.}
\label{tab:slide_counts}
\begin{tabular}{lccc}
\toprule
Organ & Cancer (train/val) & No-Cancer (train/val) & Total \\
\midrule
Adrenal & 204/22 & 639/71 & 936 \\
Bladder & 433/48 & 117/12 & 610 \\
Brain & 486/53 & 2003/222 & 2764 \\
Breast & 1895/210 & 790/87 & 2982 \\
Cervix & 258/28 & 72/8 & 366 \\
Colorectal & 4038/448 & 3023/335 & 7844 \\
Esophagus & 142/16 & 2412/267 & 2837 \\
Heart & 0/0 & 3442/382 & 3824 \\
Kidney & 470/52 & 539/59 & 1120 \\
Liver & 502/55 & 543/60 & 1160 \\
Lung & 129/14 & 987/109 & 1239 \\
Ovary & 0/0 & 227/25 & 252 \\
Pancreas & 148/16 & 771/85 & 1020 \\
Prostate & 3001/333 & 7087/787 & 11208 \\
Skin & 2406/267 & 6246/691 & 9610 \\
Spleen & 0/0 & 780/86 & 866 \\
Stomach & 660/74 & 837/92 & 1663 \\
Testis & 229/25 & 528/58 & 840 \\
Thyroid & 468/53 & 804/89 & 1414 \\
Uterus & 536/59 & 495/54 & 1144 \\
\hline
Total & 16005/1773 & 32342/3579 & 53699 \\
\bottomrule
\end{tabular}
\end{table}

\begin{table}[htbp]
 
  \centering
  \caption{%
    {\textbf{Performance (AUC) of biomarker prediction across internal and external CRC cohorts
    and a lung IO cohort at three magnifications.}
    \textbf{ELF}: ensemble of five foundation-model feature extractors. Internal validation cohorts (SR386 and Stanford Lung IO) report mean AUC across five-fold cross-validation; external cohorts report AUC evaluated on the corresponding independent external validation datasets.
  }}
  \label{tab:multires_summary}
  \scalebox{0.82}{%
  \begin{tabular}{llc ccc}
  \toprule
  \multirow{2}{*}{Dataset}
    & \multirow{2}{*}{Biomarker}
    & \multirow{2}{*}{Evaluation}
    & \multicolumn{3}{c}{ELF} \\
  \cmidrule(lr){4-6}
   & & & 5X & 10X & 20X \\
  \midrule
  
  \multirow{4}{*}{SR386}
    & BRAF & \multirow{4}{*}{Cross-val.}
    & 0.815 & 0.843 & 0.814 \\
    & KRAS &
    & 0.685 & 0.729 & 0.680 \\
    & MSI  &
    & 0.911 & 0.923 & 0.875 \\
    \cmidrule(lr){2-2}\cmidrule(lr){4-6}
    & Average &
    & 0.804 & 0.831 & 0.790 \\
  \midrule
  
  \multirow{4}{*}{TCGA-CRC}
    & BRAF & \multirow{4}{*}{External val.}
    & 0.742 & 0.760 & 0.729 \\
    & KRAS &
    & 0.617 & 0.624 & 0.622 \\
    & MSI  &
    & 0.850 & 0.855 & 0.856 \\
    \cmidrule(lr){2-2}\cmidrule(lr){4-6}
    & Average &
    & 0.736 & 0.746 & 0.735 \\
  \midrule
  
  \multirow{4}{*}{MCO}
    & BRAF & \multirow{4}{*}{External val.}
    & 0.838 & 0.843 & 0.721 \\
    & KRAS &
    & 0.657 & 0.636 & 0.633 \\
    & MSI  &
    & 0.907 & 0.897 & 0.867 \\
    \cmidrule(lr){2-2}\cmidrule(lr){4-6}
    & Average &
    & 0.800 & 0.792 & 0.740 \\
  \midrule
  
  \multirow{4}{*}{SR1482}
    & BRAF & \multirow{4}{*}{External val.}
    & 0.780 & 0.776 & 0.717 \\
    & KRAS &
    & 0.580 & 0.649 & 0.621 \\
    & MSI  &
    & 0.856 & 0.871 & 0.808 \\
    \cmidrule(lr){2-2}\cmidrule(lr){4-6}
    & Average &
    & 0.738 & 0.765 & 0.715 \\
  \midrule
  
  \multirow{4}{*}{\textbf{CRC Overall}}
    & \textbf{BRAF} & \multirow{4}{*}{--}
    & 0.794 & 0.802 & 0.745 \\
    & \textbf{KRAS} &
    & 0.634 & 0.659 & 0.639 \\
    & \textbf{MSI}  &
    & 0.881 & 0.887 & 0.851 \\
    \cmidrule(lr){2-2}\cmidrule(lr){4-6}
    & \textbf{Average} &
    & 0.770 & \textbf{0.783} & 0.745 \\
  \midrule
  
  \multirow{3}{*}{\textbf{Lung IO}}
    & Stanford & Cross-val.
    & 0.708 & 0.705 & 0.677 \\
    & MSKCC & External val.
    & 0.608 & 0.640 & 0.610 \\
    \cmidrule(lr){2-2}\cmidrule(lr){4-6}
    & \textbf{Average} &
    & 0.658 & \textbf{0.673} & 0.644 \\
  
  \bottomrule
  \end{tabular}}
  \end{table}

\begin{table}[htbp]
    \centering

    \caption{{
        \textbf{Ablation study comparing feature harmonisation strategies for the shared 
        attention module.} For individual tasks, values are reported as mean $\pm$ standard deviation estimated from 1,000 patient-level bootstrap resamples. The Average row reports the mean $\pm$ standard deviation across the six task-level mean values. \textbf{ELF} uses 
        linear interpolation to generate attention weights while preserving original 
        encoder embeddings; \textbf{Zero-padding} and \textbf{Masking} pad or mask 
        features to the maximum dimensionality; \textbf{Learned Projection} replaces 
        interpolation with a per-encoder MLP projection layer. Bold indicates best 
        performance per row.
    }}
    \label{tab:ablation}
    \begin{tabular}{lcccc}
    \toprule
    Dataset & Masking & Zero-padding & Learned Projection & ELF \\
    \midrule
    BCCC (2-class) 
        & $0.8755 \pm 0.0168$ 
        & $0.9572 \pm 0.0096$ 
        & $0.9534 \pm 0.0104$ 
        & $\mathbf{0.9644 \pm 0.0092}$ \\
    BCCC (3-class) 
        & $0.6948 \pm 0.0237$ 
        & $\mathbf{0.8752 \pm 0.0166}$ 
        & $0.8587 \pm 0.0173$ 
        & $0.8603 \pm 0.0177$ \\
    BCCC (5-class) 
        & $0.5778 \pm 0.0273$ 
        & $0.6657 \pm 0.0239$ 
        & $0.6728 \pm 0.0236$ 
        & $\mathbf{0.7017 \pm 0.0226}$ \\
    EBRAINS (coarse) 
        & $0.8578 \pm 0.0226$ 
        & $0.8527 \pm 0.0228$ 
        & $0.8902 \pm 0.0212$ 
        & $\mathbf{0.9067 \pm 0.0202}$ \\
    EBRAINS (fine) 
        & $0.6948 \pm 0.0215$ 
        & $0.7028 \pm 0.0218$ 
        & $0.7185 \pm 0.0210$ 
        & $\mathbf{0.7565 \pm 0.0205}$ \\
    BRACS 
        & $0.3041 \pm 0.0356$ 
        & $0.3685 \pm 0.0463$ 
        & $0.3634 \pm 0.0492$ 
        & $\mathbf{0.4581 \pm 0.0496}$ \\
    \midrule
    Average 
        & $0.6675 \pm 0.2103$ 
        & $0.7370 \pm 0.2112$ 
        & $0.7428 \pm 0.2140$ 
        & $\mathbf{0.7746 \pm 0.1826}$ \\
    \bottomrule
    \end{tabular}
\end{table}

\begin{table}[ht]
\centering
\caption{\textbf{Data distribution of disease subtyping datasets.}}
\begin{tabular}{llcccc}
\toprule
{Datasource} 
  & {Organ} 
  & {\# Patients} 
  & {\# WSIs} 
  & {\# Classes} 
  & Train:Validation: Test\\
\midrule
BCCC & Skin & - & 1831 & 2  & 1434:0:397\\
BCCC & Skin & - & 1831 & 3  & 1434:0:397\\
BCCC & Skin & - & 1831 & 5  & 1434:0:397\\
BRACS     & Breast &  189 &  547 & 7  &  395:65:87\\
EBRAINS   & Brain  & 2147 & 2319 & 12 & 1151:595:573\\
EBRAINS   & Brain  & 2147 & 2319 & 30 & 1151:595:573\\
\bottomrule
\end{tabular}
\label{tab:subtyping}
\end{table}

\clearpage


\setlength{\tabcolsep}{4pt}
\begin{longtable}{l >{}m{0.3\textwidth} >{}c c c c}
\caption{\textbf{Data distribution of biomarker predictions in the TCGA dataset.} All results are summarised at the patient level.}\\
\toprule
{Biomarker} & {Definition} & {Cancer Type} & {Total patients} & {Pos patients} & {Neg patients} \\
\midrule
\endhead
\midrule \multicolumn{6}{c}{{next page}} \\
\midrule
\endfoot
\bottomrule
\endlastfoot
AKT1 &amplification/mutation & CRC & 500 & 11 & 489 \\
ATM &oncogenic mutation & BLCA & 384 & 53 & 331 \\
ATM &deletion/mutation & ESCA & 155 & 6 & 149 \\
ATM &deletion/mutation & LIHC & 350 & 13 & 337 \\
ATM & oncogenic mutation & UCEC & 473 & 92 & 381 \\
BARD1 & oncogenic mutation & CRC & 502 & 9 & 493 \\
BRAF & amplification/mutation & CRC & 500 & 59 & 441 \\
BRAF & amplification/mutation & LGGGBM & 721 & 18 & 703 \\
BRAF & oncogenic mutation & LUAD & 462 & 37 & 425 \\
BRAF & oncogenic mutation & LUSC & 459 & 14 & 445 \\
BRAF & oncogenic mutation & SKCM & 403 & 219 & 184 \\
BRAF & amplification/mutation & THCA & 486 & 286 & 200 \\
BRCA1 & oncogenic mutation & CRC & 502 & 14 & 488 \\
BRCA1 & germline/somatic mutation, deep deletion or fusion & UCEC & 473 & 44 & 429 \\
BRCA2 & deletion/mutation & BRCA & 1012 & 42 & 970 \\
BRCA2 & deletion/mutation & CRC & 500 & 38 & 462 \\
BRCA2 & germline/somatic point mutation/deletion or fusion & ESCA & 155 & 6 & 149 \\
BRCA2 & germline/somatic point mutation/deletion or fusion & UCEC & 473 & 75 & 398 \\
CCND1 &  amplification & BLCA & 382 & 42 & 340 \\
CCND1 &  amplification & BRCA & 1030 & 158 & 872 \\
CCND1 &  amplification/mutation & ESCA & 155 & 52 & 103 \\
CCND1 &  amplification & LUAD & 459 & 15 & 444 \\
CCND1 &  amplification & LUSC & 462 & 55 & 407 \\
CCND1 &  amplification & PRAD & 396 & 5 & 391 \\
CCND1 &  amplification & SKCM & 335 & 22 & 313 \\
CCND1 &  oncogenic mutation & UCEC & 479 & 36 & 443 \\
CDK4 & amplification & LUAD & 459 & 28 & 431 \\
CDK4 & amplification & SKCM & 246 & 43 & 203 \\
CHEK2 & deletion/mutation & UCEC & 473 & 34 & 439 \\
EGFR & amplification/mutation & LUAD & 459 & 67 & 392 \\
EGFR & amplification/mutation & LUSC & 459 & 39 & 420 \\
ERBB2 & amplification/mutation & BLCA & 381 & 64 & 317 \\
ERBB2 & amplification/mutation & BRCA & 1012 & 144 & 868 \\
ERBB2 & amplification/mutation & ESCA & 155 & 25 & 130 \\
ERBB2 & amplification/mutation & LUAD & 459 & 16 & 443 \\
ERBB2 & amplification/mutation & LUSC & 459 & 18 & 441 \\
ESR1 & oncogenic mutation & BRCA & 1025 & 9 & 1016 \\
ESR1 & amplification/mutation & UCEC & 473 & 40 & 433 \\
FGFR2 & amplification/mutation & BLCA & 381 & 12 & 369 \\
FGFR3 & amplification/mutation & BLCA & 381 & 65 & 316 \\
FGFR3 & fusion & LGGGBM & 727 & 6 & 721 \\
IDH1 & amplification/mutation & LGGGBM & 721 & 387 & 334 \\
KIT & amplification & LGGGBM & 857 & 51 & 806 \\
KIT & amplification/mutation & SKCM & 246 & 10 & 236 \\
KRAS & amplification/mutation & BLCA & 381 & 25 & 356 \\
KRAS & oncogenic mutation & BRCA & 1025 & 6 & 1019 \\
KRAS & amplification/mutation & CRC & 500 & 208 & 292 \\
KRAS & oncogenic mutation & ESCA & 155 & 13 & 142 \\
KRAS & oncogenic mutation & LGGGBM & 721 & 18 & 703 \\
KRAS & oncogenic mutation & LIHC & 350 & 7 & 343 \\
KRAS & oncogenic mutation & LUAD & 462 & 145 & 317 \\
KRAS & oncogenic mutation & LUSC & 459 & 6 & 453 \\
KRAS & oncogenic mutation & OV & 92 & 16 & 76 \\
KRAS & amplification/mutation & PAAD & 177 & 115 & 62 \\
KRAS & oncogenic mutation & UCEC & 479 & 95 & 384 \\
MET & amplification/mutation & LUAD & 459 & 27 & 432 \\
MET & amplification/mutation & LUSC & 459 & 15 & 444 \\
MLH1 & oncogenic mutation & CRC & 502 & 20 & 482 \\
NBN & deletion/mutation & CRC & 500 & 9 & 491 \\
NRAS & oncogenic mutation & THCA & 488 & 39 & 449 \\
PIK3CA & amplification/mutation & BRCA & 1012 & 358 & 654 \\
PIK3CA & oncogenic mutation & CRC & 502 & 136 & 366 \\
PIK3CA & amplification/mutation & LUAD & 459 & 33 & 426 \\
PIK3CA & amplification/mutation & LUSC & 459 & 206 & 253 \\
PIK3CA & amplification/mutation & OV & 92 & 20 & 72 \\
PIK3CA & amplification/mutation & SKCM & 246 & 7 & 239 \\
POLE & POLE oncogenic mutation & UCEC & 479 & 76 & 403 \\
PTEN & deletion/mutation & CRC & 500 & 43 & 457 \\
PTEN & deletion/mutation & LUAD & 459 & 11 & 448 \\
PTEN & deletion/mutation & LUSC & 459 & 95 & 364 \\
PTEN & oncogenic mutation & PRAD & 398 & 14 & 384 \\
PTEN & oncogenic mutation & UCEC & 479 & 315 & 164 \\
ROS1 &fusion & LUAD & 462 & 6 & 456 \\
TP53 &oncogenic mutation & BLCA & 384 & 188 & 196 \\
TP53 &oncogenic mutation & BRCA & 1025 & 341 & 684 \\
TP53 &deletion/mutation & LIHC & 350 & 112 & 238 \\
TP53 &deletion/mutation & LUAD & 459 & 239 & 220 \\
TP53 &deletion/mutation & LUSC & 459 & 386 & 73 \\
TP53 &deletion/mutation & SKCM & 248 & 81 & 167 \\
TSC1 & oncogenic mutation & LUAD & 462 & 6 & 456 \\
TSC2 & deletion/mutation & CRC & 500 & 15 & 485 \\
TSC2 & fusion & LUAD & 462 & 10 & 452 \\
TSC2 & fusion & LUSC & 459 & 17 & 442 \\
TSC2 & oncogenic mutation & UCEC & 479 & 37 & 442 \\
\label{tab:biomarker-tcga}
\end{longtable}

\clearpage

\begin{table*}
\centering
\caption{\textbf{Data distribution of biomarker predictions across CRC cohorts.} All results are summarised at the patient level.}
\scalebox{0.9}{\begin{tabular}{lccccc}
\toprule
{Dataset}
  & {Biomarker}
  & {Total patients}
  & {Positive patients}
  & {Negative patients}
  & {Evaluation} \\
\midrule
SR386 & BRAF & 400 & 45 & 355 & Cross-validation\\
SR386 & KRAS & 400 & 137 & 263 & Cross-validation\\
SR386 & MSI & 400 & 31 & 369 & Cross-validation\\
SR1482 & BRAF & 338 & 61 & 277 & External-validation\\
SR1482 & KRAS & 399 & 177 & 222 &External-validation\\
SR1482 & MSI & 129 & 16 & 113 &External-validation\\
MCO & BRAF & 1488 & 173 & 1315 &External-validation\\
MCO & KRAS & 1490 & 455 & 1035 &External-validation\\
MCO & MSI & 1488 & 195 & 1293 &External-validation\\
TCGA-CRC & BRAF & 501 & 58 & 443 &External-validation\\
TCGA-CRC & KRAS & 501 & 207 & 294 &External-validation\\
TCGA-CRC & MSI & 429 & 61 & 368 &External-validation\\
\bottomrule
\end{tabular}}
\label{tab:biomarker-crc}
\end{table*}

\begin{table}[ht]
\centering
\caption{\textbf{Data distribution of the aneuploidy scores regression task at the TCGA dataset.} All results are summarised at the patient level.}
\begin{tabular}{llccc}
\toprule
{Cancer Type} 
  & {\# Patients} 
  & {\# WSIs} 
  & {Score range} 
  & Train: Validation: Test\\
\midrule
Pan-cancer & 8819 & 10854 & 0-39  & 6163:889:1767\\
\bottomrule
\end{tabular}
\label{tab:aneuploidy}
\end{table}

\begin{table}[ht]
\centering
\caption{\textbf{Data distribution of the Whole Genome Doubling (WGD) and Tumor Mutational Burden (TMB) classification tasks at the TCGA pan-cancer dataset.} All results are summarised at the patient level.}
\begin{tabular}{llcccc}
\toprule
{Biomarker} 
  & {\# Patients} 
  & {\# WSIs} 
  & {Positive patients} 
  & {Negative patients} 
  & Train: Validation: Test\\
\midrule
WGD & 8819 & 10854 & 3094 & 5725 & 6163:889:1767\\
TMB & 8200 & 10102 & 701 & 7499 & 6560:0:1640 (5-fold)\\
\bottomrule
\end{tabular}
\label{tab:wgd}
\end{table}

\begin{table*}
\centering
\caption{\textbf{Data distribution of anticancer therapy cohorts.} All results are summarised at the patient level.}
\scalebox{0.8}{\begin{tabular}{lcccccccc}
\toprule
{Dataset}
  & {Treatment}
  & {Total patients}
  & {Responder}
  & {Non-responder} 
  & {Evaluation}\\
\midrule
Metastatic breast (Platinum) & Chemotherapy & 77 & 33 & 44 & Cross-validation \\
Breast (Trastuzumab, Yale) & Trastuzumab & 85 & 36 & 49 & Cross-validation \\
Ovary (Platinum) & Chemotherapy & 158 & 91 & 67 & Cross-validation \\
Ovary (Bevacizumab) & Bevacizumab & 36 & 25 & 11 & Cross-validation \\
Breast (Chemotherapy, TransNEO) & Chemotherapy & 116 & 26 & 90 & Cross-validation \\
Breast (Chemotherapy, IMPRESS) & Chemotherapy & 64 & 27 & 37 &External-validation\\
Breast (Trastuzumab, TransNEO) & Trastuzumab & 81 & 31 & 50 & Cross-validation \\
Breast (Trastuzumab, IMPRESS) & Trastuzumab & 62 & 38 & 24 & External-validation\\

\bottomrule
\end{tabular}}
\label{tab:targedtherapy}
\end{table*}

\begin{table*}
\centering
\caption{\textbf{Data distribution of anti-PD-1/PD-L1 immunotherapy cohorts.} All results are summarised at the patient level.}
\scalebox{0.85}{\begin{tabular}{lcccccccc}
\toprule
{Dataset}
  & {Total patients}
  & {Responder}
  & {Non-responder}
  & Evaluation\\
\midrule
Bladder advanced & 46 & 27 & 19 & Cross-validation \\
Bladder localized & 47 & 14 & 33 & Cross-validation \\
Kidney & 52 & 43 & 9 & Cross-validation \\
Non-metastatic melanoma & 63 & 31 &32 & Cross-validation \\
Metastatic melanoma & 78 & 58 & 20 & Cross-validation \\
Head and neck & 47 & 16 & 31 & Cross-validation \\
Gastroesophageal & 106 & 50 & 56 & Cross-validation \\
Endometrial & 96 & 59 & 37 & Cross-validation \\
Lung (Stanford) & 148 & 81 & 67 & Cross-validation \\
Lung (CMB) & 34 & 18 & 16 & External-validation\\
{Lung (MSKCC)} & {275} & {103} & {172} & {External-validation} \\
Breast neoadjuvant {(pCR)} & 32 & 15 & 17 & Cross-validation \\
Breast advanced & 33 & 17 & 16 & External-validation\\

\bottomrule
\end{tabular}}
\label{tab:immunotherapy}
\end{table*}

\begin{table}[htbp]
\centering

\caption{{\textbf{Baseline clinical characteristics of lung cancer immunotherapy cohorts.} Data are presented as no.\ (\%) unless otherwise indicated. IQR = interquartile range; 1L/2L+ = first/second-or-later line; PD-L1 = programmed death-ligand 1;
TPS = tumor proportion score.}}
\label{tab:lung_io_baseline}
\small
\begin{tabular}{@{}l c c@{}}
\toprule & \textbf{Stanford Internal} & \textbf{MSKCC External} \\
 & ($n = 148$)               & ($n = 275$)            \\
\midrule
Age at treatment, yr, median (IQR) & 70.3 (63.4--76.0) & 68.0 (60.0--75.0) \\
\addlinespace[6pt]
\textbf{Sex} & & \\
\quad Male                         & 86 (58.1)   & 131 (47.6) \\
\quad Female                       & 62 (41.9)   & 144 (52.4) \\
\addlinespace[6pt]
\textbf{Smoking history} & & \\
\quad Yes                          & 100 (67.6)   & 213 (77.5) \\
\quad No                           & 47 (31.7)   & 62 (22.5)  \\
\quad Unknown                      & 1 (0.7)   & 0           \\
\addlinespace[6pt]
\textbf{Histology} & & \\
\quad Adenocarcinoma               & 122 (82.4)  & 227 (82.5) \\
\quad Squamous cell carcinoma      & 20 (13.5)   & 31 (11.3)  \\
\quad Other                        & 6 (4.1)   & 17 (6.2)   \\
\addlinespace[6pt]
\textbf{PD-L1 TPS} & & \\
\quad $= 0\%$                      & 38 (25.7)   & 103 (37.5) \\
\quad $1$--$49\%$                   & 42 (28.4)   & 61 (22.2)  \\
\quad $\geq 50\%$                   & 47 (31.7)   & 62 (22.5)  \\
\quad Unknown                      & 21 (14.2)   & 49 (17.8)  \\
\addlinespace[6pt]
\textbf{Line of therapy} & & \\
\quad 1L                           & 84 (56.8)   & 147 (53.5) \\
\quad 2L or later                           & 64 (43.2)   & 128 (46.5) \\
\addlinespace[6pt]
\textbf{Concurrent chemotherapy} & & \\
\quad Yes                          & 90 (60.8)   & 97 (35.3)  \\
\quad No                           & 58 (39.2)   & 178 (64.7) \\
\bottomrule
\end{tabular}
\end{table}

%% file: resources/bibs/dataset.bib
@article{shao2021transmil,
  title={Transmil: Transformer based correlated multiple instance learning for whole slide image classification},
  author={Shao, Zhuchen and Bian, Hao and Chen, Yang and Wang, Yifeng and Zhang, Jian and Ji, Xiangyang and others},
  journal={Advances in neural information processing systems},
  volume={34},
  pages={2136--2147},
  year={2021}
}

@article{schirris2022deepsmile,
  title={DeepSMILE: Contrastive self-supervised pre-training benefits MSI and HRD classification directly from H\&E whole-slide images in colorectal and breast cancer},
  author={Schirris, Yoni and Gavves, Efstratios and Nederlof, Iris and Horlings, Hugo Mark and Teuwen, Jonas},
  journal={Medical image analysis},
  volume={79},
  pages={102464},
  year={2022},
  publisher={Elsevier}
}

@article{horst2024cellvit,
  title={Cellvit: Vision transformers for precise cell segmentation and classification},
  author={H{\"o}rst, Fabian and Rempe, Moritz and Heine, Lukas and Seibold, Constantin and Keyl, Julius and Baldini, Giulia and Ugurel, Selma and Siveke, Jens and Gr{\"u}nwald, Barbara and Egger, Jan and others},
  journal={Medical image analysis},
  volume={94},
  pages={103143},
  year={2024},
  publisher={Elsevier}
}


%% file: resources/bibs/main.bib
@article{campanella2025clinical,
  title={A clinical benchmark of public self-supervised pathology foundation models},
  author={Campanella, Gabriele and Chen, Shengjia and Singh, Manbir and Verma, Ruchika and Muehlstedt, Silke and Zeng, Jennifer and Stock, Aryeh and Croken, Matt and Veremis, Brandon and Elmas, Abdulkadir and others},
  journal={Nature Communications},
  volume={16},
  number={1},
  pages={3640},
  year={2025},
  publisher={Nature Publishing Group UK London}
}

@article{ma2025pathbench,
  title={PathBench: A comprehensive comparison benchmark for pathology foundation models towards precision oncology},
  author={Ma, Jiabo and Xu, Yingxue and Zhou, Fengtao and Wang, Yihui and Jin, Cheng and Guo, Zhengrui and Wu, Jianfeng and Tang, On Ki and Zhou, Huajun and Wang, Xi and others},
  journal={arXiv preprint arXiv:2505.20202},
  year={2025}
}

@article{lipkova2024age,
  title={The age of foundation models},
  author={Lipkova, Jana and Kather, Jakob Nikolas},
  journal={Nature Reviews Clinical Oncology},
  volume={21},
  number={11},
  pages={769--770},
  year={2024},
  publisher={Nature Publishing Group UK London}
}

@article{weinstein2013cancer,
  title={The cancer genome atlas pan-cancer analysis project},
  author={Weinstein, John N and Collisson, Eric A and Mills, Gordon B and Shaw, Kenna R and Ozenberger, Brad A and Ellrott, Kyle and Shmulevich, Ilya and Sander, Chris and Stuart, Joshua M},
  journal={Nature genetics},
  volume={45},
  number={10},
  pages={1113--1120},
  year={2013},
  publisher={Nature Publishing Group}
}

@article{lonsdale2013genotype,
  title={The genotype-tissue expression (GTEx) project},
  author={Lonsdale, John and Thomas, Jeffrey and Salvatore, Mike and Phillips, Rebecca and Lo, Edmund and Shad, Saboor and Hasz, Richard and Walters, Gary and Garcia, Fernando and Young, Nancy and others},
  journal={Nature genetics},
  volume={45},
  number={6},
  pages={580--585},
  year={2013},
  publisher={Nature Publishing Group}
}

@article{weitz2024acrobat,
  title={The ACROBAT 2022 challenge: automatic registration of breast cancer tissue},
  author={Weitz, Philippe and Valkonen, Masi and Solorzano, Leslie and Carr, Circe and Kartasalo, Kimmo and Boissin, Constance and Koivukoski, Sonja and Kuusela, Aino and Rasic, Dusan and Feng, Yanbo and others},
  journal={Medical image analysis},
  volume={97},
  pages={103257},
  year={2024},
  publisher={Elsevier}
}

@article{xu2021predicting,
  title={Predicting axillary lymph node metastasis in early breast cancer using deep learning on primary tumor biopsy slides},
  author={Xu, Feng and Zhu, Chuang and Tang, Wenqi and Wang, Ying and Zhang, Yu and Li, Jie and Jiang, Hongchuan and Shi, Zhongyue and Liu, Jun and Jin, Mulan},
  journal={Frontiers in oncology},
  volume={11},
  pages={759007},
  year={2021},
  publisher={Frontiers Media SA}
}

@article{geijs2024detection,
  title={Detection and subtyping of basal cell carcinoma in whole-slide histopathology using weakly-supervised learning},
  author={Geijs, Daan J and Dooper, Stephan and Aswolinskiy, Witali and Hillen, Lisa M and Amir, Avital L and Litjens, Geert},
  journal={Medical Image Analysis},
  volume={93},
  pages={103063},
  year={2024},
  publisher={Elsevier}
}

@article{zhu2021development,
  title={Development and evaluation of a deep neural network for histologic classification of renal cell carcinoma on biopsy and surgical resection slides},
  author={Zhu, Mengdan and Ren, Bing and Richards, Ryland and Suriawinata, Matthew and Tomita, Naofumi and Hassanpour, Saeed},
  journal={Scientific reports},
  volume={11},
  number={1},
  pages={7080},
  year={2021},
  publisher={Nature Publishing Group UK London}
}

@article{wei2019pathologist,
  title={Pathologist-level classification of histologic patterns on resected lung adenocarcinoma slides with deep neural networks},
  author={Wei, Jason W and Tafe, Laura J and Linnik, Yevgeniy A and Vaickus, Louis J and Tomita, Naofumi and Hassanpour, Saeed},
  journal={Scientific reports},
  volume={9},
  number={1},
  pages={3358},
  year={2019},
  publisher={Nature Publishing Group UK London}
}

@article{neto2022imil4path,
  title={iMIL4PATH: A semi-supervised interpretable approach for colorectal whole-slide images},
  author={Neto, Pedro C and Oliveira, Sara P and Montezuma, Diana and Fraga, Jo{\~a}o and Monteiro, Ana and Ribeiro, Liliana and Gon{\c{c}}alves, Sofia and Pinto, Isabel M and Cardoso, Jaime S},
  journal={Cancers},
  volume={14},
  number={10},
  pages={2489},
  year={2022},
  publisher={MDPI}
}

@article{neto2024interpretable,
  title={An interpretable machine learning system for colorectal cancer diagnosis from pathology slides},
  author={Neto, Pedro C and Montezuma, Diana and Oliveira, Sara P and Oliveira, Domingos and Fraga, Jo{\~a}o and Monteiro, Ana and Monteiro, Jo{\~a}o and Ribeiro, Liliana and Gon{\c{c}}alves, Sofia and Reinhard, Stefan and others},
  journal={NPJ precision oncology},
  volume={8},
  number={1},
  pages={56},
  year={2024},
  publisher={Nature Publishing Group UK London}
}

@article{chauhan2024ipd,
  title={IPD-Brain: An Indian histopathology dataset for glioma subtype classification},
  author={Chauhan, Ekansh and Sharma, Amit and Uppin, Megha S and Kondamadugu, Manasa and Jawahar, CV and Vinod, PK},
  journal={Scientific Data},
  volume={11},
  number={1},
  pages={1403},
  year={2024},
  publisher={Nature Publishing Group UK London}
}

@article{zhang2019pathologist,
  title={Pathologist-level interpretable whole-slide cancer diagnosis with deep learning},
  author={Zhang, Zizhao and Chen, Pingjun and McGough, Mason and Xing, Fuyong and Wang, Chunbao and Bui, Marilyn and Xie, Yuanpu and Sapkota, Manish and Cui, Lei and Dhillon, Jasreman and others},
  journal={Nature Machine Intelligence},
  volume={1},
  number={5},
  pages={236--245},
  year={2019},
  publisher={Nature Publishing Group UK London}
}

@article{kim2021paip,
  title={PAIP 2019: Liver cancer segmentation challenge},
  author={Kim, Yoo Jung and Jang, Hyungjoon and Lee, Kyoungbun and Park, Seongkeun and Min, Sung-Gyu and Hong, Choyeon and Park, Jeong Hwan and Lee, Kanggeun and Kim, Jisoo and Hong, Wonjae and others},
  journal={Medical image analysis},
  volume={67},
  pages={101854},
  year={2021},
  publisher={Elsevier}
}

@article{bulten2022artificial,
  title={Artificial intelligence for diagnosis and Gleason grading of prostate cancer: the PANDA challenge},
  author={Bulten, Wouter and Kartasalo, Kimmo and Chen, Po-Hsuan Cameron and Str{\"o}m, Peter and Pinckaers, Hans and Nagpal, Kunal and Cai, Yuannan and Steiner, David F and Van Boven, Hester and Vink, Robert and others},
  journal={Nature medicine},
  volume={28},
  number={1},
  pages={154--163},
  year={2022},
  publisher={Nature Publishing Group US New York}
}

@article{lu2021data,
  title={Data-efficient and weakly supervised computational pathology on whole-slide images},
  author={Lu, Ming Y and Williamson, Drew FK and Chen, Tiffany Y and Chen, Richard J and Barbieri, Matteo and Mahmood, Faisal},
  journal={Nature biomedical engineering},
  volume={5},
  number={6},
  pages={555--570},
  year={2021},
  publisher={Nature Publishing Group UK London}
}

@article{lu2024visual,
  title={A visual-language foundation model for computational pathology},
  author={Lu, Ming Y and Chen, Bowen and Williamson, Drew FK and Chen, Richard J and Liang, Ivy and Ding, Tong and Jaume, Guillaume and Odintsov, Igor and Le, Long Phi and Gerber, Georg and others},
  journal={Nature Medicine},
  volume={30},
  number={3},
  pages={863--874},
  year={2024},
  publisher={Nature Publishing Group US New York}
}

@article{chen2024towards,
  title={Towards a general-purpose foundation model for computational pathology},
  author={Chen, Richard J and Ding, Tong and Lu, Ming Y and Williamson, Drew FK and Jaume, Guillaume and Song, Andrew H and Chen, Bowen and Zhang, Andrew and Shao, Daniel and Shaban, Muhammad and others},
  journal={Nature Medicine},
  volume={30},
  number={3},
  pages={850--862},
  year={2024},
  publisher={Nature Publishing Group US New York}
}

@article{xu2024whole,
  title={A whole-slide foundation model for digital pathology from real-world data},
  author={Xu, Hanwen and Usuyama, Naoto and Bagga, Jaspreet and Zhang, Sheng and Rao, Rajesh and Naumann, Tristan and Wong, Cliff and Gero, Zelalem and Gonz{\'a}lez, Javier and Gu, Yu and others},
  journal={Nature},
  volume={630},
  number={8015},
  pages={181--188},
  year={2024},
  publisher={Nature Publishing Group UK London}
}

@misc{hoptimus0,
  author = {Saillard, Charlie and Jenatton, Rodolphe and Llinares-López, Felipe and Mariet, Zelda and Cahané, David and Durand, Eric and Vert, Jean-Philippe},
  title = {H-optimus-0},
  url = {https://github.com/bioptimus/releases/tree/main/models/h-optimus/v0},
  year = {2024},
}

@article{zimmermann2024virchow2,
  title={Virchow2: Scaling self-supervised mixed magnification models in pathology},
  author={Zimmermann, Eric and Vorontsov, Eugene and Viret, Julian and Casson, Adam and Zelechowski, Michal and Shaikovski, George and Tenenholtz, Neil and Hall, James and Klimstra, David and Yousfi, Razik and others},
  journal={arXiv preprint arXiv:2408.00738},
  year={2024}
}

@article{xiang2025vision,
  title={A vision--language foundation model for precision oncology},
  author={Xiang, Jinxi and Wang, Xiyue and Zhang, Xiaoming and Xi, Yinghua and Eweje, Feyisope and Chen, Yijiang and Li, Yuchen and Bergstrom, Colin and Gopaulchan, Matthew and Kim, Ted and others},
  journal={Nature},
  pages={1--10},
  year={2025},
  publisher={Nature Publishing Group UK London}
}

@article{el2025whole,
  title={From whole-slide image to biomarker prediction: end-to-end weakly supervised deep learning in computational pathology},
  author={El Nahhas, Omar SM and van Treeck, Marko and W{\"o}lflein, Georg and Unger, Michaela and Ligero, Marta and Lenz, Tim and Wagner, Sophia J and Hewitt, Katherine J and Khader, Firas and Foersch, Sebastian and others},
  journal={Nature Protocols},
  volume={20},
  number={1},
  pages={293--316},
  year={2025},
  publisher={Nature Publishing Group UK London}
}

@inproceedings{ilse2018attention,
  title={Attention-based deep multiple instance learning},
  author={Ilse, Maximilian and Tomczak, Jakub and Welling, Max},
  booktitle={International conference on machine learning},
  pages={2127--2136},
  year={2018},
  organization={PMLR}
}

@article{wang2024pathology,
  title={A pathology foundation model for cancer diagnosis and prognosis prediction},
  author={Wang, Xiyue and Zhao, Junhan and Marostica, Eliana and Yuan, Wei and Jin, Jietian and Zhang, Jiayu and Li, Ruijiang and Tang, Hongping and Wang, Kanran and Li, Yu and others},
  journal={Nature},
  volume={634},
  number={8035},
  pages={970--978},
  year={2024},
  publisher={Nature Publishing Group UK London}
}

@article{vaidya2025molecular,
  title={Molecular-driven Foundation Model for Oncologic Pathology},
  author={Vaidya, Anurag and Zhang, Andrew and Jaume, Guillaume and Song, Andrew H and Ding, Tong and Wagner, Sophia J and Lu, Ming Y and Doucet, Paul and Robertson, Harry and Almagro-Perez, Cristina and others},
  journal={arXiv preprint arXiv:2501.16652},
  year={2025}
}

@inproceedings{lenz2025unsupervised,
  title={Unsupervised foundation model-agnostic slide-level representation learning},
  author={Lenz, Tim and Neidlinger, Peter and Ligero, Marta and W{\"o}lflein, Georg and van Treeck, Marko and Kather, Jakob N},
  booktitle={Proceedings of the Computer Vision and Pattern Recognition Conference},
  pages={30807--30817},
  year={2025}
}

@inproceedings{chen2021empirical,
  title={An empirical study of training self-supervised vision transformers},
  author={Chen, Xinlei and Xie, Saining and He, Kaiming},
  booktitle={Proceedings of the IEEE/CVF international conference on computer vision},
  pages={9640--9649},
  year={2021}
}

@article{oord2018representation,
  title={Representation learning with contrastive predictive coding},
  author={Oord, Aaron van den and Li, Yazhe and Vinyals, Oriol},
  journal={arXiv preprint arXiv:1807.03748},
  year={2018}
}

@inproceedings{he2022masked,
  title={Masked autoencoders are scalable vision learners},
  author={He, Kaiming and Chen, Xinlei and Xie, Saining and Li, Yanghao and Doll{\'a}r, Piotr and Girshick, Ross},
  booktitle={Proceedings of the IEEE/CVF conference on computer vision and pattern recognition},
  pages={16000--16009},
  year={2022}
}

@article{ding2024multimodal,
  title={A multimodal whole-slide foundation model for pathology},
  author={Ding, Tong and Wagner, Sophia J and Song, Andrew H and Chen, Richard J and Lu, Ming Y and Zhang, Andrew and Vaidya, Anurag J and Jaume, Guillaume and Shaban, Muhammad and Kim, Ahrong and others},
  journal={Nature medicine},
  pages={1--13},
  year={2025},
  publisher={Nature Publishing Group US New York}
}

@article{roetzer2022digital,
  title={The digital brain tumour atlas, an open histopathology resource},
  author={Roetzer-Pejrimovsky, Thomas and Moser, Anna-Christina and Atli, Baran and Vogel, Clemens Christian and Mercea, Petra A and Prihoda, Romana and Gelpi, Ellen and Haberler, Christine and H{\"o}ftberger, Romana and Hainfellner, Johannes A and others},
  journal={Scientific Data},
  volume={9},
  number={1},
  pages={55},
  year={2022},
  publisher={Nature Publishing Group UK London}
}

@article{yacob2023weakly,
  title={Weakly supervised detection and classification of basal cell carcinoma using graph-transformer on whole slide images},
  author={Yacob, Filmon and Siarov, Jan and Villiamsson, Kajsa and Suvilehto, Juulia T and Sj{\"o}blom, Lisa and Kjellberg, Magnus and Neittaanm{\"a}ki, Noora},
  journal={Scientific Reports},
  volume={13},
  number={1},
  pages={7555},
  year={2023},
  publisher={Nature Publishing Group UK London}
}

@article{brancati2022bracs,
  title={Bracs: A dataset for breast carcinoma subtyping in h\&e histology images},
  author={Brancati, Nadia and Anniciello, Anna Maria and Pati, Pushpak and Riccio, Daniel and Scognamiglio, Giosu{\`e} and Jaume, Guillaume and De Pietro, Giuseppe and Di Bonito, Maurizio and Foncubierta, Antonio and Botti, Gerardo and others},
  journal={Database},
  volume={2022},
  pages={baac093},
  year={2022},
  publisher={Oxford University Press UK}
}

@article{swanton2012my,
  title={My Cancer Genome: a unified genomics and clinical trial portal},
  author={Swanton, Charles},
  journal={The Lancet Oncology},
  volume={13},
  number={7},
  pages={668--669},
  year={2012},
  publisher={Elsevier}
}

@article{wang2024screen,
  title={Screen them all: high-throughput pan-cancer genetic and phenotypic biomarker screening from H\&E whole slide images},
  author={Wang, Yi Kan and Tydlitatova, Ludmila and Kunz, Jeremy D and Oakley, Gerard and Godrich, Ran A and Lee, Matthew CH and Vanderbilt, Chad and Yousfi, Razik and Fuchs, Thomas and Klimstra, David S and others},
  journal={arXiv preprint arXiv:2408.09554},
  year={2024}
}

@article{suehnholz2020oncokb,
  title={OncoKB, a precision oncology knowledgebase},
  author={Suehnholz, Sarah and Zhang, Hongxin and Nissan, Moriah and Kundra, Ritika and Su, Jing and LaFave, Lindsay and Gala, Kinisha and Vanderbilt, Chad and Arcila, Maria and Ladanyi, Marc and others},
  journal={Cancer Research},
  volume={80},
  number={16\_Supplement},
  pages={3208--3208},
  year={2020},
  publisher={The American Association for Cancer Research}
}

@article{ward2015mco,
  title={MCO study whole slide image collection},
  author={Ward, Robyn},
  year={2015}
}

@incollection{jonnagaddala2016integration,
  title={Integration and analysis of heterogeneous colorectal cancer data for translational research},
  author={Jonnagaddala, Jitendra and Croucher, Joanne L and Jue, Toni Rose and Meagher, Nicola S and Caruso, Lena and Ward, Robyn and Hawkins, Nicholas J},
  booktitle={Nursing Informatics 2016},
  pages={387--391},
  year={2016},
  publisher={IOS Press}
}

@article{myles2025surgen,
  title={Surgen: 1020 h\&e-stained whole slide images with survival and genetic markers},
  author={Myles, Craig and Um, In Hwa and Marshall, Craig and Harris-Birtill, David and Harrison, David J},
  journal={arXiv preprint arXiv:2502.04946},
  year={2025}
}

@article{bergstrom2024deep,
  title={Deep learning artificial intelligence predicts homologous recombination deficiency and platinum response from histologic slides},
  author={Bergstrom, Erik N and Abbasi, Ammal and D{\'\i}az-Gay, Marcos and Galland, Lo{\"\i}ck and Ladoire, Sylvain and Lippman, Scott M and Alexandrov, Ludmil B},
  journal={Journal of Clinical Oncology},
  volume={42},
  number={30},
  pages={3550--3560},
  year={2024},
  publisher={Wolters Kluwer Health}
}

@article{eisenhauer2009new,
  title={New response evaluation criteria in solid tumours: revised RECIST guideline (version 1.1)},
  author={Eisenhauer, Elizabeth A and Therasse, Patrick and Bogaerts, Jan and Schwartz, Lawrence H and Sargent, Danielle and Ford, Robert and Dancey, Janet and Arbuck, Stephen and Gwyther, Steve and Mooney, Margaret and others},
  journal={European journal of cancer},
  volume={45},
  number={2},
  pages={228--247},
  year={2009},
  publisher={Elsevier}
}

@article{farahmand2022her2,
  title={HER2 and trastuzumab treatment response H\&E slides with tumor ROI annotations},
  author={Farahmand, S and Fernandez, AI and Ahmed, FS and Rimm, DL and Chuang, JH and Reisenbichler, E and Zarringhalam, K},
  journal={The Cancer Imaging Archive)},
  year={2022}
}

@article{chowdhury2023proteogenomic,
  title={Proteogenomic analysis of chemo-refractory high-grade serous ovarian cancer},
  author={Chowdhury, Shrabanti and Kennedy, Jacob J and Ivey, Richard G and Murillo, Oscar D and Hosseini, Noshad and Song, Xiaoyu and Petralia, Francesca and Calinawan, Anna and Savage, Sara R and Berry, Anna B and others},
  journal={Cell},
  volume={186},
  number={16},
  pages={3476--3498},
  year={2023},
  publisher={Elsevier}
}

@article{wang2022histopathological,
  title={Histopathological whole slide image dataset for classification of treatment effectiveness to ovarian cancer},
  author={Wang, Ching-Wei and Chang, Cheng-Chang and Khalil, Muhammad Adil and Lin, Yi-Jia and Liou, Yi-An and Hsu, Po-Chao and Lee, Yu-Ching and Wang, Chih-Hung and Chao, Tai-Kuang},
  journal={Scientific Data},
  volume={9},
  number={1},
  pages={25},
  year={2022},
  publisher={Nature Publishing Group UK London}
}

@article{sammut2022multi,
  title={Multi-omic machine learning predictor of breast cancer therapy response},
  author={Sammut, Stephen-John and Crispin-Ortuzar, Mireia and Chin, Suet-Feung and Provenzano, Elena and Bardwell, Helen A and Ma, Wenxin and Cope, Wei and Dariush, Ali and Dawson, Sarah-Jane and Abraham, Jean E and others},
  journal={Nature},
  volume={601},
  number={7894},
  pages={623--629},
  year={2022},
  publisher={Nature Publishing Group UK London}
}

@article{huang2023artificial,
  title={Artificial intelligence reveals features associated with breast cancer neoadjuvant chemotherapy responses from multi-stain histopathologic images},
  author={Huang, Zhi and Shao, Wei and Han, Zhi and Alkashash, Ahmad Mahmoud and De la Sancha, Carlo and Parwani, Anil V and Nitta, Hiroaki and Hou, Yanjun and Wang, Tongxin and Salama, Paul and others},
  journal={NPJ precision oncology},
  volume={7},
  number={1},
  pages={14},
  year={2023},
  publisher={Nature Publishing Group UK London}
}

@article{callahan2023stanford,
  title={The Stanford Medicine data science ecosystem for clinical and translational research},
  author={Callahan, Alison and Ashley, Euan and Datta, Somalee and Desai, Priyamvada and Ferris, Todd A and Fries, Jason A and Halaas, Michael and Langlotz, Curtis P and Mackey, Sean and Posada, Jos{\'e} D and others},
  journal={JAMIA open},
  volume={6},
  number={3},
  pages={ooad054},
  year={2023},
  publisher={Oxford University Press}
}

@misc{cancerMoonshot2022,
  author       = {{Cancer Moonshot Biobank}},
  title        = {Cancer Moonshot Biobank - Lung Cancer Collection (CMB-LCA)},
  year         = {2022},
  publisher    = {The Cancer Imaging Archive},
  doi          = {10.7937/3CX3-S132},
  note         = {Version 3, dataset},
  url          = {https://doi.org/10.7937/3CX3-S132}
}

@article{hoang2024deep,
  title={A deep-learning framework to predict cancer treatment response from histopathology images through imputed transcriptomics},
  author={Hoang, Danh-Tai and Dinstag, Gal and Shulman, Eldad D and Hermida, Leandro C and Ben-Zvi, Doreen S and Elis, Efrat and Caley, Katherine and Sammut, Stephen-John and Sinha, Sanju and Sinha, Neelam and others},
  journal={Nature cancer},
  volume={5},
  number={9},
  pages={1305--1317},
  year={2024},
  publisher={Nature Publishing Group US New York}
}

@article{bagaev2021conserved,
  title={Conserved pan-cancer microenvironment subtypes predict response to immunotherapy},
  author={Bagaev, Alexander and Kotlov, Nikita and Nomie, Krystle and Svekolkin, Viktor and Gafurov, Azamat and Isaeva, Olga and Osokin, Nikita and Kozlov, Ivan and Frenkel, Felix and Gancharova, Olga and others},
  journal={Cancer cell},
  volume={39},
  number={6},
  pages={845--865},
  year={2021},
  publisher={Elsevier}
}

@article{passaro2024cancer,
  title={Cancer biomarkers: Emerging trends and clinical implications for personalized treatment},
  author={Passaro, Antonio and Al Bakir, Maise and Hamilton, Emily G and Diehn, Maximilian and Andr{\'e}, Fabrice and Roy-Chowdhuri, Sinchita and Mountzios, Giannis and Wistuba, Ignacio I and Swanton, Charles and Peters, Solange},
  journal={Cell},
  volume={187},
  number={7},
  pages={1617--1635},
  year={2024},
  publisher={Elsevier}
}

@article{prelaj2024artificial,
  title={Artificial intelligence for predictive biomarker discovery in immuno-oncology: a systematic review},
  author={Prelaj, Arsela and Miskovic, V and Zanitti, M and Trovo, F and Genova, C and Viscardi, Giuseppe and Rebuzzi, SE and Mazzeo, Laura and Provenzano, L and Kosta, S and others},
  journal={Annals of Oncology},
  volume={35},
  number={1},
  pages={29--65},
  year={2024},
  publisher={Elsevier}
}

@article{cristescu2018pan,
  title={Pan-tumor genomic biomarkers for PD-1 checkpoint blockade--based immunotherapy},
  author={Cristescu, Razvan and Mogg, Robin and Ayers, Mark and Albright, Andrew and Murphy, Erin and Yearley, Jennifer and Sher, Xinwei and Liu, Xiao Qiao and Lu, Hongchao and Nebozhyn, Michael and others},
  journal={Science},
  volume={362},
  number={6411},
  pages={eaar3593},
  year={2018},
  publisher={American Association for the Advancement of Science}
}

@article{samstein2019tumor,
  title={Tumor mutational load predicts survival after immunotherapy across multiple cancer types},
  author={Samstein, Robert M and Lee, Chung-Han and Shoushtari, Alexander N and Hellmann, Matthew D and Shen, Ronglai and Janjigian, Yelena Y and Barron, David A and Zehir, Ahmet and Jordan, Emmet J and Omuro, Antonio and others},
  journal={Nature genetics},
  volume={51},
  number={2},
  pages={202--206},
  year={2019},
  publisher={Nature Publishing Group US New York}
}

@article{taylor2018genomic,
  title={Genomic and functional approaches to understanding cancer aneuploidy},
  author={Taylor, Alison M and Shih, Juliann and Ha, Gavin and Gao, Galen F and Zhang, Xiaoyang and Berger, Ashton C and Schumacher, Steven E and Wang, Chen and Hu, Hai and Liu, Jianfang and others},
  journal={Cancer cell},
  volume={33},
  number={4},
  pages={676--689},
  year={2018},
  publisher={Elsevier}
}

@article{ma2024towards,
  title={Towards a generalizable pathology foundation model via unified knowledge distillation},
  author={Ma, Jiabo and Guo, Zhengrui and Zhou, Fengtao and Wang, Yihui and Xu, Yingxue and Li, Jinbang and Yan, Fang and Cai, Yu and Zhu, Zhengjie and Jin, Cheng and others},
  journal={arXiv preprint arXiv:2407.18449},
  year={2024}
}

@article{wang2022transformer,
  title={Transformer-based unsupervised contrastive learning for histopathological image classification},
  author={Wang, Xiyue and Yang, Sen and Zhang, Jun and Wang, Minghui and Zhang, Jing and Yang, Wei and Huang, Junzhou and Han, Xiao},
  journal={Medical image analysis},
  volume={81},
  pages={102559},
  year={2022},
  publisher={Elsevier}
}

@inproceedings{chen2016xgboost,
  title={Xgboost: A scalable tree boosting system},
  author={Chen, Tianqi and Guestrin, Carlos},
  booktitle={Proceedings of the 22nd acm sigkdd international conference on knowledge discovery and data mining},
  pages={785--794},
  year={2016}
}

@article{quinton2021whole,
  title={Whole-genome doubling confers unique genetic vulnerabilities on tumour cells},
  author={Quinton, Ryan J and DiDomizio, Amanda and Vittoria, Marc A and Kot{\`y}nkov{\'a}, Krist{\`y}na and Ticas, Carlos J and Patel, Sheena and Koga, Yusuke and Vakhshoorzadeh, Jasmine and Hermance, Nicole and Kuroda, Taruho S and others},
  journal={Nature},
  volume={590},
  number={7846},
  pages={492--497},
  year={2021},
  publisher={Nature Publishing Group UK London}
}

@article{lambuta2023whole,
  title={Whole-genome doubling drives oncogenic loss of chromatin segregation},
  author={Lambuta, Ruxandra A and Nanni, Luca and Liu, Yuanlong and Diaz-Miyar, Juan and Iyer, Arvind and Tavernari, Daniele and Katanayeva, Natalya and Ciriello, Giovanni and Oricchio, Elisa},
  journal={Nature},
  volume={615},
  number={7954},
  pages={925--933},
  year={2023},
  publisher={Nature Publishing Group UK London}
}

@article{marcus2021fda,
  title={FDA approval summary: pembrolizumab for the treatment of tumor mutational burden--high solid tumors},
  author={Marcus, Leigh and Fashoyin-Aje, Lola A and Donoghue, Martha and Yuan, Mengdie and Rodriguez, Lisa and Gallagher, Pamela S and Philip, Reena and Ghosh, Soma and Theoret, Marc R and Beaver, Julia A and others},
  journal={Clinical Cancer Research},
  volume={27},
  number={17},
  pages={4685--4689},
  year={2021},
  publisher={American Association for Cancer Research}
}

@article{yarchoan2017tumor,
  title={Tumor mutational burden and response rate to PD-1 inhibition},
  author={Yarchoan, Mark and Hopkins, Alexander and Jaffee, Elizabeth M},
  journal={New England Journal of Medicine},
  volume={377},
  number={25},
  pages={2500--2501},
  year={2017},
  publisher={Mass Medical Soc}
}

@article{bera2019artificial,
  title={Artificial intelligence in digital pathology---new tools for diagnosis and precision oncology},
  author={Bera, Kaustav and Schalper, Kurt A and Rimm, David L and Velcheti, Vamsidhar and Madabhushi, Anant},
  journal={Nature reviews Clinical oncology},
  volume={16},
  number={11},
  pages={703--715},
  year={2019},
  publisher={Nature Publishing Group UK London}
}

@article{yates2025new,
  title={New horizons at the interface of artificial intelligence and translational cancer research},
  author={Yates, Josephine and Van Allen, Eliezer M},
  journal={Cancer Cell},
  volume={43},
  number={4},
  pages={708--727},
  year={2025},
  publisher={Elsevier}
}

@article{bhinder2021artificial,
  title={Artificial intelligence in cancer research and precision medicine},
  author={Bhinder, Bhavneet and Gilvary, Coryandar and Madhukar, Neel S and Elemento, Olivier},
  journal={Cancer discovery},
  volume={11},
  number={4},
  pages={900--915},
  year={2021},
  publisher={American Association for Cancer Research}
}

@article{van2021deep,
  title={Deep learning in histopathology: the path to the clinic},
  author={Van der Laak, Jeroen and Litjens, Geert and Ciompi, Francesco},
  journal={Nature medicine},
  volume={27},
  number={5},
  pages={775--784},
  year={2021},
  publisher={Nature Publishing Group US New York}
}

@article{vorontsov2024foundation,
  title={A foundation model for clinical-grade computational pathology and rare cancers detection},
  author={Vorontsov, Eugene and Bozkurt, Alican and Casson, Adam and Shaikovski, George and Zelechowski, Michal and Severson, Kristen and Zimmermann, Eric and Hall, James and Tenenholtz, Neil and Fusi, Nicolo and others},
  journal={Nature medicine},
  volume={30},
  number={10},
  pages={2924--2935},
  year={2024},
  publisher={Nature Publishing Group US New York}
}

@article{wang2025foundation,
  title={Foundation model for predicting prognosis and adjuvant therapy benefit from digital pathology in GI cancers},
  author={Wang, Xiyue and Jiang, Yuming and Yang, Sen and Wang, Fang and Zhang, Xiaoming and Wang, Wei and Chen, Yijiang and Wu, Xiaoyan and Xiang, Jinxi and Li, Yuchen and others},
  journal={Journal of Clinical Oncology},
  pages={JCO--24},
  year={2025},
  publisher={Wolters Kluwer Health}
}

@article{campanella2025real,
  title={Real-world deployment of a fine-tuned pathology foundation model for lung cancer biomarker detection},
  author={Campanella, Gabriele and Kumar, Neeraj and Nanda, Swaraj and Singi, Siddharth and Fluder, Eugene and Kwan, Ricky and Muehlstedt, Silke and Pfarr, Nicole and Sch{\"u}ffler, Peter J and H{\"a}ggstr{\"o}m, Ida and others},
  journal={Nature Medicine},
  pages={1--9},
  year={2025},
  publisher={Nature Publishing Group US New York}
}

@article{kondepudi2025foundation,
  title={Foundation models for fast, label-free detection of glioma infiltration},
  author={Kondepudi, Akhil and Pekmezci, Melike and Hou, Xinhai and Scotford, Katie and Jiang, Cheng and Rao, Akshay and Harake, Edward S and Chowdury, Asadur and Al-Holou, Wajd and Wang, Lin and others},
  journal={Nature},
  volume={637},
  number={8045},
  pages={439--445},
  year={2025},
  publisher={Nature Publishing Group UK London}
}

@article{nabet2020noninvasive,
  title={Noninvasive early identification of therapeutic benefit from immune checkpoint inhibition},
  author={Nabet, Barzin Y and Esfahani, Mohammad S and Moding, Everett J and Hamilton, Emily G and Chabon, Jacob J and Rizvi, Hira and Steen, Chloe B and Chaudhuri, Aadel A and Liu, Chih Long and Hui, Angela B and others},
  journal={Cell},
  volume={183},
  number={2},
  pages={363--376},
  year={2020},
  publisher={Elsevier}
}

@article{louie2024molecular,
  title={Molecular Tumor Board for Unicorns: Outcomes for rare and ultra-rare cancers using an N-of-One personalized treatment strategy},
  author={Louie, Bryan H and Kato, Shumei and Lim, Jordan S and Kim, Ki Hwan and Lim, Hyo Jeong and Okamura, Ryosuke and Lee, Suzanna and Kim, Lisa and Sicklick, Jason K and Lippman, Scott M and others},
  journal={Iscience},
  volume={27},
  number={8},
  year={2024},
  publisher={Elsevier}
}

@article{shao2025mil,
  title={Do MIL Models Transfer?},
  author={Shao, Daniel and Chen, Richard J and Song, Andrew H and Runevic, Joel and Lu, Ming Y and Ding, Tong and Mahmood, Faisal},
  journal={arXiv preprint arXiv:2506.09022},
  year={2025}
}

@article{neidlinger2024benchmarking,
  title={Benchmarking foundation models as feature extractors for weakly supervised computational pathology},
  author={Neidlinger, Peter and El Nahhas, Omar SM and Muti, Hannah Sophie and Lenz, Tim and Hoffmeister, Michael and Brenner, Hermann and van Treeck, Marko and Langer, Rupert and Dislich, Bastian and Behrens, Hans Michael and others},
  journal={Nature Biomedical Engineering},
  pages={1--11},
  year={2025},
  publisher={Nature Publishing Group UK London}
}

@article{luo2025nnmil,
  title={nnMIL: A generalizable multiple instance learning framework for computational pathology},
  author={Luo, Xiangde and Xiang, Jinxi and Ji, Yuanfeng and Li, Ruijiang},
  journal={arXiv preprint arXiv:2511.14907},
  year={2025}
}

@article{ba2016layer,
  title={Layer normalization},
  author={Ba, Jimmy Lei and Kiros, Jamie Ryan and Hinton, Geoffrey E},
  journal={arXiv preprint arXiv:1607.06450},
  year={2016}
}

@inproceedings{filiot2025distilling,
  title={Distilling foundation models for robust and efficient models in digital pathology},
  author={Filiot, Alexandre and Dop, Nicolas and Tchita, Oussama and Riou, Auriane and Dubois, R{\'e}my and Peeters, Thomas and Valter, Daria and Scalbert, Marin and Saillard, Charlie and Robin, Genevi{\`e}ve and others},
  booktitle={International Conference on Medical Image Computing and Computer-Assisted Intervention},
  pages={162--172},
  year={2025},
  organization={Springer}
}

@article{kaplan2025openmidnight,
    author = {Kaplan, Daniel and Grandhi, Ratna Sagari and Lane, Connor and Warner, Benjamin and Abraham, Tanishq Mathew and Scotti, Paul S.},
    title = {How to Train a State-of-the-Art Pathology Foundation Model with \$1.6k},
    year = {2025},
    url = {https://sophont.med/blog/openmidnight}}

@misc{kapfer2025marlowe,
  title={Marlowe: Stanford's gpu-based computational instrument},
  author={Kapfer, Craig and Stine, Kurt and Narasimhan, Balasubramanian and Mentzel, Christopher and Candes, Emmanuel},
  year={2025},
  publisher={January}
}

@inproceedings{zhaosuperclip,
  title={SuperCLIP: CLIP with Simple Classification Supervision},
  author={Zhao, Weiheng and Huang, Zilong and Feng, Jiashi and Wang, Xinggang},
  booktitle={The Thirty-ninth Annual Conference on Neural Information Processing Systems}
}

@article{bareja2025evaluating,
  title={Evaluating Vision and Pathology Foundation Models for Computational Pathology: A Comprehensive Benchmark Study},
  author={Bareja, Rohan and Carrillo-Perez, Francisco and Zheng, Yuanning and Pizurica, Marija and Nandi, Tarak Nath and Shen, Jeanne and Madduri, Ravi and Gevaert, Olivier},
  journal={medRxiv},
  pages={2025--05},
  year={2025},
  publisher={Cold Spring Harbor Laboratory Press}
}

@article{cheng2015memorial,
  title={Memorial Sloan Kettering-Integrated Mutation Profiling of Actionable Cancer Targets (MSK-IMPACT): a hybridization capture-based next-generation sequencing clinical assay for solid tumor molecular oncology},
  author={Cheng, Donavan T and Mitchell, Talia N and Zehir, Ahmet and Shah, Ronak H and Benayed, Ryma and Syed, Aijazuddin and Chandramohan, Raghu and Liu, Zhen Yu and Won, Helen H and Scott, Sasinya N and others},
  journal={The Journal of molecular diagnostics},
  volume={17},
  number={3},
  pages={251--264},
  year={2015},
  publisher={Elsevier}
}

@article{chakravarty2017oncokb,
  title={OncoKB: a precision oncology knowledge base},
  author={Chakravarty, Debyani and Gao, Jianjiong and Phillips, Sarah and Kundra, Ritika and Zhang, Hongxin and Wang, Jiaojiao and Rudolph, Julia E and Yaeger, Rona and Soumerai, Tara and Nissan, Moriah H and others},
  journal={JCO precision oncology},
  volume={1},
  pages={1--16},
  year={2017},
  publisher={American Society of Clinical Oncology}
}
